\documentclass{article} 
\usepackage{iclr2027_conference,times}

\usepackage[T1]{fontenc}
\usepackage[utf8]{inputenc}
\usepackage{microtype}
\usepackage{inconsolata}
\usepackage{tocloft}

\usepackage{amsmath,amsfonts,bm}

\def\eqref#1{equation~\ref{#1}}

\def\1{\bm{1}}

\DeclareMathAlphabet{\mathsfit}{\encodingdefault}{\sfdefault}{m}{sl}
\SetMathAlphabet{\mathsfit}{bold}{\encodingdefault}{\sfdefault}{bx}{n}

\usepackage{xcolor}
\definecolor{uclablue}{rgb}{0.15,0.45,0.68}
\definecolor{urlpink}{HTML}{FF1493}
\usepackage[colorlinks=true,citecolor=uclablue,linkcolor=uclablue,urlcolor=urlpink]{hyperref}
\hypersetup{
    pdftitle={A Pretrained Model for Bayesian Filtering in Heterogeneous Dynamical Systems},
    pdfauthor={Ke Fan, Wen Wu, Haitao Su, Yilin Wang, Yuan Zhou, Chao Zhang}
}
\usepackage{graphicx}
\usepackage{booktabs}
\usepackage{url}
\usepackage{float}
\usepackage{colortbl}
\usepackage{capt-of}
\usepackage{wrapfig}
\usepackage[skins,breakable,listings]{tcolorbox}
\newcommand{\relatedparagraph}[1]{\par\vspace{0.25em}\noindent\textbf{#1}}
\newenvironment{baselinelist}{%
  \begin{list}{\textbullet}{%
    \setlength{\leftmargin}{1.2em}%
    \setlength{\labelwidth}{0.6em}%
    \setlength{\labelsep}{0.6em}%
    \setlength{\rightmargin}{0pt}%
    \setlength{\itemindent}{0pt}%
    \setlength{\listparindent}{0pt}%
    \setlength{\topsep}{4pt plus 1pt minus 1pt}%
    \setlength{\partopsep}{0pt}%
    \setlength{\itemsep}{5pt plus 1pt minus 1pt}%
    \setlength{\parsep}{3pt plus 1pt minus 1pt}%
  }%
}{\end{list}}

\tcbset{systemprompt/.style={
  enhanced,
  breakable,
  listing only,
  colback=uclablue!5,
  coltext=black,
  colframe=uclablue!50,
  colbacktitle=uclablue!20,
  coltitle=uclablue!45!black,
  fonttitle=\bfseries,
  arc=1mm,
  boxrule=0.7pt,
  left=1.5mm,
  right=1.5mm,
  top=1mm,
  bottom=1mm,
  before skip=8pt,
  after skip=10pt,
  listing options={
    basicstyle=\color{black}\ttfamily\fontsize{6.4}{7.4}\selectfont,
    breaklines=true,
    breakatwhitespace=false,
    columns=fullflexible,
    keepspaces=true,
    showstringspaces=false
  }
}}

\newcommand{\best}[1]{\textcolor{red}{\textbf{#1}}}
\newcommand{\runnerup}[1]{\textcolor{blue}{\underline{#1}}}
\newcommand{\third}[1]{\textcolor{teal}{\textit{#1}}}

\title{A Pretrained Model for Bayesian Filtering in Heterogeneous Dynamical Systems}

\author{
    \textbf{Ke Fan}$^{1,2}$\thanks{This work is done when Ke Fan is an intern in Shanghai AI Laboratory.} \quad
    \textbf{Wen Wu}$^{1}$ \quad
    \textbf{Haitao Su}$^{2}$ \quad
    \textbf{Yilin Wang}$^{1}$ \quad
    \textbf{Yuan Zhou}$^{2}$ \quad
    \textbf{Chao Zhang}$^{1,2}$\thanks{The corresponding author.} \\
    $^{1}$Shanghai AI Laboratory \quad
    $^{2}$Tsinghua University \\
    \texttt{fank22@mails.tsinghua.edu.cn} \quad
    \texttt{sht23@mails.tsinghua.edu.cn} \\
    \texttt{yuan-zhou@tsinghua.edu.cn} \quad
    \texttt{\{wuwen,wangyilin,zhangchao\}@pjlab.org.cn}
}

\iclrfinalcopy
\begin{document}

\addtocontents{toc}{\protect\setcounter{tocdepth}{-1}}

\maketitle
\fancyhead{}

\begin{abstract}
Filtering has not yet had its foundation-model moment: state estimation is still dominated either by online Bayesian filtering algorithms that require known dynamics as inputs, or by learning-based neural filters trained from scratch for a single system or parametric family. The core challenge is structural heterogeneity: dynamical systems differ in their governing dynamics, physical scales, and noise levels, precluding a shared representation space for a reusable filter. To bridge this gap, we introduce \emph{FilterFM}, the \emph{first} pretrained model for Bayesian filtering that estimates latent states across heterogeneous dynamical systems with a single checkpoint. FilterFM employs a hierarchical two-level observation sequence encoder and a prior-to-posterior state-reconstruction module that together handle varying observation and state dimensions across heterogeneous systems, while modeling state uncertainty via quantile prediction. Trained on a mixture of heterogeneous filtering tasks, FilterFM exhibits superior in-domain performance, delivering significant gains in estimation accuracy, runtime efficiency, and noise robustness compared to existing baselines. Beyond in-domain performance, it marks the first model to successfully enable zero-shot filtering on unseen dynamics, bridging the generalization gap in state estimation. Furthermore, we uncover clear scaling behaviors in foundation-model-based Bayesian filtering, which combined with comprehensive component ablation studies, validate the structural optimality of our design and broaden the generalization scope for this research field.
\end{abstract}

\section{Introduction}
\label{sec:intro}
Bayesian filtering serves as a basic tool for state estimation in robotics~\citep{Barfoot_2024}, navigation~\citep{GP-Bayes_Ko}, atmospheric science~\citep{hickman2025neurips-causal}, chemical process monitoring~\citep{s19102240}, and many other scientific domains~\citep{10.1016/j.inffus.2023.101971,Course2023}. 
Although foundation models have reshaped multi-modal modeling and time-series forecasting, filtering still lacks a foundation-model paradigm shift. Classical methods from Gaussian filters~\citep{7528889,855552,11434934}, particle filters~\citep{8683411} to PDE-based filters~\citep{wang2025yauyaualcomputertoolsolving,Yau_2025} instantiate a predict-then-update recursion, their strong performances usually depend on exactly known system-specific transition models, observation mappings, and noise assumptions. 
Learning-based filters reduce this modeling burden by training neural estimators from trajectories~\citep{KalmanNet,RStateNet}, but they are still typically tied to one specific dynamical system~\citep{MEStimator}, or to a parametric family sharing the same underlying dynamics~\citep{FM-UKF}. 
This raises a natural foundation-model question:
\begin{tcolorbox}[
  enhanced, sharp corners,
  colback=black!4, colframe=black!75,
  boxrule=0.6pt,
  boxsep=0pt, left=7pt, right=7pt, top=4pt, bottom=4pt,
  before skip=5pt, after skip=6pt
]
\centering\itshape
Can one model learn to filter across heterogeneous dynamical systems?
\end{tcolorbox}

\begin{figure}[!t]
\centering
\setlength{\parskip}{0pt}
\setlength{\abovecaptionskip}{4pt}
\setlength{\belowcaptionskip}{0pt}
\includegraphics[width=1.00\linewidth]{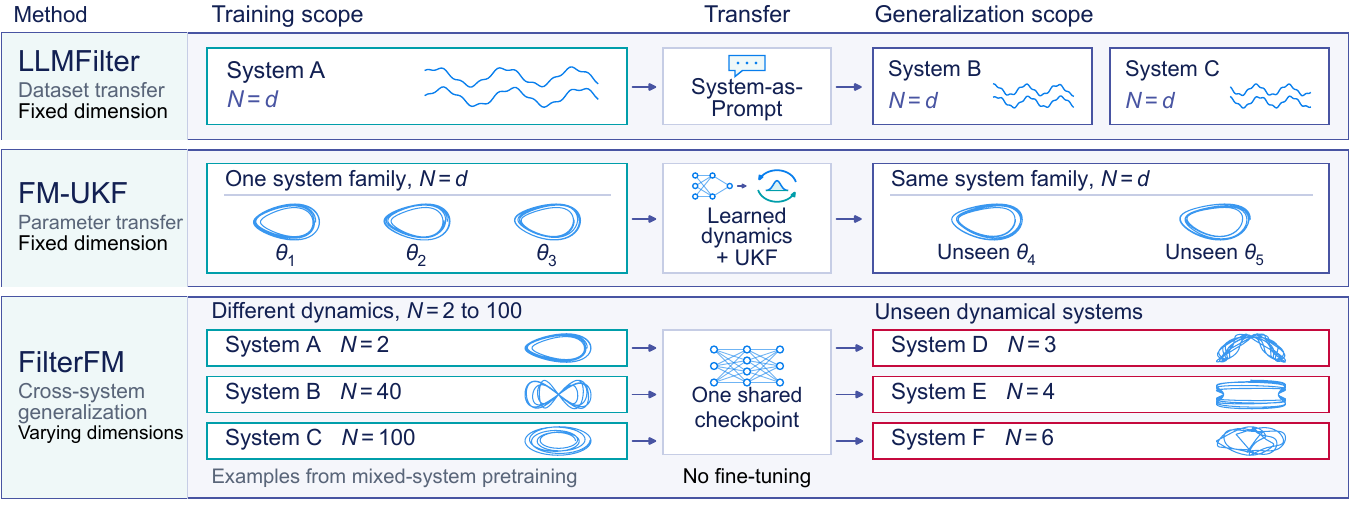}\par\nointerlineskip
\caption{Generalization scopes of FilterFM, LLMFilter, and FM-UKF. LLMFilter trains on one system at a time and transfers across datasets at fixed state dimension \(N\) via System-as-Prompt. FM-UKF learns dynamics across parameter settings within one system family and uses a UKF with a known observation model. FilterFM is pretrained on a corpus of heterogeneous systems and evaluated on unseen systems using the same checkpoint without fine-tuning.}
\label{fig:filtering_dataset_taxonomy}
\end{figure}

The challenges lie in structural heterogeneity rather than data scale alone: dynamical systems differ in their governing equations~\citep{SINDY}, observation processes, physical units, and noise levels~\citep{10912639}, so a reusable filter cannot rely on a fixed state-observation interface. 
Existing foundation-inspired attempts only partially address this issue. LLMFilter~\citep{LLMFilter} adapts a frozen LLM with system prompts, but remains trained on single datasets and flattens time and variable channels into tokens with unclear semantics. 
FM-UKF~\citep{FM-UKF} inserts a time-series foundation model into a Bayesian filtering pipeline, but still requires a specified observation map and is evaluated mainly within one parametric family. 

To fill this research gap, we introduce FilterFM, a transformer-based pretrained model for Bayesian filtering. Unlike LLMFilter and FM-UKF, FilterFM is pretrained across heterogeneous systems with varying dynamics and dimensions and is able to generalize to unseen systems with a single checkpoint. Figure~\ref{fig:filtering_dataset_taxonomy} compares the three approaches' training and generalization scopes across datasets, system parameters, and state dimensions. Our model consists of a hierarchical two-level causal observation sequence encoder coupled with a prior-to-posterior state-reconstruction module. The encoder's design is motivated by two requirements: high-dimensional systems make dense attention over all time-variable pairs inefficient, and Bayesian filtering needs the current observation to update posterior rather than just coarse patch summaries. Accordingly, patch-level blocks compress completed patches into \emph{causal history memory}, while intra-patch-level blocks preserve local evidence at each time step through \emph{cross-scale conditioning}. This design retains the per-step resolution that Bayesian filtering requires for real-time correction based on the current observation. Within each scale, causal temporal attention first models each variable's temporal dependence. The resulting features are summarized into compact prototype memory and exchanged across variables through a gated reassembly router~\citep{liu2026falconxtimeseriesfoundation}. The state-reconstruction module then performs the filtering update: it first reads \emph{the shifted local memory to form a prior}, then incorporates current observation to obtain a posterior, and decodes the posterior into the latent state estimate. Also to the best of our knowledge, FilterFM is the first model to support uncertainty estimation for Bayesian filtering, decoding multiple posterior quantiles to provide both a median state estimate and predictive uncertainty intervals.

To train FilterFM, we build a training dataset containing approximately 15K training trajectories, with about 45\% from filtering numerical examples commonly used in previous literatures~\citep{LLMFilter}, 20\% from synthetic low-dimensional chaotic systems~\citep{lai2026panda}, and 35\% from high-dimensional composites formed by concatenating and orthogonally mixing low-dimensional trajectories. 
The resulting corpus spans system dimensions from 2 to 100, exposing the model to both standard filtering regimes and broad dimensional variation. On in-domain evaluations, FilterFM delivers strong gains along three practical axes: it achieves the lowest average raw-space RMSE for state-estimation accuracy, runs substantially faster than all compared baselines, and remains consistently more robust than LLMFilter and robust online Bayesian filters under observation-noise misspecification. Beyond in-domain performance, FilterFM demonstrates \textbf{zero-shot filtering on unseen heterogeneous systems}. FilterFM exhibits foundation-model-style scaling behavior: broader system coverage at an approximately matched training-data budget, more trajectories per system at fixed system diversity, and greater model capacity each improve generalization to unseen dynamics.

Our contributions are summarized as follows:
\begin{itemize}
    \item We propose FilterFM, a pretrained model for Bayesian filtering built on a hierarchical two-level causal encoder and a prior-to-posterior state-reconstruction module. The model enables a single checkpoint to filter across heterogeneous dynamical systems with varying dimensions and, to our knowledge, is the first Bayesian filtering model to introduce quantile prediction, providing both point estimates and predictive uncertainty intervals.
    \item We curate a heterogeneous training corpus spanning diverse dynamics and dimensions, and establish a comprehensive evaluation showing FilterFM's advantages in state-estimation accuracy, runtime efficiency, and robustness to observation-noise misspecification, together with the first demonstration of zero-shot filtering on unseen heterogeneous systems.
    \item We investigate whether scaling pretraining can improve generalization to unseen dynamical systems. This is the first systematic study of scaling behavior in Bayesian filtering across training-system diversity, training-trajectory coverage, and model capacity. Our results provide evidence that expanding pretraining can improve a single model's ability to estimate states across heterogeneous dynamics.
\end{itemize}

\section{Related Work}
\label{sec:related}

\relatedparagraph{Online Bayesian filtering.}
Classical Bayesian filters such as Gaussian filters or particle filters~\citep{7528889,855552,11434934,8683411} provide the \emph{predict-then-update} paradigm for state estimation, which first predicts a prior distribution over the estimated state from the past observations and then updates it to a posterior via the transition model and observation map, but these methods require known transition models, observation maps, and noise statistics. In real-world applications, however, when explicit mathematical models are unavailable or noise statistics are misspecified, the feasibility of these approaches drops considerably.

\relatedparagraph{Learning-based filtering.}
Neural filters reduce the need for explicit mathematical models by learning estimators from the trajectories. KalmanNet learns a neural Kalman gain correction~\citep{KalmanNet} while other methods use different neural networks~\citep{MEStimator,RStateNet} to map observation sequences to state estimates. LLMFilter~\citep{LLMFilter} adopts a frozen LLM but suffers from patch-boundary inconsistencies across varying dimensions. Yet all these methods are system-specific and demand retraining for new dynamics, highlighting the need to broaden the generalization scope of Bayesian filtering toward reusable foundation-model-like frameworks.

\relatedparagraph{Foundation models.}
Time-series foundation models (TSFMs) such as Chronos-2~\citep{ansari2025chronos2univariateuniversalforecasting} and Moirai~2.0~\citep{liu2026moirai20timeseries} demonstrate strong zero-shot forecasting capabilities, while Panda~\citep{lai2026panda} and ChaosNexus~\citep{liu2026chaosnexusfoundationmodelodebased} target chaotic-system forecasting. In the area of state estimation, FM-UKF~\citep{FM-UKF} inserts a causal transformer into an unscented Kalman filter but still requires an explicit observation map and operates within a single parametric family. Critically, forecasting and filtering follow \emph{different information structures}: forecasting predicts a future window from an observed one, while online Bayesian filtering \emph{causally updates} the state from \emph{one new observation at a time}. FilterFM is designed for filtering: it learns the observation-to-state mapping from data, preserves strict per-step causality, and is the first pretrained model for Bayesian filtering across heterogeneous systems.

\section{Problem Formulation}
\label{sec:problem_formulation}

\begin{figure}[!t]
\centering
\includegraphics[width=\linewidth]{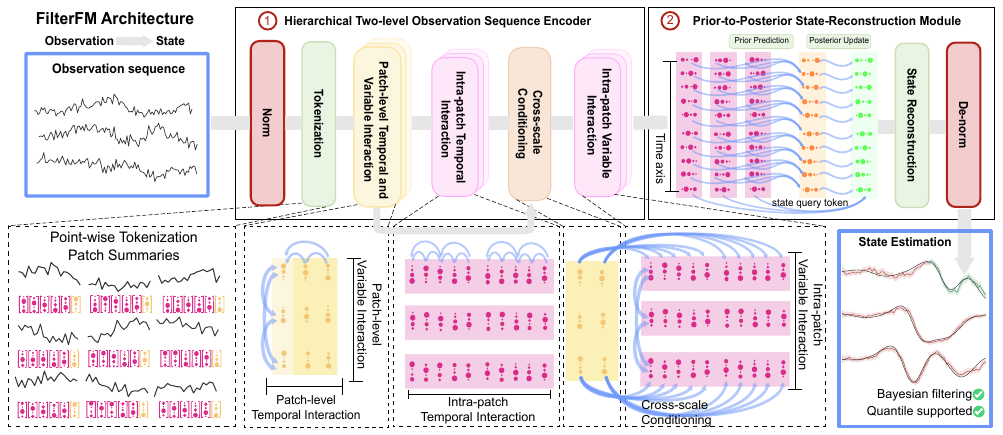}
\caption{FilterFM architecture. The encoder combines patch-level temporal and variable interaction with intra-patch temporal interaction, cross-scale conditioning, and intra-patch variable interaction. State queries read shifted and current local memory for prior prediction and posterior update, followed by quantile decoding and de-normalization to the original state scale.}
\label{fig:framework}
\end{figure}

Let \(\mathbf{x}_t\in\mathbb{R}^{N}\) denote the latent state of a dynamical system and \(\mathbf{y}_t\in\mathbb{R}^{M}\) denote the noisy observation at time \(t\). A Markovian state-space model~\citep{WEISS1965141} can be written as
\begin{equation}
\begin{aligned}
\mathbf{x}_{t+1} &= f(\mathbf{x}_t) + \bm{\omega}_t,\\
\mathbf{y}_t &= h(\mathbf{x}_t) + \bm{\epsilon}_t,
\end{aligned}
\label{eq:state_space}
\end{equation}
where \(f\) is the transition function, \(h\) is the observation function, and \(\bm{\omega}_t\sim\mathcal{N}(0,\bm{R})\), \(\bm{\epsilon}_t\sim\mathcal{N}(0,\bm{Q})\) denote process and observation noise sampled independently from Gaussian distributions. The filtering objective is to estimate the current state using only observations available up to the current time. Bayesian filtering expresses this objective through a prediction step and an update step:
\begin{equation}
\begin{aligned}
p(\mathbf{x}_t|\mathbf{y}_{1:t-1})
&=
\int p(\mathbf{x}_t|\mathbf{x}_{t-1})
p(\mathbf{x}_{t-1}|\mathbf{y}_{1:t-1})\,d\mathbf{x}_{t-1},\\
p(\mathbf{x}_t|\mathbf{y}_{1:t})
&=
\frac{
p(\mathbf{y}_t|\mathbf{x}_t)
p(\mathbf{x}_t|\mathbf{y}_{1:t-1})
}{
\int
p(\mathbf{y}_t|\mathbf{x}_t)
p(\mathbf{x}_t|\mathbf{y}_{1:t-1})\,d\mathbf{x}_t
}.
\end{aligned}
\label{eq:bayes_filtering}
\end{equation}
Here, \(p(\mathbf{x}_t\mid\mathbf{y}_{1:t-1})\) is the prior distribution given past observations, and \(p(\mathbf{x}_t\mid\mathbf{y}_{1:t})\) is the posterior distribution after incorporating \(\mathbf{y}_t\). This prior-to-posterior update motivates our state-reconstruction module. A point estimate is the posterior mean \(\hat{\mathbf{x}}_t=\mathbb{E}[\mathbf{x}_t|\mathbf{y}_{1:t}]\). Classical filters instantiate this recursion by specifying \(f\), \(h\), and the noise levels for each system. Learning-based filters instead replace the explicit Bayesian recursion with a neural estimator. We focus on the \emph{foundation-model version} of this problem, using one neural model across different dynamical systems. For each system, the model receives \(\mathbf{y}_{1:t}\in\mathbb{R}^{t\times M}\) and outputs \(\hat{\mathbf{x}}_t\in\mathbb{R}^{N}\) according to the mapping \(\hat{\mathbf{x}}_t=F_\theta(\mathbf{y}_{1:t})\).

\section{Method}
\label{sec:method}

In this section, we introduce the architecture of FilterFM, which comprises a hierarchical two-level causal observation sequence encoder and a prior-to-posterior state-reconstruction module, as shown in Figure~\ref{fig:framework}. We first describe how the encoder represents observation history and current evidence, then explain how these representations support prior prediction and posterior update.

\subsection{Hierarchical Two-Level Observation Sequence Encoder}
\label{sec:method_encoder}

Our encoder is designed to preserve current observations for filtering while efficiently summarizing longer histories. We therefore represent observations at two temporal levels: patch-level embeddings summarize consecutive groups of observations, while intra-patch-level embeddings retain individual timesteps within each group. Causal temporal and prototype-based variable interaction operate at both levels, with cross-scale conditioning providing historical context to intra-patch embeddings.

\noindent\textbf{Observation Tokenization and Patch Summaries.}\label{sec:method_norm_token} FilterFM first applies per-channel reversible instance normalization~\citep{kim2022revin} to both observation and state sequences: \(\tilde{y}_{t,m}=(y_{t,m}-\mu^y_m)/\sigma^y_m\) and \(\tilde{x}_{t,n}=(x_{t,n}-\mu^x_n)/\sigma^x_n\), where \(\mu^y_m,\sigma^y_m\) and \(\mu^x_n,\sigma^x_n\) are the trajectory-wise mean and standard deviation of observation channel \(m\) and state channel \(n\), respectively. We partition observations into patches of length \(L_p\), with \(t=pL_p+r\), \(p\ge0\), and \(r=1,\ldots,L_p\). A shared scalar encoder embeds normalized observations \(\tilde{y}_{t,m}\), with channel and position embeddings added to form tokens \(\mathbf{o}_{p,r,m}\in\mathbb{R}^{d}\). Causal temporal encoding and learned pooling with skip projections yield patch summaries \(\mathbf{s}_{p,m}\in\mathbb{R}^{d}\), as detailed in Appendix~\ref{app:method_patch_summaries}.

\noindent\textbf{Temporal and Variable Interaction.}\label{sec:method_temporal_attention} To illustrate both levels, we describe one block and omit the layer indices distinguishing successive blocks for brevity. Let \(\ell\in\{\mathrm{patch},\mathrm{intra}\}\) denote the level, with \(s=p\) or \(s=(p,r)\), respectively. Its input and temporal output are \(\mathbf{U}^\ell_s,\bar{\mathbf{U}}^\ell_s\in\mathbb{R}^{M\times d}\). The first input stacks \(\mathbf{s}_{p,m}^{\top}\) or \(\mathbf{o}_{p,r,m}^{\top}\) as rows, while later inputs are the preceding block's full output \(\mathbf{U}^{\ell,{\rm out}}_s\) in Eq.~(\ref{eq:method_reassemble}).

\emph{Causal temporal interaction.} Causally masked self-attention models temporal dependencies independently for each variable, followed by a residual feed-forward network (FFN):
\begin{equation}
\widetilde{\mathbf{U}}^\ell_s = \mathbf{U}^\ell_s + {\rm SelfAttn}_\ell(\mathbf{U}^\ell_{\preceq_\ell s}),
\bar{\mathbf{U}}^\ell_s = \widetilde{\mathbf{U}}^\ell_s + {\rm FFN}_\ell(\widetilde{\mathbf{U}}^\ell_s),
\label{eq:method_temporal_block}
\end{equation}
where \(\preceq_\ell\) includes patches up to \(p\) at the patch level and positions up to \(r\) within the same patch at the intra-patch level. Layer normalization, dropout, and projection biases are omitted for clarity.

\emph{Prototype-based variable interaction.} To exchange information across different numbers of variables, we route the \(M\) tokens through a fixed number \(C\) of prototype slots~\citep{liu2026falconxtimeseriesfoundation}. Their input \(\mathbf{V}^\ell_s\in\mathbb{R}^{M\times d}\) equals \(\bar{\mathbf{U}}^\ell_s\) at the patch level and incorporates historical context at the intra-patch level through the cross-scale conditioning described below. The \(c\)-th rows \(\mathbf{k}^{\ell,\pm}_c\in\mathbb{R}^{d}\) of the learnable matrices \(\mathbf{K}^\ell_\pm\in\mathbb{R}^{C\times d}\) serve as the positive and negative attention keys for slot \(c\), while projections \(\mathbf{A}^\ell_q,\mathbf{A}^\ell_v\in\mathbb{R}^{d\times d}\) map each token \(\mathbf{v}^\ell_{s,m}\) to query and value vectors:
\begin{equation}
\alpha^{\ell,\pm}_{s,m,c}
=\operatorname{softmax}_{c}\!\left(\frac{(\mathbf{A}^\ell_q\mathbf{v}^\ell_{s,m})^\top\mathbf{k}^{\ell,\pm}_c}{\sqrt{d}}\right),\mathbf{p}^\ell_{s,c}
=\frac{1}{M}\sum_{m'=1}^{M}
\big(\alpha^{\ell,+}_{s,m',c}-\lambda_\ell\alpha^{\ell,-}_{s,m',c}\big)\mathbf{A}^\ell_v\mathbf{v}^\ell_{s,m'}.
\label{eq:method_diffattn}
\end{equation}
The positive and negative attention branches normalize independently over the \(C\) slots for each variable, and \(\lambda_\ell\) controls the contribution of negative attention. After obtaining the prototypes \(\mathbf{p}^\ell_{s,c}\), we stack them into \(\mathbf{P}^\ell_s\in\mathbb{R}^{C\times d}\) and apply self-attention followed by a residual FFN to obtain \(\tilde{\mathbf{P}}^\ell_s\), then use this memory to update variable tokens through the gated reassembly router in Eq.~(\ref{eq:method_reassemble}):
\begin{equation}
\begin{aligned}
\mathbf{R}^\ell_s&={\rm CrossAttn}(\mathbf{V}^\ell_s,\tilde{\mathbf{P}}^\ell_s,\tilde{\mathbf{P}}^\ell_s),\,
\mathbf{G}^\ell_s=\sigma({\rm MLP}_\ell([\mathbf{V}^\ell_s;\mathbf{R}^\ell_s])),\\
\widehat{\mathbf{V}}^\ell_s&=\mathbf{V}^\ell_s+\mathbf{G}^\ell_s\odot\big(\mathbf{R}^\ell_s(\mathbf{W}^{\ell,\mathrm{proto}}_o)^\top\big),\,
\mathbf{U}^{\ell,{\rm out}}_s=\widehat{\mathbf{V}}^\ell_s+{\rm FFN}_\ell(\widehat{\mathbf{V}}^\ell_s).
\end{aligned}
\label{eq:method_reassemble}
\end{equation}
Here \(\mathbf{W}^{\ell,\mathrm{proto}}_o\in\mathbb{R}^{d\times d}\) is a learnable output projection and \(\widehat{\mathbf{V}}^\ell_s\) contains the updated variable representations, with \(\mathbf{R}^\ell_s,\widehat{\mathbf{V}}^\ell_s,\mathbf{U}^{\ell,{\rm out}}_s\in\mathbb{R}^{M\times d}\) and \(\mathbf{G}^\ell_s\in(0,1)^{M\times d}\). The operators \(\sigma\), \(\odot\), and \([\cdot;\cdot]\) denote sigmoid, elementwise multiplication, and feature concatenation, respectively.

\noindent\textbf{Cross-Scale Conditioning.}\label{sec:method_patch_memory} To combine local evidence with longer-term context, intra-patch embeddings read completed patch summaries before the variable interaction described above. Stacked patch-level blocks yield summaries \(\mathbf{H}^{\rm raw}_{p,m}\), delayed by one patch as \(\mathbf{H}_{0,m}=\mathbf{h}^{\rm bos}_m\) and \(\mathbf{H}_{p,m}=\mathbf{H}^{\rm raw}_{p-1,m}\) for \(p\ge1\), with learned initial memory \(\mathbf{h}^{\rm bos}_m\). We define this history read through \({\rm Retrieve}\), a gated cross-attention operator that updates a query \(\mathbf{q}\in\mathbb{R}^{d}\) from a memory matrix \(\mathbf{B}\in\mathbb{R}^{M\times d}\) containing one representation for each observation channel:
\begin{equation}
\mathbf{r}={\rm CrossAttn}(\mathbf{q},\mathbf{B},\mathbf{B}),\,
\tilde{\mathbf{q}}=\mathbf{q}+\mathbf{g}\odot\mathbf{W}^{\mathrm{ret}}_o\mathbf{r},\,
{\rm Retrieve}(\mathbf{q},\mathbf{B})=\tilde{\mathbf{q}}+{\rm FFN}(\tilde{\mathbf{q}}),
\label{eq:method_retrieve}
\end{equation}
where \(\mathbf{r},\tilde{\mathbf{q}}\in\mathbb{R}^{d}\) are the retrieved memory vector and updated query, respectively, and \(\mathbf{g}=\sigma({\rm MLP}([\mathbf{q};\mathbf{r}]))\in(0,1)^d\). The temporal output then queries delayed patch memory to produce history-conditioned embeddings for intra-patch variable interaction:
\begin{equation}
\mathbf{v}^{\rm intra}_{p,r,m}={\rm Retrieve}\!\left(\bar{\mathbf{u}}^{\rm intra}_{p,r,m},\mathbf{H}_{p,1:M}\right).
\label{eq:method_topdown_read}
\end{equation}
Within each intra-patch block, the history-conditioned embeddings \(\mathbf{V}^{\rm intra}_{p,r}\) undergo prototype-based variable interaction through Eqs.~(\ref{eq:method_diffattn})--(\ref{eq:method_reassemble}). The final block produces local memory \(\mathbf{L}_{t,m}\), whose one-step shift \(\mathbf{L}^-_{t,m}=\mathbf{L}_{t-1,m}\) provides past evidence for prior prediction with \(\mathbf{L}_{0,m}=\mathbf{0}\) at initialization.
\subsection{Prior-to-Posterior State-Reconstruction Module}
\label{sec:method_belief}
The state-reconstruction module estimates latent states from the encoder's local memory through a prior-to-posterior update. For each state channel \(n=1,\ldots,N\), a learned query \(\mathbf{q}_{t,n}\in\mathbb{R}^{d}\) combines channel identity, patch index, and intra-patch position to read the \(M\) observation tokens, allowing \(N\) to differ from \(M\). Prior prediction and posterior update use the \({\rm Retrieve}\) structure in Eq.~(\ref{eq:method_retrieve}) with independently learned parameters to access shifted and current local memory, respectively.

\noindent\textbf{Prior Prediction.} The state query forms a prior token \(\mathbf{z}^-_{t,n}\in\mathbb{R}^{d}\) from shifted local memory, using the encoded history before the current observation:
\begin{equation}
\mathbf{z}^-_{t,n} = {\rm Retrieve}\big(\mathbf{q}_{t,n},\;\mathbf{L}^-_{t,1:M}\big).
\label{eq:method_prior}
\end{equation}

\begin{table*}[!t]
\centering
\begingroup
\fontfamily{qtm}\selectfont
\Large\bfseries\boldmath
\setlength{\heavyrulewidth}{1.6pt}
\renewcommand{\arraystretch}{1.80}
\setlength{\tabcolsep}{1.5pt}
\resizebox{\textwidth}{!}{%
\begin{tabular}{@{}lrrrrrrrrrrrrr@{}}
\toprule
{\LARGE\textbf{Method}} &
\multicolumn{1}{c}{{\LARGE\textbf{Ours}}} &
\multicolumn{7}{c}{{\LARGE\textbf{Learning-Based Filters}}} &
\multicolumn{5}{c}{{\LARGE\textbf{Online Bayesian Filters}}} \\
\cmidrule(lr){3-9}\cmidrule(lr){10-14}
&
\textbf{FilterFM} &
\textbf{LLMFilter} &
\textbf{FM-UKF} &
\textbf{MEstimator} &
\textbf{RStateNet} &
\textbf{ProTran} &
\textbf{KalmanNet} &
\textbf{RNNYYF} &
\textbf{EnKF} &
\textbf{PF} &
\textbf{EnKFI} &
\textbf{EnKFS} &
\textbf{HubEnKF} \\
\midrule
Tracking & \best{0.9683} & 1.9944 & \runnerup{1.2556} & 2.7357 & 17.8542 & \third{1.5862} & 17.7382 & 18.7570 & 1.7846 & 1.7820 & 1.7858 & 1.7874 & 1.7853 \\
Pendulum & \runnerup{0.0840} & \third{0.0891} & 0.1318 & 0.2234 & 0.2243 & 0.1095 & 0.1457 & 2.1915 & \best{0.0822} & 0.3266 & 0.2776 & 0.0933 & 0.0897 \\
Selkov & \best{0.0155} & 0.0189 & \third{0.0165} & 0.0264 & 0.0336 & 0.0182 & 0.1059 & 0.2291 & \runnerup{0.0165} & 0.0167 & 0.1384 & 0.0217 & 0.0219 \\
Oscillator & \best{0.0166} & \runnerup{0.0166} & 0.0171 & 0.0249 & 0.0320 & 0.0181 & 2.9264 & 0.1307 & \third{0.0170} & 0.0172 & 0.1313 & 0.0206 & 0.0212 \\
Hopf & \best{0.0156} & 0.0176 & \third{0.0167} & 0.0280 & 0.0304 & 0.0176 & 0.0540 & 0.1991 & \runnerup{0.0164} & 0.0169 & 0.2095 & 0.0212 & 0.0216 \\
\midrule
Lorenz96 & \best{1.1697} & \third{2.0539} & \runnerup{1.3314} & 2.6634 & 3.3996 & 2.5728 & -- & 5.1669 & 5.3113 & 6.2552 & 5.3311 & 5.3875 & 5.3453 \\
VL20 & \best{0.5315} & 2.4686 & \runnerup{0.6877} & 2.2144 & 3.3186 & 2.1721 & -- & 5.1090 & 1.4819 & 6.3915 & 5.9505 & 1.5404 & \third{1.1619} \\
LorenzUV40 & \best{0.1219} & 0.2543 & \runnerup{0.2487} & \third{0.2525} & 0.9310 & 0.2615 & -- & 0.6085 & 0.3213 & 2.7444 & 0.3215 & 0.3142 & 0.3128 \\
Vanderpol64 & \best{0.0683} & 0.3413 & 0.1422 & 0.2665 & 0.2236 & 0.2968 & -- & 0.7162 & 0.1352 & 1.8032 & \third{0.1346} & 0.1361 & \runnerup{0.1334} \\
Rossler96 & \best{0.0584} & 0.2459 & \runnerup{0.1560} & 0.2715 & \third{0.1604} & 0.2422 & -- & 0.5554 & 0.1817 & 0.5458 & 0.1813 & 0.1880 & 0.1827 \\
Duffing64 & \best{0.0224} & 0.0635 & \runnerup{0.0353} & 0.0908 & 0.1157 & 0.0714 & -- & 0.2743 & \third{0.0446} & 0.8295 & 0.0489 & 0.0474 & 0.0454 \\
Burgers64 & \best{0.0203} & 0.0728 & \runnerup{0.0294} & 0.0690 & 0.1285 & 0.0765 & -- & 0.3352 & \third{0.0308} & 83.4534 & 2.8609 & 0.0410 & 0.0366 \\
Fhn64 & \best{0.0185} & 0.0717 & \runnerup{0.0322} & 0.0737 & 0.1003 & 0.0765 & -- & 0.3272 & \third{0.0341} & 0.6590 & 0.0448 & 0.0366 & 0.0349 \\
Cubic-Sensoring100 & \best{0.1405} & 0.7277 & \third{0.1811} & 0.6646 & 0.6316 & 0.7433 & -- & 0.9869 & \runnerup{0.1497} & 0.8148 & 0.9696 & 0.3281 & 0.3437 \\
\midrule
\rowcolor[HTML]{EEF4EB}
\textbf{Average} & \best{\textbf{0.2322}} & \textbf{0.6026} & \runnerup{\textbf{0.3058}} & \textbf{0.6861} & \textbf{1.9417} & \third{\textbf{0.5902}} & \textbf{4.1940} & \textbf{2.5419} & \textbf{0.6862} & \textbf{7.5469} & \textbf{1.3133} & \textbf{0.7117} & \textbf{0.6812} \\
\midrule
\rowcolor[HTML]{F3EDF7}
\textbf{1$^{st}$ Count} & \best{\textbf{13}} & \textbf{0} & \textbf{0} & \textbf{0} & \textbf{0} & \textbf{0} & \textbf{0} & \textbf{0} & \runnerup{\textbf{1}} & \textbf{0} & \textbf{0} & \textbf{0} & \textbf{0} \\
\bottomrule
\end{tabular}
}
\endgroup
\caption{In-domain raw-space RMSE on 14 dynamical systems (lower is better). Red, blue underlining, and teal italics indicate the top three results; \(1^{st}\) Count denotes row-wise wins. Rankings use unrounded values. Missing entries (--) are excluded from the corresponding method's average.}
\label{tab:panda_foundation_selected14_online}
\end{table*}

\begin{figure}[!t]
\centering
\setlength{\parskip}{0pt}
\setlength{\abovecaptionskip}{3pt}
\setlength{\belowcaptionskip}{-6pt}
\makebox[\linewidth][c]{\includegraphics[width=1.04\linewidth,height=0.425\linewidth,pagebox=artbox]{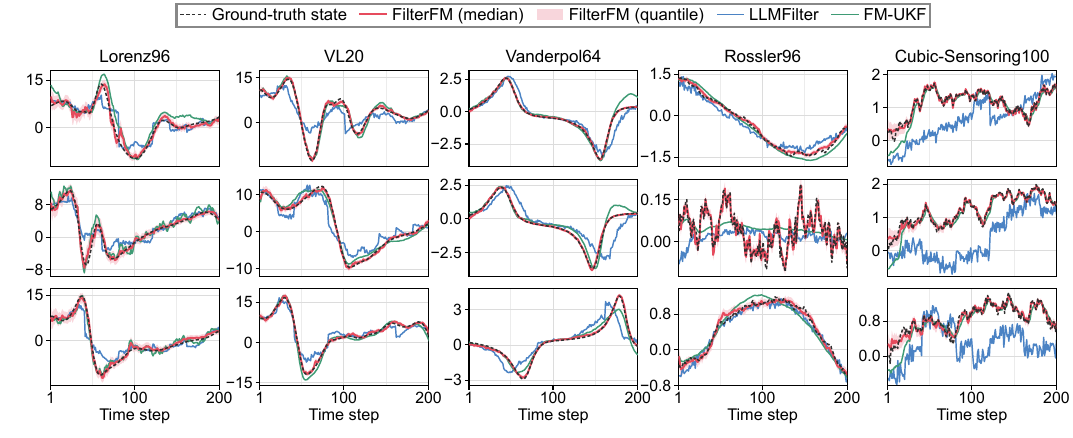}}\par\nointerlineskip
\caption{State-estimation trajectories on five high-dimensional in-domain systems (columns), with three state channels per system (rows). FilterFM tracks rapid changes in Lorenz96 and VL20, turning points in Vanderpol64, and small fluctuations in Rossler96. LLMFilter shows jumps and large offsets on Cubic-Sensoring100, while FM-UKF exhibits timing and amplitude errors. Light-red shading shows FilterFM's predicted $0.1$--$0.9$ quantile interval.}
\label{fig:trajectory_baseline}
\end{figure}

\begin{samepage}
\noindent\textbf{Posterior Update.} To incorporate the current observation, the prior token queries current local memory for evidence \(\mathbf{e}_{t,n}\in\mathbb{R}^{d}\) to compute a gated innovation correction:
\begin{equation}
\mathbf{e}_{t,n}={\rm Retrieve}(\mathbf{z}^-_{t,n},\mathbf{L}_{t,1:M}),\,
\mathbf{z}^+_{t,n}=\mathbf{z}^-_{t,n}+\mathbf{g}_{t,n}\odot\boldsymbol{\Delta}_{t,n}.
\label{eq:method_posterior_innovation}
\end{equation}
\end{samepage}
Two MLP branches take the concatenated prior, evidence, and their difference as input, producing the sigmoid gate \(\mathbf{g}_{t,n}\in(0,1)^d\) and correction \(\boldsymbol{\Delta}_{t,n}\in\mathbb{R}^{d}\), respectively.

\noindent\textbf{State Reconstruction.} A quantile head outputs \(Q^+_{t,n,\tau}=W^+_\tau\mathbf{z}^+_{t,n}\) for \(\tau\in\mathcal{Q}=\{0.1,0.5,0.9\}\), with the median mapped back to the state scale as \(\hat{x}_{t,n}=\sigma^x_n Q^+_{t,n,0.5}+\mu^x_n\) and the outer quantiles describing uncertainty. Training supervises both prior and posterior with quantile losses, supplemented by a median Huber loss and prototype-key regularization, as detailed in Appendix~\ref{app:method_training_loss}.

\section{Experiment}
\textbf{Benchmarks.}\label{sec:exp_datasets} We curate 14 established in-domain filtering tasks spanning low-dimensional nonlinear systems such as Selkov~\citep{selkov} and Oscillator~\citep{SINDY}, and high-dimensional chaotic systems such as Lorenz96~\citep{lorenz96}, VL20~\citep{vissio2020mechanics}, and Cubic-Sensoring100~\citep{hazewinkel1983cubic}. We augment them with auxiliary systems from skew-product chaotic trajectories in~\cite{lai2026panda}, using clean latent-state windows and per-channel Gaussian observation noise with a log-uniform scale. Because these source trajectories have at most 10 dimensions, we also form systems up to dimension 100 by normalizing, concatenating, and orthogonally mixing low-dimensional windows. In-domain evaluation uses 20 held-out trajectories per system, while OOS evaluation uses held-out systems with the same noise protocol. Full system definitions and generation details are in Appendix~\ref{app:dataset_details}.

\textbf{Baselines and Metrics.}\label{sec:exp_baselines} We compare FilterFM with four baseline groups. Learning-based filters include MEstimator~\citep{MEStimator}, RStateNet~\citep{RStateNet}, ProTran~\citep{ProTran}, KalmanNet~\citep{KalmanNet}, and RNNYYF~\citep{RNNYYF}, each trained separately for each system. LLMFilter~\citep{LLMFilter} uses the System-as-Prompt setting, whereas FM-UKF~\citep{FM-UKF} couples a time-series foundation model with an unscented Kalman filter~\citep{882463}. Online Bayesian baselines comprise EnKF, its robust variants EnKFI, EnKFS, and HubEnKF~\citep{Evensen1994EnKF}, together with a particle filter~\citep{8683411}. We measure state-estimation accuracy by root mean square error (RMSE), defined in Appendix~\ref{app:experiment_details}, and runtime by wall-clock latency per filtering step (ms/step), excluding model loading.

\begin{figure}[!t]
\centering
\includegraphics[width=0.95\linewidth]{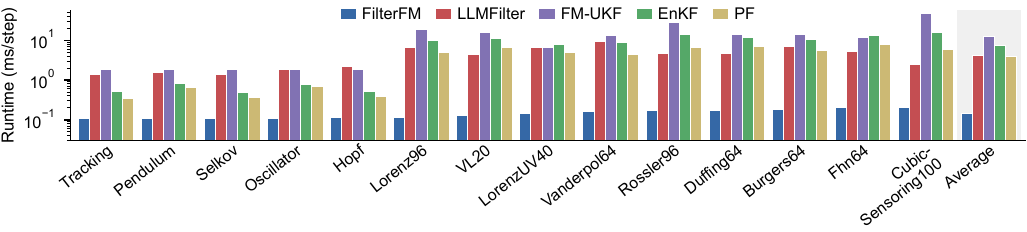}
\vspace{-0.7em}
\caption{Per-step inference latency on 14 benchmark systems. The y-axis reports latency in ms/step on a base-10 logarithmic scale. Measurements use a single NVIDIA A800 GPU and exclude model loading. FilterFM is fastest on every system, averaging 0.145~ms/step.}
\label{fig:runtime_selected14}
\end{figure}

\begin{figure}[!t]
\centering
\includegraphics[width=0.98\linewidth]{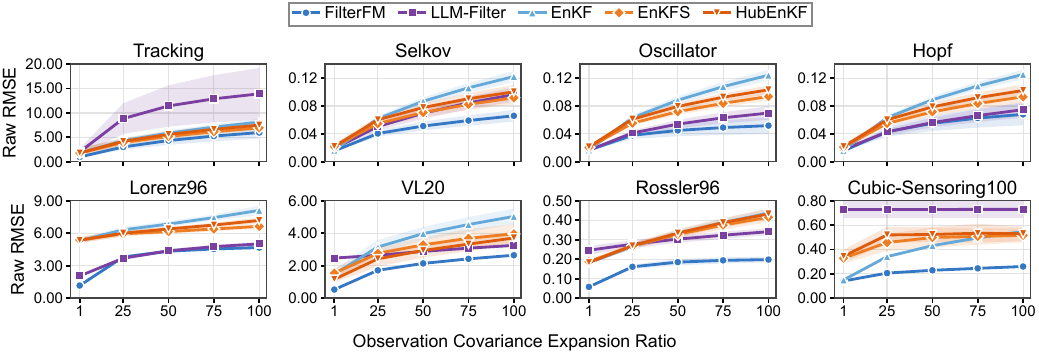}
\vspace{-0.7em}
\caption{Observation-noise misspecification: true covariance is scaled up while online filters retain their original assumed covariance. FilterFM achieves the lowest pooled RMSE in 38 of the 40 displayed settings across low- and high-dimensional systems. Curves show pooled RMSE; shading denotes the population standard deviation of trajectory-wise RMSE over 20 test trajectories.}
\label{fig:model_mismatch}
\end{figure}

\begin{figure}[!t]
\centering
\includegraphics[width=0.96\linewidth]{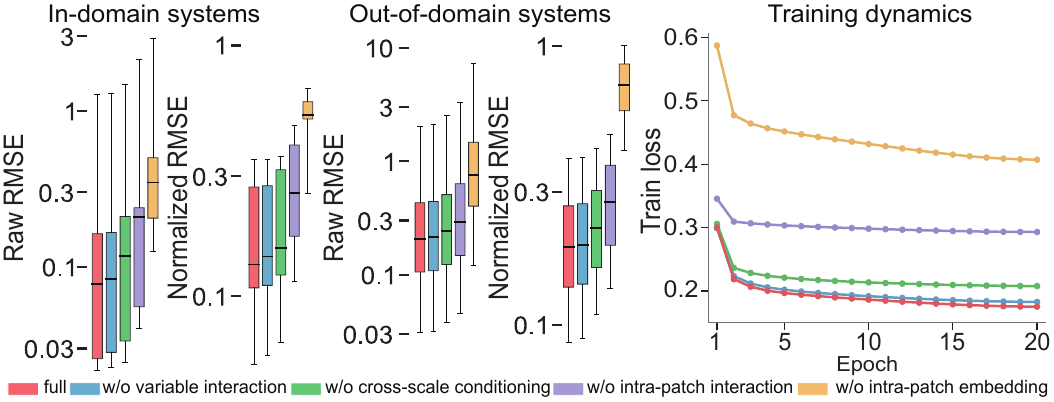}
\caption{Ablation study of five architectural variants. Left and middle show raw and normalized RMSE distributions on in-domain and out-of-domain systems, respectively; boxes mark interquartile ranges and medians, with 5th--95th percentile whiskers. Right shows training-loss curves under identical data, optimization settings, and 20-epoch budgets. The full model achieves the lowest median errors on both splits and the lowest final training loss.}
\label{fig:ablation_dynamics}
\end{figure}

\subsection{In-Domain Performance in Terms of Accuracy, Runtime, and Robustness}
\label{sec:exp_indomain}

FilterFM is pretrained on approximately 15K trajectories from the heterogeneous corpus described above, including 480 training trajectories from each in-domain system. The system-specific learning-based filters use the same 480 target-system trajectories for training. We adapt KalmanNet from its official implementation. For LLMFilter, we adapt the official implementation and train its input and output adapters on the same per-system data while keeping the LLaMA-7B~\citep{touvron2023llama} backbone frozen, with the task prompts provided in Appendix~\ref{app:llmfilter_prompts}. For FM-UKF, we train a separate causal Transformer on pooled training trajectories from systems with the same state dimension. Online Bayesian filters use each system's ground-truth observation model. Appendix~\ref{app:baseline_details} provides implementation and training details.

\noindent\textbf{Accuracy Analysis.} Table~\ref{tab:panda_foundation_selected14_online} summarizes quantitative performance across all 14 systems. A single FilterFM checkpoint handles state dimensions from 2 to 100, wins 13 of 14 systems, and achieves the lowest average raw-space RMSE. Training on heterogeneous systems does not degrade per-benchmark accuracy, but rather \emph{improves} it: FilterFM surpasses every per-system learning-based baseline, with the advantage most pronounced in high dimensions. Figure~\ref{fig:trajectory_baseline} visualizes representative trajectories comparing FilterFM with LLMFilter and FM-UKF on five high-dimensional systems, with three state channels shown per system. LLMFilter's autoregressive patch-wise decoding introduces \textbf{local discontinuities} and its flattening of time and variable channels loses cross-variable structure at scale. FM-UKF relies on an explicit observation model within a UKF; when model mismatch occurs, it incurs \textbf{systematic time lags and amplitude deviations}. In contrast, FilterFM closely tracks the ground truth while learning the observation-to-state mapping entirely from data without assuming an observation model. Appendix~\ref{app:indomain_trajectory_showcase} shows trajectories for all 14 systems.

\noindent\textbf{Runtime.} All runtime measurements are conducted on a single NVIDIA A800 GPU under the same hardware setting. Figure~\ref{fig:runtime_selected14} compares per-step inference time. FilterFM runs at \(0.145\) ms/step on average, substantially faster than all compared baselines. LLMFilter's inference cost is \textbf{dominated by system-as-prompt tokens}, since patching contributes fewer than 20 tokens to the backbone. FM-UKF is slower because each step requires both \textbf{neural prediction and explicit UKF posterior correction}, incurring the computational cost of both stages. In contrast, FilterFM produces state estimates \textbf{end-to-end in a single forward pass}, eliminating prompt overhead and symbolic filtering.

\noindent\textbf{Model Mismatch.} We further evaluate the model's robustness under observation-noise misspecification~\citep{LLMFilter,Evensen1994EnKF} by scaling the true observation covariance on test trajectories while keeping the assumed covariance fixed. Pooled RMSE is computed by averaging squared errors over all test trajectories, time steps, and valid state channels within each system, and then taking the square root. Figure~\ref{fig:model_mismatch} shows that FilterFM achieves lower pooled RMSE than LLMFilter in most tested settings and outperforms EnKF and its robust variants, EnKFS and HubEnKF, across all reported observation covariance expansion ratios (OCER \(\in\{1,25,50,75,100\}\)). Appendix~\ref{app:model_mismatch_details} reports per-setting pooled RMSE and the standard deviation of trajectory-wise RMSE over 20 test trajectories.

\subsection{Zero-Shot Generalization and Ablation Analysis}
\label{sec:exp_zeroshot}

FilterFM supports zero-shot filtering across heterogeneous dynamical systems. On held-out Panda systems~\citep{lai2026panda} with three to six state variables, the same checkpoint transfers without fine-tuning, as illustrated in Figure~\ref{fig:ablation_trajectories} and the additional cases in Appendix~\ref{app:oos_trajectory_showcase}.

\begin{figure}[!t]
\centering
\includegraphics[width=\linewidth]{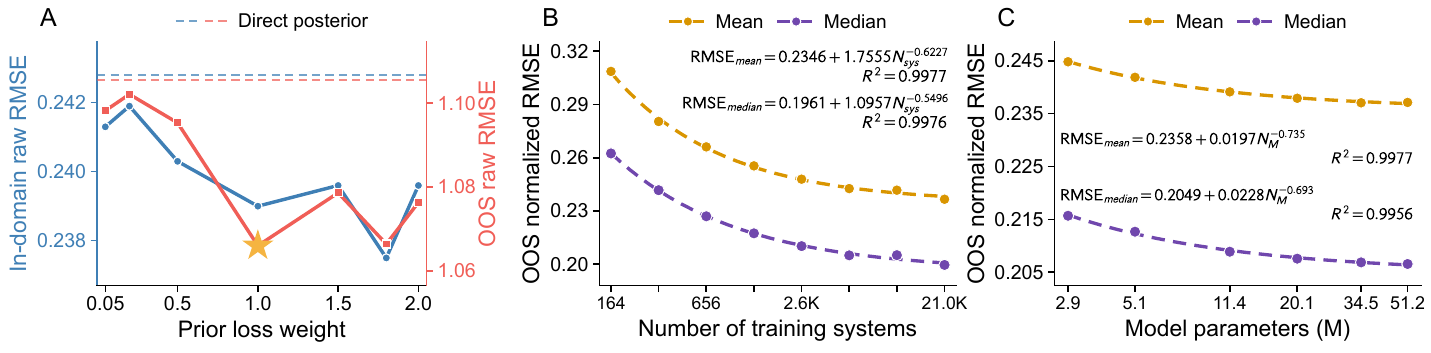}
\caption{Prior-loss sweep and scaling behavior. (A)~Raw RMSE on nine high-dimensional in-domain systems (blue; 40--100 state variables) and OOS systems (red). Dashed lines: direct posterior reconstruction; star: best OOS setting ($\lambda_{\rm prior}=1.0$). Full in-domain results can be found in Appendix~\ref{app:prior_loss_weight}. (B,C)~Mean and median OOS normalized RMSE versus training-system count and model capacity, with measurements (dots), power-law fits (dashed curves), and logarithmic x-axes.}
\label{fig:scaling}
\end{figure}

The ablations in Figure~\ref{fig:ablation_dynamics} identify current-observation access and intra-patch interaction as the main contributors to accuracy. \textbf{W/o intra-patch embedding} relies only on past patch representations and has the largest median RMSE increase on both splits, missing rapid changes in Figure~\ref{fig:ablation_trajectories}. \textbf{W/o intra-patch interaction} retains current observations but omits their local processing, causing the second-largest increase. Removing \textbf{cross-scale conditioning} or \textbf{variable interaction} has smaller effects, with history conditioning contributing more. Under identical training settings and a 20-epoch budget, the full model also reaches the lowest training loss and the history-only variant the highest. Appendix~\ref{app:architectural_ablations} provides further analysis of these ablations. In Figure~\ref{fig:scaling}A, explicit prior-to-posterior updates improve OOS performance over direct posterior reconstruction; we adopt a prior-loss weight of 1.0, which yields the lowest OOS RMSE.

\subsection{Scaling Behavior}
\label{sec:exp_scaling}

We study how training-system diversity, training-trajectory coverage, and model capacity affect Bayesian filtering on unseen dynamics. Improvements along all three axes suggest that a shared model can learn more general state-estimation capabilities from larger-scale pretraining.

\textbf{Scaling system diversity.} Increasing the nominal system count from 164 to 20,980 lowers mean OOS normalized RMSE from $0.3086$ to $0.2366$ and the median from $0.2624$ to $0.1996$ in Figure~\ref{fig:scaling}B. With approximately matched total training-trajectory counts, greater dynamical diversity improves OOS generalization, with diminishing gains as the number of training systems increases.

\textbf{Scaling model parameters.} We train six FilterFM models, ranging from $2.9$M to $51.2$M parameters, on the same expanded corpus of 301,873 long trajectories with a fixed number of training epochs. Table~\ref{tab:model_parameter_scaling_architecture} details the changes in hidden widths, self-attention head counts, and encoder and state-reconstruction depths across model sizes. Mean OOS normalized RMSE decreases from $0.2448$ to $0.2371$ and the median from $0.2157$ to $0.2066$ in Figure~\ref{fig:scaling}C, indicating that greater capacity improves generalization with the training data held fixed.

\textbf{Scaling training trajectories.} With model capacity and training systems fixed, increasing the corpus from 20,471 to 301,873 trajectories lowers mean and median OOS normalized RMSE by $7.16\%$ and $9.17\%$, respectively, averaged over three seeds. Both statistics decrease across all six dataset sizes, showing that broader coverage of initial conditions benefits generalization without changing the underlying training systems. Appendix~\ref{app:trajectory_scaling} details the experimental settings and results.

\section{Conclusion and Future Work}

We presented FilterFM, the first pretrained model for Bayesian filtering across heterogeneous dynamical systems with a single checkpoint. FilterFM combines a hierarchical two-level causal encoder with a prior-to-posterior state-reconstruction module. It delivers superior in-domain performance in state-estimation accuracy, runtime efficiency, and robustness to observation-noise misspecification, alongside zero-shot generalization to unseen dynamics. Ablation and scaling studies support the architectural design and the benefits of greater system diversity, trajectory coverage, and model capacity. Future work should expand training data to cover broader observation models, sensor modalities, and noise structures, advancing general-purpose filtering models.

\subsection*{AI use statement}

In this work, we used LLM-based assistants to polish the wording of text. We have not used generative AI tools to produce research ideas or experimental designs. No generative AI tool had access to non-public data. We have reviewed all AI-assisted work: all AI-assisted text was checked, corrected, and approved by the authors, and all AI-assisted code was verified for correctness, with every reported result produced by the authors' own runs. We take responsibility for the final content of this work, including text, claims, or artifacts produced with the aid of generative AI.

\subsection*{Ethics statement}

This work trains and evaluates FilterFM using synthetic trajectories and publicly available academic datasets used under their respective licenses; no private or personally collected data is involved, and no human-subject study was conducted. Our filtering datasets combine simulations of specified dynamical systems with transformations of publicly available Panda data~\citep{lai2026panda}, and we will document their provenance, construction procedures, and splits upon publication so that upstream restrictions remain traceable. The proposed method improves state estimation across heterogeneous dynamical systems and, to our knowledge, does not introduce risks beyond those already associated with Bayesian filtering, such as privacy concerns in tracking applications. As with any advance in state estimation, improved tracking capabilities could in principle be misused in surveillance settings; our training procedure neither targets nor specifically facilitates such uses.

\subsection*{Reproducibility statement}

We document everything needed to reproduce our results in the paper and its appendix. The construction pipeline of our filtering datasets, including benchmark dynamics, source datasets, observation generation, and train--test splits, is described in Appendices~\ref{app:selected14_systems}--\ref{app:training_corpus}; the protocols behind the runtime, robustness, ablation, and scaling experiments are specified in Appendices~\ref{app:runtime_details}--\ref{app:additional_ablation_scaling}; baselines, benchmarks, and the complete set of training and evaluation hyperparameters are listed in Appendices~\ref{app:baseline_details}, \ref{app:selected14_systems}, \ref{app:experiment_details}, and \ref{app:main_hyperparameters}; the method itself is fully specified by Equations~\ref{eq:method_temporal_block}--\ref{eq:method_posterior_innovation} and Appendices~\ref{app:method_observation_tokens}--\ref{app:method_training_loss}. We will release the training code, the dataset construction scripts, the evaluation pipeline, the generated datasets, and the trained checkpoints upon publication.



\appendix
\clearpage
\addtocontents{toc}{\protect\setcounter{tocdepth}{3}}
\renewcommand{\contentsname}{Appendix Table of Contents}
\tableofcontents
\clearpage

\section{Baseline Details}
\label{app:baseline_details}

Our comparisons cover classical Bayesian filters, robust ensemble variants, task-specific neural estimators, and filters built around pretrained or jointly trained sequence models. This appendix introduces each baseline and details our benchmark adaptations and experimental settings for the results in Table~\ref{tab:panda_foundation_selected14_online}. All methods receive identical test observations, and predictions are converted back to the original state coordinates before evaluation.

\subsection{Classical and Robust Bayesian Filters}
\label{app:classical_baselines}
\label{app:robust_baselines}

These filters implement the prior-to-posterior recursion in Eq.~(\ref{eq:bayes_filtering}) using the known system functions and noise levels in Appendix~\ref{app:selected14_systems}. We follow Section~\ref{sec:problem_formulation}: \(\mathbf{x}_t\in\mathbb{R}^{N}\) is the latent state, \(\mathbf{y}_t\in\mathbb{R}^{M}\) is the observation, \(f\) and \(h\) are the transition and observation functions, and \(\bm{R}\) and \(\bm{Q}\) are the process- and observation-noise covariances.

\begin{baselinelist}
\item \textbf{Extended Kalman Filter (EKF).} As background for the ensemble methods below, the EKF extends the Kalman filter to nonlinear systems by applying first-order Taylor approximations to \(f\) and \(h\). It approximates the posterior as \(p(\mathbf{x}_t\mid\mathbf{y}_{1:t})\approx\mathcal{N}(\bm{\mu}_t,\bm{\Sigma}_t)\), where \(\bm{\mu}_t\in\mathbb{R}^{N}\) and \(\bm{\Sigma}_t\in\mathbb{R}^{N\times N}\) are the posterior mean and covariance. The transition Jacobian at the previous posterior mean is defined as
\[
\mathbf{F}_t:=\left.\frac{\partial f(\mathbf{x})}{\partial\mathbf{x}}\right|_{\mathbf{x}=\bm{\mu}_{t-1}}\in\mathbb{R}^{N\times N}.
\]
The prediction step computes the prior mean and covariance as
\begin{equation}
\tilde{\bm{\mu}}_t=f(\bm{\mu}_{t-1}),\, \tilde{\bm{\Sigma}}_t=\mathbf{F}_t\bm{\Sigma}_{t-1}\mathbf{F}_t^\top+\bm{R}.
\label{eq:baseline_ekf_predict}
\end{equation}
Let \(\mathbf{H}_t\in\mathbb{R}^{M\times N}\) be the Jacobian of \(h\) at the predicted mean \(\tilde{\bm{\mu}}_t\). The update incorporates the current observation through
\begin{equation}
\begin{aligned}
\mathbf{K}_t&=\tilde{\bm{\Sigma}}_t\mathbf{H}_t^\top\bigl(\mathbf{H}_t\tilde{\bm{\Sigma}}_t\mathbf{H}_t^\top+\bm{Q}\bigr)^{-1},\\
\bm{\mu}_t&=\tilde{\bm{\mu}}_t+\mathbf{K}_t\bigl(\mathbf{y}_t-h(\tilde{\bm{\mu}}_t)\bigr),\\
\bm{\Sigma}_t&=(\mathbf{I}_N-\mathbf{K}_t\mathbf{H}_t)\tilde{\bm{\Sigma}}_t.
\end{aligned}
\label{eq:baseline_ekf_update}
\end{equation}
The final state estimate is the mean of the resulting Gaussian posterior approximation, \(\hat{\mathbf{x}}_t=\bm{\mu}_t\).

\item \textbf{Ensemble Kalman Filter (EnKF).} The EnKF~\citep{Evensen1994EnKF} is an ensemble alternative to the EKF for nonlinear and chaotic systems. It approximates the posterior \(p(\mathbf{x}_t\mid\mathbf{y}_{1:t})\) with \(N_{\rm e}\) particles \(\mathbf{x}_t^{(i)}\in\mathbb{R}^{N}\), where \(i=1,\ldots,N_{\rm e}\), and estimates covariances from the ensemble. The prediction step propagates the previous posterior particles through the transition model in Eq.~(\ref{eq:state_space}) to obtain \(\tilde{\mathbf{x}}_t^{(i)}\), whose ensemble approximates the prior \(p(\mathbf{x}_t\mid\mathbf{y}_{1:t-1})\). In the update step, a perturbed predicted observation is sampled for each particle as
\begin{equation}
\mathbf{y}_t^{(i)}\sim\mathcal{N}\!\left(h(\tilde{\mathbf{x}}_t^{(i)}),\bm{Q}\right),\, i=1,\ldots,N_{\rm e}.
\label{eq:baseline_enkf_predict}
\end{equation}
The Kalman gain and updated particles are then computed as
\begin{equation}
\begin{aligned}
\mathbf{K}_t&=\operatorname{Cov}(\tilde{\mathbf{x}}_t^{(i)},\mathbf{y}_t^{(i)})\bigl(\operatorname{Var}(\mathbf{y}_t^{(i)})+10^{-4}\mathbf{I}_M\bigr)^{\dagger},\\
\mathbf{x}_t^{(i)}&=\tilde{\mathbf{x}}_t^{(i)}+\mathbf{K}_t(\mathbf{y}_t-\mathbf{y}_t^{(i)}),\, i=1,\ldots,N_{\rm e},
\end{aligned}
\label{eq:baseline_enkf}
\end{equation}
where \(\operatorname{Cov}\) and \(\operatorname{Var}\) denote the sample cross-covariance and sample covariance across ensemble members, respectively. Our implementation uses the diagonal regularizer and pseudoinverse \(\dagger\) for numerical stability. Finally, averaging the updated ensemble gives the posterior state estimate:
\begin{equation}
\hat{\mathbf{x}}_t=\frac{1}{N_{\rm e}}\sum_{i=1}^{N_{\rm e}}\mathbf{x}_t^{(i)}.
\label{eq:baseline_enkf_mean}
\end{equation}

\item \textbf{Particle Filter (PF).} The bootstrap PF~\citep{8683411} approximates the posterior \(p(\mathbf{x}_t\mid\mathbf{y}_{1:t})\) with \(N_{\rm p}\) weighted particles. It implements the prediction and update in Eq.~(\ref{eq:bayes_filtering}) through sampling, likelihood weighting, and resampling.

\emph{Sampling.} Each previous posterior particle \(\mathbf{x}_{t-1}^{(i)}\) is propagated according to the transition model, producing predictive particles \(\tilde{\mathbf{x}}_t^{(i)}\) that approximate \(p(\mathbf{x}_t\mid\mathbf{y}_{1:t-1})\):
\begin{equation}
\tilde{\mathbf{x}}_t^{(i)}\sim p(\mathbf{x}_t\mid\mathbf{x}_{t-1}^{(i)})=\mathcal{N}\!\left(f(\mathbf{x}_{t-1}^{(i)}),\bm{R}\right),\, i=1,\ldots,N_{\rm p}.
\label{eq:baseline_pf_predict}
\end{equation}
\begin{samepage}
\emph{Weighting.} Each predictive particle is assigned a likelihood measuring how well its predicted observation \(h(\tilde{\mathbf{x}}_t^{(i)})\) agrees with the current observation \(\mathbf{y}_t\). For Gaussian observation noise, these likelihoods are normalized to obtain the particle weights:
\begin{equation}
\begin{aligned}
\tilde w_t^{(i)}&=\mathcal{N}\!\left(\mathbf{y}_t;h(\tilde{\mathbf{x}}_t^{(i)}),\bm{Q}\right),\\
w_t^{(i)}&=\frac{\tilde w_t^{(i)}}{\sum_{k=1}^{N_{\rm p}}\tilde w_t^{(k)}}.
\end{aligned}
\label{eq:baseline_pf_weights}
\end{equation}
\end{samepage}
For numerical stability, we subtract the maximum log likelihood before exponentiation and normalization, preserving the normalized weights while preventing all unnormalized weights from underflowing to zero. The weighted particles approximate \(p(\mathbf{x}_t\mid\mathbf{y}_{1:t})\) and give the state estimate:
\begin{equation}
\hat{\mathbf{x}}_t=\sum_{i=1}^{N_{\rm p}}w_t^{(i)}\tilde{\mathbf{x}}_t^{(i)}.
\label{eq:baseline_pf_mean}
\end{equation}

\emph{Resampling.} After computing \(\hat{\mathbf{x}}_t\), we use systematic resampling to prepare the particle ensemble for the next prediction, allocating more particles to states with larger posterior weights. The procedure selects \(N_{\rm p}\) copies from the weighted particles, allowing a particle to appear multiple times or be omitted. A particle with weight \(w_t^{(i)}\) contributes \(N_{\rm p}w_t^{(i)}\) copies on average, so its relative importance is carried forward through its number of copies. Each selected copy receives equal weight \(1/N_{\rm p}\), and the resulting ensemble \(\{\mathbf{x}_t^{(i)}\}_{i=1}^{N_{\rm p}}\) supplies the next prediction.
\end{baselinelist}

\noindent\textbf{Robust EnKF variants.} To reduce the influence of unusually large observation residuals, we implement innovation rejection, soft weighting, and Huber clipping, following the robust-update families discussed by~\citet{duran2024outlier}. All three variants retain the EnKF prediction, perturbed observations, and gain computation above. For the \(i\)-th particle and observation channel \(m\), define the innovation and its standardized value as
\[
d_{t,m}^{(i)}=y_{t,m}-y_{t,m}^{(i)},\, u_{t,m}^{(i)}=\frac{d_{t,m}^{(i)}}{\sqrt{Q_{mm}}}.
\]
Each variant replaces \(\mathbf{d}_t^{(i)}\) with a modified innovation \(\tilde{\mathbf{d}}_t^{(i)}\) in the common update
\begin{equation}
\mathbf{x}_t^{(i)}=\tilde{\mathbf{x}}_t^{(i)}+\mathbf{K}_t\tilde{\mathbf{d}}_t^{(i)}.
\label{eq:baseline_robust_update}
\end{equation}
\begin{baselinelist}
\item \textbf{EnKFI.} This variant uses a hard decision shared by all ensemble members within each observation channel. It rejects the channel's innovation when the ensemble-averaged squared standardized residual exceeds \(c^2\):
\begin{equation}
\tilde d_{t,m}^{(i)}=d_{t,m}^{(i)}\mathbb{I}\!\left[\frac{1}{N_{\rm e}}\sum_{k=1}^{N_{\rm e}}(u_{t,m}^{(k)})^2\le c^2\right],
\label{eq:baseline_enkfi}
\end{equation}
where \(\mathbb{I}[\cdot]\) is the indicator function. After the particle update in Eq.~(\ref{eq:baseline_robust_update}), EnKFI returns the ensemble mean \(\hat{\mathbf{x}}_t=\frac{1}{N_{\rm e}}\sum_{i=1}^{N_{\rm e}}\mathbf{x}_t^{(i)}\) as the final state estimate.

\item \textbf{EnKFS.} Soft weighting instead reduces each residual continuously, with a separate weight for every ensemble member and observation channel:
\begin{equation}
\tilde d_{t,m}^{(i)}=\frac{d_{t,m}^{(i)}}{\sqrt{1+(u_{t,m}^{(i)}/c)^2}}.
\label{eq:baseline_enkfs}
\end{equation}
Small residuals are nearly unchanged, while larger residuals receive less weight. The final state estimate is the mean \(\hat{\mathbf{x}}_t=\frac{1}{N_{\rm e}}\sum_{i=1}^{N_{\rm e}}\mathbf{x}_t^{(i)}\) of the ensemble updated by Eq.~(\ref{eq:baseline_robust_update}).

\item \textbf{HubEnKF.} The Huberized update clips each standardized innovation at \(\pm c\) and then restores its observation scale:
\begin{equation}
\tilde d_{t,m}^{(i)}=\sqrt{Q_{mm}}\operatorname{clip}(u_{t,m}^{(i)},-c,c).
\label{eq:baseline_hubenkf}
\end{equation}
This preserves innovations within the threshold and caps larger components at \(c\sqrt{Q_{mm}}\). The final state estimate is the mean \(\hat{\mathbf{x}}_t=\frac{1}{N_{\rm e}}\sum_{i=1}^{N_{\rm e}}\mathbf{x}_t^{(i)}\) of the ensemble updated by Eq.~(\ref{eq:baseline_robust_update}).
\end{baselinelist}

\noindent\textbf{Shared implementation settings.} To balance estimation accuracy and computational efficiency, we set \(N_{\rm e}=N_{\rm p}=1{,}000\). We initialize all particles from \(\mathcal{N}(\mathbf{0},\mathbf{I}_N)\) and use \(c=5\) for the three robust variants. Diagonal noise variances have a numerical floor of \(10^{-6}\), and state particles are clipped componentwise to \([-100,100]\) after propagation and correction. In the mismatch experiments of Appendix~\ref{app:model_mismatch_details}, we increase the actual observation covariance while keeping the assumed \(\bm{Q}\) fixed.

\subsection{Learning-Based Filters}
\label{app:learning_baselines}

Following the learning-based baseline selection in LLMFilter~\citep{LLMFilter}, we evaluate MEstimator, RStateNet, ProTran, and KalmanNet, and additionally include RNNYYF. These task-specific models are trained separately on each system, with adaptations to their inputs, output dimensions, and supervised objectives described below. Each model predicts states from observations without receiving target states in its forward pass.

\begin{baselinelist}
\item \textbf{MEstimator.} The estimator of~\citet{MEStimator} was developed alongside a control policy for legged locomotion. We adopt its multilayer-perceptron estimator and replace robot-specific inputs with causal observation history. At time \(t\), we concatenate the standardized observation vectors from \(t,t-2,t-4,t-6\) into a \(4M\)-dimensional input. This input passes through two fully connected hidden layers with 256 and 128 units, respectively, each followed by an ELU activation and dropout 0.1. An \(N\)-unit linear layer produces \(\hat{\mathbf{x}}_t\), and training uses only state-estimation MSE.

\item \textbf{RStateNet.} RobustStateNet~\citep{RStateNet} combines learned motion prediction with a recurrent measurement update for ego-vehicle state estimation. We adapt it to our benchmark by replacing the vehicle-specific sensor branches with a single LSTM branch that processes the full \(M\)-dimensional observation vector together with the previous \(N\)-dimensional state estimate. We resize the output heads to predict an \(N\)-dimensional state increment and an \(N\times M\) correction matrix, matching each system's state and observation dimensions. The current estimate is obtained by adding the predicted increment and a correction based on the observation residual to the previous estimate, and we train the model on each system with state-estimation MSE.

\item \textbf{ProTran.} The probabilistic Transformer of~\citet{ProTran} models temporal structure through a recursively constructed latent representation. Our supervised filtering adaptation maintains a hidden-state history and lets the current query attend both to earlier hidden states and to observations up to the current time. We use representation width 256, eight attention heads, a 16-dimensional latent representation, and dropout 0.1. A latent projection updates the recurrent representation before state decoding. Training uses state-estimation MSE rather than the original variational objective, so this baseline evaluates an adaptation of ProTran's temporal architecture rather than its full probabilistic generative model.

\item \textbf{KalmanNet.} KalmanNet~\citep{KalmanNet} retains known transition and observation functions while learning the Kalman gain with recurrent networks. We adapt the gain-network architecture from the official KalmanNet repository\footnote{\begingroup\hypersetup{urlcolor=black}\url{https://github.com/KalmanNet/KalmanNet_TSP}\endgroup}, including its three GRUs for process, state, and innovation covariance features, and provide system-specific wrappers for the benchmark dynamics and observation maps. Our runs use input and output expansion factors of 5 and 1, respectively, and batch size 1. LLMFilter~\citep{LLMFilter} reports training divergence with the official implementation on high-dimensional systems. Table~\ref{tab:panda_foundation_selected14_online} reports KalmanNet results for the five low-dimensional benchmark systems.

\item \textbf{RNNYYF.} We adapt the recurrent Yau--Yau filtering framework of~\citet{RNNYYF} to learn state estimates directly, without explicitly representing the conditional density. Starting from a learned initial vector, the model updates a 512-dimensional hidden representation at each time step through a feed-forward network with \(\tanh\) activations, using the previous representation and the observation increment \(\mathbf{y}_t-\mathbf{y}_{t-1}\) as inputs. A linear layer then maps the updated representation to the \(N\)-dimensional state estimate \(\hat{\mathbf{x}}_t\). We train the model on complete 200-step windows with unnormalized observations and target states, minimizing state-estimation MSE in the original state coordinates.
\end{baselinelist}

\subsection{Other Baselines}
\label{app:foundation_baselines}

These baselines use sequence models to reduce the amount of manually specified dynamics, but their interfaces and original generalization settings differ. We adapt both to the same benchmark trajectories and report their in-domain filtering performance, with the original model design and our changes distinguished below.

\begin{baselinelist}
\item \textbf{LLMFilter.} LLMFilter~\citep{LLMFilter} maps observations into a frozen language model's embedding space and decodes its representations into state estimates, using System-as-Prompt to describe the filtering task. We build on the official repository\footnote{\begingroup\hypersetup{urlcolor=black}\url{https://github.com/Anonymous-User-2367/LLMFilter}\endgroup}, retain the frozen LLaMA-7B backbone\footnote{\begingroup\hypersetup{urlcolor=black}\url{https://huggingface.co/huggyllama/llama-7b}\endgroup}, and train the observation encoder and state decoder separately for each system, using hidden width 512 and \(\tanh\) activations in both adapters. The public implementation flattens time and observation channels before splitting them into fixed-length scalar segments, so patch boundaries need not coincide with time-step boundaries. Our adaptation preserves the time--channel layout by dividing each 40-step window into two contiguous 20-step patches, each containing all \(M\) observation channels and decoded into the corresponding \(N\) state channels. The encoder input and decoder output dimensions are therefore \(20M\) and \(20N\), respectively. Following the official preprocessing, we standardize observations and target states using training-set statistics. The public evaluation script reports RMSE in this standardized space, so we invert the state standardization before evaluation to compare all methods in the original state coordinates. Appendix~\ref{app:llmfilter_prompts} provides the System-as-Prompt inputs for all fourteen benchmark systems.

\item \textbf{FM-UKF.} FM-UKF~\citep{FM-UKF} combines a learned causal dynamics model with an unscented Kalman filter for posterior correction. Its official implementation\footnote{\begingroup\hypersetup{urlcolor=black}\url{https://github.com/Data-Science-in-Mechanical-Engineering/fm-ukf}\endgroup} was developed for container-ship models with a 10-dimensional state space and 70 dynamics parameters, evaluating transfer to held-out parameterizations with a shared state layout. To adapt the predictor to our benchmark systems, we use a state-only causal Transformer and train one model per state-dimension group, pooling the systems within each group for in-domain evaluation. The adapted architecture uses four layers, width 256, eight attention heads, feed-forward width 1,024, and dropout 0.05, with input and output dimensions matched to each group. We train on standardized 80-step state windows with a next-step MSE loss at every position except the last under causal attention. During filtering, recent posterior state estimates provide temporal context for the learned dynamics prediction, and the UKF incorporates the current observation through the known observation map to obtain the posterior estimate.
\end{baselinelist}

\subsection{Training and Evaluation Settings}
\label{app:baseline_training}

Table~\ref{tab:baseline_training} reports the training settings for the baselines in Table~\ref{tab:panda_foundation_selected14_online}. Each task-specific baseline uses the 480 training trajectories of its target system, while FM-UKF pools those trajectories within each dimension group. Training uses AdamW with weight decay \(10^{-5}\), retaining the checkpoint with the lowest validation loss for evaluation. The 20 test trajectories for each system also serve as the validation set and are excluded from parameter updates. LLMFilter uses the cosine learning-rate scheduler provided in its official implementation, and we also apply cosine decay to the task-specific neural estimators. Our FM-UKF dynamics model uses a constant learning rate. Training stops after 10 consecutive epochs without improvement in validation loss or at the configured epoch limit.

\begin{table}[htbp]
\centering
\small
\setlength{\tabcolsep}{4pt}
\renewcommand{\arraystretch}{1.12}
\newcommand{\trainingheader}[1]{\begin{tabular}{@{}c@{}}#1\end{tabular}}
\begin{tabular*}{\linewidth}{@{\extracolsep{\fill}}lcccc@{}}
\toprule[1pt]
Method & \trainingheader{Window\\length} & Batch size & \trainingheader{Learning\\rate} & \trainingheader{Maximum\\training epochs} \\
\midrule
MEstimator & 40 & 16 & \(10^{-4}\) & 50 \\
RStateNet & 40 & 16 & \(10^{-4}\) & 50 \\
ProTran & 40 & 16 & \(10^{-4}\) & 50 \\
KalmanNet & 40 & 1 & \(10^{-4}\) & 50 \\
RNNYYF & 200 & 16 & \(4\times10^{-4}\) & 50 \\
LLMFilter & 40 & 1 & \(10^{-4}\) & 10 \\
FM-UKF & 80 & \begin{tabular}{@{}l@{}}128 (\(N=2,4\))\\16 (\(N=40,64,72\))\\4 (\(N=96,100\))\end{tabular} & \(10^{-4}\) & 50 \\
\bottomrule[1pt]
\end{tabular*}
\caption{Baseline training settings for Table~\ref{tab:panda_foundation_selected14_online}. Window lengths are measured in time steps, \(N\) denotes state dimension, and maximum training epochs is the preset upper limit for each run.}
\label{tab:baseline_training}
\end{table}

\begin{samepage}
\noindent\textbf{Preprocessing.} MEstimator, RStateNet, ProTran, and KalmanNet standardize input observations using training-set statistics and minimize MSE in the original state coordinates. RNNYYF uses unnormalized observations and target states. Following the official implementations, LLMFilter standardizes observations and states, while FM-UKF standardizes the state sequences used to train its dynamics predictor. We evaluate all methods using RMSE in the original state coordinates. RNNYYF is less competitive on several systems in Table~\ref{tab:panda_foundation_selected14_online}, and its validation loss fluctuates substantially on VL20 and Burgers64. Following the authors' formulation~\citep{RNNYYF}, our recurrent estimator takes the observation increment \(\mathbf{y}_t-\mathbf{y}_{t-1}\) as input. Without normalization, differencing does not bring observation channels with different amplitudes and noise levels to a common scale, which may contribute to unstable optimization. These results suggest that unnormalized observation increments may not provide a robust default for neural estimation across heterogeneous systems.
\par
\end{samepage}

\section{Dataset Details}
\label{app:dataset_details}

During reproduction of LLMFilter~\citep{LLMFilter}, we found nearly constant observation residuals over long intervals in the Selkov, Oscillator, and Hopf datasets: after accounting for their one-step observation alignment, approximately 99.5\% of adjacent residual vectors agree within \(10^{-5}\). We consider the resulting overly smooth observations unsuitable for evaluating filtering under independently sampled per-step noise. Since its public repository does not provide a dataset generator, we regenerate trajectories following the system definitions and simulation settings below, drawing fresh process and observation noise at each time step and aligning observations with their corresponding states. The original seven-system benchmark also includes only two high-dimensional systems, Lorenz96 and VL20, providing limited coverage of high-dimensional dynamics. We therefore expand the benchmark to fourteen systems and use it to evaluate all methods.

\subsection{Benchmark Systems}
\label{app:selected14_systems}

This section documents the fourteen benchmark systems used in the main in-domain experiments. We summarize each system's characteristics and simulation settings in Table~\ref{tab:selected14_system_configs} before describing its dynamics and observation model.

\begin{table}[!t]
\centering
\footnotesize
\setlength{\tabcolsep}{2.3pt}
\renewcommand{\arraystretch}{1.03}
\begin{tabular}{@{}%
>{\raggedright\arraybackslash}p{0.205\textwidth}%
>{\raggedright\arraybackslash}p{0.225\textwidth}%
>{\centering\arraybackslash}p{0.085\textwidth}%
>{\centering\arraybackslash}p{0.055\textwidth}%
>{\centering\arraybackslash}p{0.105\textwidth}%
>{\centering\arraybackslash}p{0.130\textwidth}%
>{\centering\arraybackslash}p{0.125\textwidth}@{}}
\toprule
\textbf{System} & \textbf{System type} & \textbf{State / Obs. Dim.} & \(\Delta t\) & \textbf{Integrator} & \(\bm{R}\) & \(\bm{Q}\) \\
\midrule
Tracking~\citep{smith1962application} & Linear target tracking & \(4 / 4\) & \(0.1\) & Forward Euler & \(0.1\,\mathbf{I}_4\) & \(10\,\mathbf{I}_4\) \\
Double-Pendulum~\citep{levien1993double} & Chaotic rigid-body mechanics & \(4 / 4\) & \(0.01\) & RK4 & \(10^{-4}\mathbf{I}_4\) & \(0.1\,\mathbf{I}_4\) \\
Selkov~\citep{selkov} & Biochemical glycolytic oscillation & \(2 / 2\) & \(0.01\) & RK4 & \(10^{-4}\mathbf{I}_2\) & \(10^{-3}\mathbf{I}_2\) \\
Oscillator~\citep{SINDY} & Canonical nonlinear oscillation & \(2 / 2\) & \(0.01\) & RK4 & \(10^{-4}\mathbf{I}_2\) & \(10^{-3}\mathbf{I}_2\) \\
Hopf~\citep{hassard1981hopf} & Normal-form bifurcation oscillator & \(2 / 2\) & \(0.01\) & RK4 & \(10^{-4}\mathbf{I}_2\) & \(10^{-3}\mathbf{I}_2\) \\
Lorenz96~\citep{lorenz96} & Chaotic atmospheric dynamics & \(72 / 72\) & \(0.01\) & RK4 & \(10^{-4}\mathbf{I}_{72}\) & \(10\,\mathbf{I}_{72}\) \\
Vissio-Lucarini 20 (VL20)~\citep{vissio2020mechanics} & Coupled climate surrogate dynamics & \(72 / 72\) & \(0.01\) & RK4 & \(10^{-4}\mathbf{I}_{72}\) & \(\mathbf{I}_{72}\) \\
LorenzUV40~\citep{wilks2005effects} & Multiscale slow-fast atmosphere & \(40 / 40\) & \(0.005\) & RK4 & \(2.5\times 10^{-7}\mathbf{I}_{40}\) & \(0.1\,\mathbf{I}_{40}\) \\
Vanderpol64~\citep{vanderpol1926relaxation} & Coupled nonlinear oscillator lattice & \(64 / 64\) & \(0.03\) & RK4 & \(1.8\times 10^{-5}\mathbf{I}_{64}\) & \(0.1\,\mathbf{I}_{64}\) \\
Rossler96~\citep{rossler1976equation} & Coupled chaotic flow cells & \(96 / 96\) & \(0.02\) & RK4 & \(10^{-3}\mathbf{I}_{96}\) & \(\mathbf{I}_{96}\) \\
Duffing64~\citep{duffing1918forced} & Coupled bistable oscillators & \(64 / 64\) & \(0.05\) & RK4 & \(5\times 10^{-5}\mathbf{I}_{64}\) & \(0.08\,\mathbf{I}_{64}\) \\
Burgers64~\citep{burgers1948mathematical} & Viscous transport field & \(64 / 64\) & \(0.05\) & RK4 & \(1.25\times 10^{-5}\mathbf{I}_{64}\) & \(0.05\,\mathbf{I}_{64}\) \\
Fhn64~\citep{fitzhugh1961impulses,nagumo1962active} & Excitable reaction-diffusion lattice & \(64 / 64\) & \(0.05\) & RK4 & \(5\times 10^{-5}\mathbf{I}_{64}\) & \(0.08\,\mathbf{I}_{64}\) \\
Cubic-Sensoring100\newline\citep{hazewinkel1983cubic} & Nonlinear sensing stress test & \(100 / 100\) & \(0.01\) & Forward Euler & \(10^{-2}\mathbf{I}_{100}\) & \(10^{-2}\mathbf{I}_{100}\) \\
\bottomrule
\end{tabular}
\caption{Configuration of the fourteen in-domain benchmark systems. Here \(\bm{R}\) and \(\bm{Q}\) denote the process and observation noise covariances used for testing, with 20 test trajectories per system.}
\label{tab:selected14_system_configs}
\end{table}

These systems cover linear tracking, nonlinear oscillations, chaotic flows, and spatial coupling, providing diverse settings for state estimation. We numerically integrate the differential equations below using the methods and time steps in Table~\ref{tab:selected14_system_configs} to obtain the discrete-time state transition function \(f\) in Eq.~(\ref{eq:state_space}). The filtering model includes additive process noise \(\bm{\omega}_t\sim\mathcal{N}(\mathbf{0},\bm{R})\) and observation noise \(\bm{\epsilon}_t\sim\mathcal{N}(\mathbf{0},\bm{Q})\). We use \(\mathbf{I}_d\) for the \(d\times d\) identity matrix and write the coupled dynamics componentwise, with the index ranges and boundary conditions specified for each system.

\noindent\textbf{Tracking~\citep{smith1962application}.} This system is the classical constant-velocity kinematic model used in radar tracking, vision-based motion estimation, and navigation. The filtering task is to estimate the target's position and velocity from noisy observations \mbox{\(\mathbf{y}_t=\mathbf{x}_t+\bm{\epsilon}_t\)}, with dynamics
\[
\dot{\mathbf{x}}=
\begin{bmatrix}
0 & 0 & 1 & 0\\
0 & 0 & 0 & 1\\
0 & 0 & 0 & 0\\
0 & 0 & 0 & 0
\end{bmatrix}\mathbf{x},
\mathbf{x}=[p_x,p_y,v_x,v_y]^\top.
\]

\noindent\textbf{Pendulum~\citep{levien1993double}.} This four-dimensional benchmark models a planar double pendulum and tests whether a filter can track strongly coupled angular motion whose errors grow rapidly in the chaotic regime. We use the state \(\mathbf{x}=[\theta_1,\omega_1,\theta_2,\omega_2]^\top\), unit masses and lengths, \(\delta=\theta_2-\theta_1\), and \(D=1+\sin^2\delta\). The dynamics are
\[
\left\{
\begin{aligned}
\dot{\theta}_1 &= \omega_1,\\
\dot{\omega}_1
&=\frac{\omega_1^2\sin\delta\cos\delta
+g\sin\theta_2\cos\delta
+\omega_2^2\sin\delta
-2g\sin\theta_1}{D},\\
\dot{\theta}_2 &= \omega_2,\\
\dot{\omega}_2
&=\frac{-\omega_2^2\sin\delta\cos\delta
+2g\sin\theta_1\cos\delta
-2\omega_1^2\sin\delta
-2g\sin\theta_2}{D}.
\end{aligned}
\right.
\]
Here \(g=9.8\), and the observation model is \mbox{\(\mathbf{y}_t=\mathbf{x}_t+\bm{\epsilon}_t\)}.

\noindent\textbf{Selkov~\citep{selkov}.} The Selkov model describes glycolytic oscillations in biochemical reaction networks and is a standard low-dimensional model for metabolic feedback. It captures how nonlinear reaction terms generate self-sustained oscillatory behavior, making it a useful benchmark for filtering biochemical dynamics from noisy concentration measurements:
\[
\dot{\mathbf{x}}=
\begin{bmatrix}
-x_1+a x_2+x_1^2x_2\\
b-a x_2-x_1^2x_2
\end{bmatrix},
\mathbf{x}=[x_1,x_2]^\top,
\]
with \(a=0.08\), \(b=0.6\), and observations \mbox{\(\mathbf{y}_t=\mathbf{x}_t+\bm{\epsilon}_t\)}.

\noindent\textbf{Oscillator~\citep{SINDY}.} This benchmark is a compact nonlinear oscillator with cubic restoring terms. Although low-dimensional, it introduces nontrivial state coupling and strongly nonlinear drift, making it a useful stress test for whether a filter can recover smooth latent motion from corrupted observations without relying on a linear update rule:
\[
\dot{\mathbf{x}}=
\begin{bmatrix}
-a x_1^3+b x_2^3\\
-b x_1^3-a x_2^3
\end{bmatrix},
\mathbf{x}=[x_1,x_2]^\top,
\]
with \(a=0.1\), \(b=2.0\), and observations \mbox{\(\mathbf{y}_t=\mathbf{x}_t+\bm{\epsilon}_t\)}.

\noindent\textbf{Hopf~\citep{hassard1981hopf}.} The Hopf normal form models the onset of self-sustained oscillation as a stable equilibrium loses stability. With applications in fluid instabilities, chemical reactions, and electrical circuits, it provides a benchmark for recovering oscillatory states from noisy observations:
\[
\dot{\mathbf{x}}=
\begin{bmatrix}
\mu x_1+\omega x_2-a x_1\|\mathbf{x}\|_2^2\\
-\omega x_1+\mu x_2-a x_2\|\mathbf{x}\|_2^2
\end{bmatrix},
\mathbf{x}=[x_1,x_2]^\top,
\]
with \(\mu=0.5\), \(\omega=1.0\), \(a=1.0\), and observations \mbox{\(\mathbf{y}_t=\mathbf{x}_t+\bm{\epsilon}_t\)}.

\noindent\textbf{Lorenz96~\citep{lorenz96}.} The Lorenz96 system, introduced by Lorenz, is a standard surrogate for atmospheric transport and weather predictability on a periodic domain. It is widely used in data assimilation because it combines simple local interactions with strongly chaotic global evolution, creating a representative medium-scale benchmark for online state reconstruction. Observations follow \mbox{\(\mathbf{y}_t=\mathbf{x}_t+\bm{\epsilon}_t\)}, and the dynamics are
\[
\dot{x}_{j}=\bigl(x_{j+1}-x_{j-2}\bigr)x_{j-1}-x_{j}+F,\, j=1,\ldots,72.
\]
Here \(x_{j}\) denotes the \(j\)-th atmospheric variable, \(F=8\) is the external forcing, and the system uses periodic boundary conditions \(x_{73}=x_{1}\), \(x_{0}=x_{72}\), and \(x_{-1}=x_{71}\).

\noindent\textbf{Vissio-Lucarini 20 (VL20)~\citep{vissio2020mechanics}.} The Vissio-Lucarini 20 model extends the Lorenz framework with an additional temperature-like field and provides a minimalist but expressive benchmark for coupled climate dynamics. Relative to Lorenz96, it introduces direct interaction between circulation-like and thermodynamic variables, so filtering involves recovering both within-field evolution and cross-field exchange:
\[
\begin{aligned}
\dot{\phi}_{j}&=\bigl(\phi_{j+1}-\phi_{j-2}\bigr)\phi_{j-1}-\gamma\phi_{j}-\epsilon\theta_{j}+F,\\
\dot{\theta}_{j}&=\phi_{j+1}\theta_{j+2}-\phi_{j-1}\theta_{j-2}+\epsilon\phi_{j}-\gamma\theta_{j}+G.
\end{aligned}
\]
Here \(\mathbf{x}=[\bm{\phi}^\top,\bm{\theta}^\top]^\top\) stacks the circulation-like variables \(\phi_j\) and temperature-like variables \(\theta_j\) for \(j=1,\ldots,36\), with \(F=10\), \(G=0\), \(\gamma=1\), and \(\epsilon=1\). The model uses the same periodic boundary convention as Lorenz96. The observation model is \mbox{\(\mathbf{y}_t=\mathbf{x}_t+\bm{\epsilon}_t\)}.

\noindent\textbf{LorenzUV40~\citep{wilks2005effects}.} This is a multiscale slow-fast chaotic model with eight slow atmospheric variables and 32 fast variables. The coupling between these groups makes it a benchmark for filtering across distinct temporal scales. We use \(\mathbf{x}=[\mathbf{u}^\top,\mathbf{v}^\top]^\top\), where \(\mathbf{u}\in\mathbb{R}^{8}\) and \(\mathbf{v}\in\mathbb{R}^{32}\), with observations \mbox{\(\mathbf{y}_t=\mathbf{x}_t+\bm{\epsilon}_t\)}. Each slow variable \(u_i\) is coupled to the four fast variables with indices \(j=4i-3,\ldots,4i\). For \(i=1,\ldots,8\) and these associated fast indices \(j\), the dynamics are
\[
\left\{
\begin{aligned}
\dot{u}_i&=(u_{i-1}-u_{i+2})u_{i+1}-u_i+20-\frac{hc}{b}\sum_{k=1}^{4}v_{4(i-1)+k},\\
\dot{v}_j&=cb(v_{j-1}-v_{j+2})v_{j+1}-cv_j+\frac{hc}{b}u_i.
\end{aligned}
\right.
\]
The slow and fast indices wrap around their respective groups: \(u_{i+8}=u_i\) and \(v_{j+32}=v_j\). We set the coupling parameter \(h=1\), the amplitude scale \(b=10\), and the time-scale parameter \(c=10\).

\noindent\textbf{Vanderpol64~\citep{vanderpol1926relaxation}.} This system is a lattice of coupled Van der Pol oscillators, a classical model for self-excited oscillation in circuits and distributed oscillatory media. Filtering this lattice requires recovering oscillatory motion governed by nonlinear damping and coupling between adjacent oscillators. We use \(\mathbf{x}=[\mathbf{q}^\top,\mathbf{v}^\top]^\top\), where \(q_i\) and \(v_i\) are the position and velocity of oscillator \(i\), respectively. Observations follow \mbox{\(\mathbf{y}_t=\mathbf{x}_t+\bm{\epsilon}_t\)}. For \(i=1,\ldots,32\),
\[
\left\{
\begin{aligned}
\dot{q}_i&=v_i,\\
\dot{v}_i&=\mu(1-q_i^2)v_i-q_i+\kappa(q_{i+1}-2q_i+q_{i-1}).
\end{aligned}
\right.
\]
The term \(q_{i+1}-2q_i+q_{i-1}\) couples each oscillator to its two neighbors on a ring, with \(q_0=q_{32}\) and \(q_{33}=q_1\). We set \(\mu=2.0\) and the coupling strength \(\kappa=0.15\).

\noindent\textbf{Rossler96~\citep{rossler1976equation}.} This system is a chain of coupled R\"ossler cells and serves as a high-dimensional spatiotemporal chaos benchmark. It represents locally interacting chaotic subsystems whose collective behavior remains difficult to estimate online because the latent state mixes oscillatory motion, stretching, and local transport across many channels. Each cell \(i\) has three coordinates \(r_{x,i}\), \(r_{y,i}\), and \(r_{z,i}\), and the 96-dimensional state stacks these triples in cell order, with observations \mbox{\(\mathbf{y}_t=\mathbf{x}_t+\bm{\epsilon}_t\)}. For \(i=1,\ldots,32\),
\[
\left\{
\begin{aligned}
\dot{r}_{x,i}&=-r_{y,i}-r_{z,i}+\kappa(r_{x,i+1}-2r_{x,i}+r_{x,i-1}),\\
\dot{r}_{y,i}&=r_{x,i}+a r_{y,i},\\
\dot{r}_{z,i}&=b+r_{z,i}(r_{x,i}-c).
\end{aligned}
\right.
\]
The cells interact through their \(r_x\) coordinates, with periodic boundaries \(r_{x,0}=r_{x,32}\) and \(r_{x,33}=r_{x,1}\). We set \(a=0.2\), \(b=0.2\), \(c=5.7\), and the coupling strength \(\kappa=0.08\).

\noindent\textbf{Duffing64~\citep{duffing1918forced}.} This system is a coupled Duffing oscillator chain with local bistable dynamics. Duffing-type systems arise in nonlinear structural vibration and forced mechanical response, and their double-well geometry makes filtering sensitive to whether the estimator preserves the correct oscillatory regime over time. We use \(\mathbf{x}=[\mathbf{q}^\top,\mathbf{v}^\top]^\top\), where \(q_i\) and \(v_i\) are the position and velocity of oscillator \(i\), respectively. Observations follow \mbox{\(\mathbf{y}_t=\mathbf{x}_t+\bm{\epsilon}_t\)}. For \(i=1,\ldots,32\),
\[
\left\{
\begin{aligned}
\dot{q}_i&=v_i,\\
\dot{v}_i&=q_i-q_i^3-\eta v_i+\kappa(q_{i+1}-2q_i+q_{i-1}).
\end{aligned}
\right.
\]
The neighbor-coupling term uses periodic boundaries \(q_0=q_{32}\) and \(q_{33}=q_1\). We set the damping coefficient \(\eta=0.22\) and the coupling strength \(\kappa=0.18\).

\noindent\textbf{Burgers64~\citep{burgers1948mathematical}.} This system is a discretized viscous Burgers field, included as a PDE-like transport benchmark. Burgers dynamics capture the competition between nonlinear advection and diffusion and are widely used as a simplified model for transport and shock-like structure formation. The filtering task is to recover the evolving field from noisy measurements at 64 grid points. We use \(\mathbf{x}=\mathbf{u}\in\mathbb{R}^{64}\), where \(u_i\) is the field value at grid point \(i\), with observations \mbox{\(\mathbf{y}_t=\mathbf{u}_t+\bm{\epsilon}_t\)}. We approximate the spatial derivatives by centered finite differences:
\[
\dot{u}_i=-u_i\frac{u_{i+1}-u_{i-1}}{2\Delta x}
+\nu\frac{u_{i+1}-2u_i+u_{i-1}}{(\Delta x)^2}
+0.12\sin\!\left(\frac{2\pi(i-1)}{64}\right).
\]
Here \(i=1,\ldots,64\), \(\Delta x=1\) is the spatial grid spacing, and \(\nu=0.08\) is the viscosity. The two difference quotients approximate the first and second spatial derivatives, respectively, and the sine term provides a fixed spatial forcing. Periodic boundaries give \(u_0=u_{64}\) and \(u_{65}=u_1\).

\noindent\textbf{Fhn64~\citep{fitzhugh1961impulses,nagumo1962active}.} This system is a FitzHugh--Nagumo reaction-diffusion lattice for excitable media. FitzHugh--Nagumo dynamics are a standard abstraction for neuronal and cardiac excitation, and the coupled lattice form produces traveling activity patterns that require the filter to reconstruct both local activation and spatial propagation. We use \(\mathbf{x}=[\mathbf{u}^\top,\mathbf{v}^\top]^\top\), where \(u_i\) is the activation variable and \(v_i\) is the recovery variable at site \(i\). For \(i=1,\ldots,32\),
\[
\left\{
\begin{aligned}
\dot{u}_i&=u_i-\frac{u_i^3}{3}-v_i+0.7+0.08(u_{i+1}-2u_i+u_{i-1}),\\
\dot{v}_i&=0.08(u_i+0.7-0.8v_i)+0.02(v_{i+1}-2v_i+v_{i-1}).
\end{aligned}
\right.
\]
The neighbor-difference terms describe diffusion of activation and recovery between adjacent sites, with coefficients \(0.08\) and \(0.02\), respectively. Both fields have periodic boundaries: \(u_0=u_{32}\), \(u_{33}=u_1\), \(v_0=v_{32}\), and \(v_{33}=v_1\). The observation model is \mbox{\(\mathbf{y}_t=\mathbf{x}_t+\bm{\epsilon}_t\)}.

\noindent\textbf{Cubic-Sensoring100~\citep{hazewinkel1983cubic}.} This benchmark stresses nonlinear observation handling rather than partial observability. The latent dynamics are comparatively mild, but the cubic sensor map distorts the observation space enough that accurate filtering depends on learning a robust inverse relation from noisy measurements to latent state. Let \(J\in\mathbb{R}^{100\times 100}\) be the upper-shift matrix with \(J_{i,i+1}=1\) for \(i<100\) and zeros elsewhere. Then
\[
\dot{\mathbf{x}}=-0.5\mathbf{x}+\cos(\mathbf{x})+0.1J\mathbf{x},
\]
where \(\cos(\mathbf{x})\) acts elementwise and observations satisfy \mbox{\(y_{t,m}=x_{t,m}^3+\epsilon_{t,m}\)} for \(m=1,\ldots,100\).

\subsection{Benchmark Split and Evaluation Protocol}
\label{app:benchmark_protocol}

For each of the fourteen in-domain systems, we use 480 training trajectories and 20 test trajectories, each containing \(T=200\) time steps. On each system, FilterFM and the applicable baselines are evaluated on the same 20 test trajectories, with per-system raw RMSE reported in Table~\ref{tab:panda_foundation_selected14_online}.

\subsection{Foundation Training Corpus Construction}
\label{app:training_corpus}

Beyond the fourteen benchmark systems, we include Panda-derived low-dimensional trajectories and high-dimensional composites to broaden dynamical diversity and dimension range~\citep{lai2026panda}. Table~\ref{tab:main_corpus_composition} gives the exact composition of the main model's 15,120 training trajectories and the corresponding held-out sets. The 620 held-out trajectories form the validation pool, including the 280 benchmark trajectories used for Table~\ref{tab:panda_foundation_selected14_online}.

\begin{table}[!htbp]
\centering
\small
\setlength{\tabcolsep}{4pt}
\renewcommand{\arraystretch}{1.15}
\begin{tabular*}{\linewidth}{@{\extracolsep{\fill}}lrrr@{}}
\toprule[1pt]
\textbf{Data source} & \textbf{Training} & \textbf{Held-out} & \textbf{Total} \\
\midrule
In-domain benchmark (14 systems) & 6,720 & 280 & 7,000 \\
Panda low-dimensional (790 systems) & 3,360 & 140 & 3,500 \\
Panda high-dimensional composites & 5,040 & 200 & 5,240 \\
\midrule
\textbf{Total for training domains} & \textbf{15,120} & \textbf{620} & \textbf{15,740} \\
\midrule
Panda zero-shot (457 systems) & 0 & 8,207 & 8,207 \\
\bottomrule[1pt]
\end{tabular*}
\caption{Composition of the main training corpus and held-out evaluation sets. All trajectories contain 200 time steps.}
\label{tab:main_corpus_composition}
\end{table}

\noindent\textbf{In-Domain Benchmark Data.} The fourteen systems contribute 6,720 of the 15,120 training trajectories in FilterFM's joint corpus. The resulting model is evaluated on all fourteen systems without system-specific fine-tuning. Per-system baselines, including LLMFilter, use the same 480 training trajectories from their target system, while FM-UKF pools systems with the same state dimension.

\noindent\textbf{Panda Low-Dimensional Systems.} We use trajectories from the Panda dataset\footnote{\begingroup\hypersetup{urlcolor=black}\url{https://huggingface.co/datasets/GilpinLab/skew40}\endgroup}. To formulate filtering tasks from these clean trajectories, we use \(T=200\) consecutive states as \(\mathbf{X}=\mathbf{x}_{1:T}\in\mathbb{R}^{T\times N}\) and observe each state coordinate with additive Gaussian noise whose scale is relative to the trajectory amplitude:
\begin{equation}
y_{t,m}=x_{t,m}+\epsilon_{t,m},\, \epsilon_{t,m}\sim\mathcal{N}(0,\alpha^2\sigma_m^2(\mathbf{X})),\, \alpha\sim\mathrm{LogUniform}(0.15,0.75),
\label{eq:app_panda_lowdim}
\end{equation}
where \(\sigma_m(\mathbf{X})\) is the standard deviation of state coordinate \(m\) along the trajectory. Given \(\mathbf{X}\) and \(\alpha\), noise is sampled independently across time steps and coordinates, with observation covariance \(\bm{Q}\) defined by \(Q_{mm}=\alpha^2\sigma_m^2(\mathbf{X})\).

\noindent\textbf{Panda High-Dimensional Composites.} To extend training beyond low-dimensional dynamics, we synthesize high-dimensional filtering systems by combining trajectories from multiple Panda systems. For a target dimension \(N\in\{24,32,40,48,56,64,72,80,88,96\}\), we first normalize each source window channel-wise, then concatenate enough normalized blocks to exceed \(N\), select \(N\) columns, and apply a fixed random orthogonal mixing matrix:
\begin{equation}
\tilde{\mathbf{X}}^{(b)}_{:,j}=\frac{\mathbf{X}^{(b)}_{:,j}-\mu^{(b)}_j}{\sigma^{(b)}_j+\delta},\mathbf{Z}=\Pi_N\operatorname{concat}(\tilde{\mathbf{X}}^{(1)},\ldots,\tilde{\mathbf{X}}^{(B)}),\mathbf{x}^{\rm comp}_{1:T}=\mathbf{Z}\mathbf{O}_N^\top,\mathbf{O}_N^\top\mathbf{O}_N=\mathbf{I}_N,
\label{eq:app_panda_composite}
\end{equation}
where \(\delta>0\) is a small numerical stabilizer, \(\Pi_N\) denotes column selection to dimension \(N\), and \(\mathbf{O}_N\) is an orthogonal mixing matrix. Observations are then generated with the same identity-plus-noise construction as Eq.~(\ref{eq:app_panda_lowdim}), but with \(\alpha\sim\mathrm{LogUniform}(0.20,0.90)\). These composites provide the model with high-dimensional training systems whose variable count and cross-channel structure differ substantially from the benchmark systems.

\begin{figure}[!htbp]
\centering
\includegraphics[width=\linewidth]{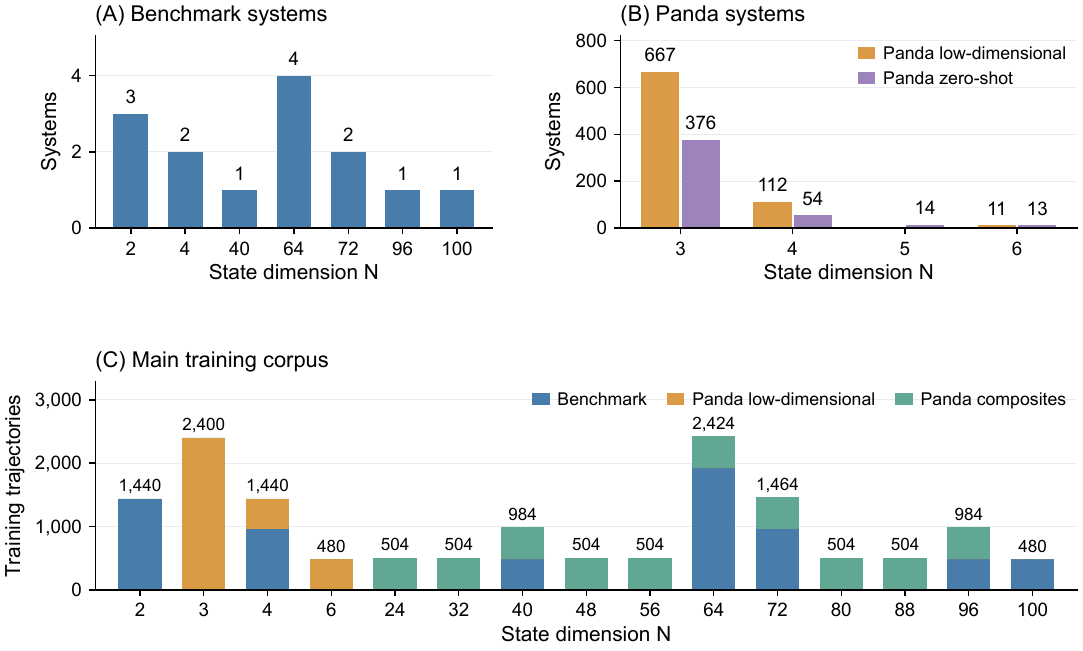}
\caption{State-dimension distributions for (A) the fourteen benchmark systems and (B) the low-dimensional and zero-shot subsets in Table~\ref{tab:main_corpus_composition}. (C) The 15,120 training trajectories, stacked by data source. All horizontal axes display state dimension \(N\) as discrete categories.}
\label{fig:corpus_dimensions}
\end{figure}

\noindent\textbf{Held-out Zero-shot Systems.} For the zero-shot experiments in Section~\ref{sec:exp_zeroshot}, we use 8,207 trajectories from 457 unseen Panda systems with three to six state variables. Each trajectory contains \(T=200\) consecutive states, and observations follow the same Gaussian noise model as Eq.~(\ref{eq:app_panda_lowdim}). These systems are excluded from training and fine-tuning, allowing us to evaluate transfer to previously unseen dynamics under the same observation construction. Figure~\ref{fig:corpus_dimensions} summarizes the state dimensions of the benchmark and Panda systems together with the training-data allocation. The benchmark covers dimensions from 2 to 100, while the low-dimensional Panda systems are concentrated at dimension 3. The composites extend Panda-derived training data across ten dimensions from 24 to 96.

\section{More Experiment Results}
\label{app:experiment_details}

This section extends the main experiments with per-system runtime comparisons, observation-noise misspecification results, and architectural ablations. We also report the prior-loss and patch-length sweeps and examine three aspects of scaling: the diversity of training systems, model capacity, and the number of training trajectories per system. Together, these experiments assess the computational cost, robustness, and transfer behavior of FilterFM and clarify how its components contribute to filtering accuracy.

\noindent\textbf{Evaluation Metrics and Aggregation.}
We use root mean square error (RMSE) as the primary accuracy metric, following the squared-error criteria commonly used in Bayesian filtering and state estimation~\citep{KalmanNet,LLMFilter}. Compared with mean absolute error (MAE), which is also common in time-series forecasting, RMSE gives greater weight to large estimation errors and is consistent with the minimum-mean-square-error criterion for state estimation. For an evaluation window of \(T\) time steps and \(N\) state variables, let \(x_{t,n}\) and \(\hat{x}_{t,n}\) denote the true state and its estimate in the original coordinates. We compute
\[
\begin{aligned}
\mathrm{RMSE}_{\rm raw}
&=\sqrt{\frac{1}{TN}\sum_{t=1}^{T}\sum_{n=1}^{N}(\hat{x}_{t,n}-x_{t,n})^2},\\
\mathrm{RMSE}_{\rm normalized}
&=\sqrt{\frac{1}{TN}\sum_{t=1}^{T}\sum_{n=1}^{N}\left(\frac{\hat{x}_{t,n}-x_{t,n}}{\sigma^x_n}\right)^2}.
\end{aligned}
\]
Here \(\sigma^x_n\) is the standard deviation of the ground-truth state channel over the evaluated window, with a numerical floor of \(10^{-6}\), matching the instance normalization in Section~\ref{sec:method_encoder}. Only valid time steps and state channels contribute to the averages. Raw RMSE measures errors after reversing any state normalization, while normalized RMSE places channels with different amplitudes on a comparable scale. FilterFM's point estimate is its predicted posterior median.

In Table~\ref{tab:panda_foundation_selected14_online}, each per-system score is the square root of the mean squared error over all 20 test trajectories, time steps, and valid state channels. The Average row is the arithmetic mean of the available per-system scores. For a long trajectory evaluated through multiple windows, we sum the squared errors and evaluated-coordinate counts across its windows before dividing and taking the square root. The resulting trajectory RMSEs are summarized by their mean or median. Window-level summaries instead apply these statistics directly to individual window RMSEs, as specified in the scaling experiments. In experiments with multiple seeds, we repeat this evaluation for each seed and report the mean and sample standard deviation of the resulting scores across seeds.

\subsection{Runtime Evaluation Details}
\label{app:runtime_details}

Table~\ref{tab:runtime_details} reports the per-system runtime values used in Figure~\ref{fig:runtime_selected14}. All measurements are obtained on a single NVIDIA A800 GPU, exclude model loading time, and are averaged over the 20 held-out testing trajectories for each benchmark system. LLMFilter is timed with the full System-as-Prompt input protocol, including instruction and example tokens, to retain the input configuration used for its accuracy evaluation.

The runtime gap widens as the state dimension increases. EnKF and PF propagate 1,000 samples at every time step, so higher-dimensional systems increase the work required to evolve the samples and evaluate their observations. EnKF additionally forms covariance matrices and solves for the observation update, while PF evaluates particle likelihoods and resamples the population. These repeated operations make high-dimensional inference substantially more expensive: EnKF takes 0.4834--0.8383 ms per step on the five low-dimensional systems, compared with 7.9090--15.9068 ms on the nine high-dimensional systems. Sampling-based inference thus incurs a cost at every update even though no neural network is involved.

LLMFilter processes observation embeddings together with its System-as-Prompt tokens through the frozen LLaMA-7B backbone. Freezing the backbone reduces training requirements \emph{but does not} remove this computation at inference. FM-UKF uses its neural dynamics model for prediction and still performs recursive UKF covariance propagation and posterior correction. This combination of neural prediction and recursive filtering makes FM-UKF the slowest method on average, and its cost grows with the number of sigma points in higher dimensions. FilterFM instead reconstructs states directly from learned observation representations without propagating an ensemble or solving a UKF update, giving the lowest measured runtime on all fourteen systems in Table~\ref{tab:runtime_details}.

\begin{table}[!ht]
\centering
\footnotesize
\renewcommand{\arraystretch}{1.2}
\setlength{\tabcolsep}{8.0pt}
\begin{tabular}{@{}lrrrrr@{}}
\toprule
\textbf{Dataset} & \textbf{FilterFM} & \textbf{LLMFilter} & \textbf{FM-UKF} & \textbf{EnKF} & \textbf{PF} \\
\midrule
Tracking & 0.1063 & 1.4186 & 1.8507 & 0.5247 & 0.3518 \\
Pendulum & 0.1062 & 1.5954 & 1.8457 & 0.8383 & 0.6614 \\
Selkov & 0.1067 & 1.4057 & 1.8379 & 0.4834 & 0.3767 \\
Oscillator & 0.1089 & 1.8586 & 1.8305 & 0.7892 & 0.6796 \\
Hopf & 0.1122 & 2.2081 & 1.8338 & 0.5068 & 0.3949 \\
\midrule
Lorenz96 & 0.1175 & 6.5480 & 19.0996 & 9.9052 & 4.8176 \\
VL20 & 0.1294 & 4.5234 & 16.1513 & 11.1650 & 6.4611 \\
LorenzUV40 & 0.1470 & 6.6081 & 6.6590 & 7.9090 & 5.0657 \\
Vanderpol64 & 0.1592 & 9.3108 & 13.3544 & 8.7596 & 4.5340 \\
Rossler96 & 0.1762 & 4.6360 & 28.6942 & 14.0968 & 6.4964 \\
Duffing64 & 0.1766 & 4.6714 & 14.3440 & 11.7296 & 6.9267 \\
Burgers64 & 0.1825 & 7.0708 & 14.1660 & 10.5132 & 5.7216 \\
Fhn64 & 0.2030 & 5.1329 & 12.1294 & 13.4692 & 8.0933 \\
Cubic-Sensoring100 & 0.2028 & 2.4218 & 46.4361 & 15.9068 & 5.8194 \\
\midrule
\rowcolor{black!6}
\textbf{Average} & \textbf{0.1453} & \textbf{4.2435} & \textbf{12.8738} & \textbf{7.6140} & \textbf{4.0286} \\
\bottomrule
\end{tabular}
\caption{Runtime details for the 14-system benchmark. All methods are evaluated on a single NVIDIA A800 GPU. Values are measured in ms/step with model loading excluded. LLMFilter timing includes the full System-as-Prompt input. Lower is better.}
\label{tab:runtime_details}
\end{table}

\subsection{Model Mismatch Evaluation Details}
\label{app:model_mismatch_details}

The model-mismatch experiment tests whether a trained estimator remains reliable when observations are noisier than expected. We keep the held-out state sequences unchanged and increase the Gaussian observation noise at each time step. For observation covariance expansion ratio \(c\in\{1,25,50,75,100\}\), the test observations follow
\[
\mathbf{y}^{(c)}_t=h(\mathbf{x}_t)+\boldsymbol{\epsilon}^{(c)}_t,
\quad \boldsymbol{\epsilon}^{(c)}_t\sim\mathcal{N}(\mathbf{0},c\mathbf{Q}).
\]
Thus, \(c\) multiplies the noise covariance, while the corresponding noise standard deviation increases by \(\sqrt{c}\). The underlying dynamics and target states remain the same, allowing the comparison to focus on observation corruption. We feed these observations to the trained models without further fitting, and the online Bayesian filters continue to use the nominal covariance \(\mathbf{Q}\). Predictions are evaluated against the unchanged states in their original coordinates using raw-space RMSE. Table~\ref{tab:model_mismatch_details} reports pooled RMSE over all 20 test trajectories, time steps, and valid state channels, together with the population standard deviation of the 20 trajectory-wise RMSEs for each system and setting.

As the actual noise increases, a filter using the nominal covariance can place too much weight on unreliable observations. FilterFM achieves lower pooled RMSE than the compared methods in most evaluated settings.


\begin{table}[!t]
\centering
\scriptsize
\renewcommand{\arraystretch}{1.0}
\setlength{\tabcolsep}{2.2pt}
\resizebox{\linewidth}{!}{%
\begin{tabular}{@{}llccccc@{}}
\toprule
\textbf{Dataset} & \textbf{Method} & \textbf{OCER=1} & \textbf{OCER=25} & \textbf{OCER=50} & \textbf{OCER=75} & \textbf{OCER=100} \\
\midrule
\textbf{Tracking} & \cellcolor[HTML]{EAF7F0}FilterFM & \cellcolor[HTML]{EAF7F0}\(\mathbf{0.968{\scriptscriptstyle\pm 0.118}}\) & \cellcolor[HTML]{EAF7F0}\(\mathbf{3.028{\scriptscriptstyle\pm 0.579}}\) & \cellcolor[HTML]{EAF7F0}\(\mathbf{4.337{\scriptscriptstyle\pm 0.868}}\) & \cellcolor[HTML]{EAF7F0}\(\mathbf{5.257{\scriptscriptstyle\pm 1.091}}\) & \cellcolor[HTML]{EAF7F0}\(\mathbf{5.972{\scriptscriptstyle\pm 1.219}}\) \\
 & LLM-Filter & \(1.994{\scriptscriptstyle\pm 0.672}\) & \(8.839{\scriptscriptstyle\pm 3.140}\) & \(11.438{\scriptscriptstyle\pm 4.237}\) & \(12.902{\scriptscriptstyle\pm 4.844}\) & \(13.921{\scriptscriptstyle\pm 5.307}\) \\
 & EnKF & \(1.785{\scriptscriptstyle\pm 1.020}\) & \(4.300{\scriptscriptstyle\pm 0.582}\) & \(5.865{\scriptscriptstyle\pm 0.564}\) & \(7.094{\scriptscriptstyle\pm 0.608}\) & \(8.134{\scriptscriptstyle\pm 0.677}\) \\
 & EnKFS & \(1.787{\scriptscriptstyle\pm 1.023}\) & \(3.938{\scriptscriptstyle\pm 0.679}\) & \(5.193{\scriptscriptstyle\pm 0.726}\) & \(6.154{\scriptscriptstyle\pm 0.828}\) & \(6.949{\scriptscriptstyle\pm 0.932}\) \\
 & HubEnKF & \(1.785{\scriptscriptstyle\pm 1.018}\) & \(4.193{\scriptscriptstyle\pm 0.615}\) & \(5.598{\scriptscriptstyle\pm 0.664}\) & \(6.649{\scriptscriptstyle\pm 0.767}\) & \(7.502{\scriptscriptstyle\pm 0.877}\) \\
\addlinespace[1.5pt]
\textbf{Selkov} & \cellcolor[HTML]{EAF7F0}FilterFM & \cellcolor[HTML]{EAF7F0}\(\mathbf{0.0155{\scriptscriptstyle\pm 0.0010}}\) & \cellcolor[HTML]{EAF7F0}\(\mathbf{0.0405{\scriptscriptstyle\pm 0.0046}}\) & \cellcolor[HTML]{EAF7F0}\(\mathbf{0.0512{\scriptscriptstyle\pm 0.0063}}\) & \cellcolor[HTML]{EAF7F0}\(\mathbf{0.0591{\scriptscriptstyle\pm 0.0074}}\) & \cellcolor[HTML]{EAF7F0}\(\mathbf{0.0659{\scriptscriptstyle\pm 0.0083}}\) \\
 & LLM-Filter & \(0.0189{\scriptscriptstyle\pm 0.0019}\) & \(0.0505{\scriptscriptstyle\pm 0.0058}\) & \(0.0696{\scriptscriptstyle\pm 0.0078}\) & \(0.0845{\scriptscriptstyle\pm 0.0092}\) & \(0.0963{\scriptscriptstyle\pm 0.0103}\) \\
 & EnKF & \(0.0165{\scriptscriptstyle\pm 0.0011}\) & \(0.0619{\scriptscriptstyle\pm 0.0039}\) & \(0.0868{\scriptscriptstyle\pm 0.0055}\) & \(0.1061{\scriptscriptstyle\pm 0.0068}\) & \(0.1223{\scriptscriptstyle\pm 0.0078}\) \\
 & EnKFS & \(0.0217{\scriptscriptstyle\pm 0.0043}\) & \(0.0546{\scriptscriptstyle\pm 0.0046}\) & \(0.0706{\scriptscriptstyle\pm 0.0064}\) & \(0.0823{\scriptscriptstyle\pm 0.0080}\) & \(0.0918{\scriptscriptstyle\pm 0.0095}\) \\
 & HubEnKF & \(0.0219{\scriptscriptstyle\pm 0.0041}\) & \(0.0602{\scriptscriptstyle\pm 0.0048}\) & \(0.0778{\scriptscriptstyle\pm 0.0066}\) & \(0.0904{\scriptscriptstyle\pm 0.0083}\) & \(0.1004{\scriptscriptstyle\pm 0.0097}\) \\
\addlinespace[1.5pt]
\textbf{Oscillator} & \cellcolor[HTML]{EAF7F0}FilterFM & \cellcolor[HTML]{EAF7F0}\(\mathbf{0.0166{\scriptscriptstyle\pm 0.0011}}\) & \cellcolor[HTML]{EAF7F0}\(\mathbf{0.0379{\scriptscriptstyle\pm 0.0052}}\) & \cellcolor[HTML]{EAF7F0}\(\mathbf{0.0449{\scriptscriptstyle\pm 0.0083}}\) & \cellcolor[HTML]{EAF7F0}\(\mathbf{0.0490{\scriptscriptstyle\pm 0.0104}}\) & \cellcolor[HTML]{EAF7F0}\(\mathbf{0.0518{\scriptscriptstyle\pm 0.0121}}\) \\
 & LLM-Filter & \(0.0166{\scriptscriptstyle\pm 0.0012}\) & \(0.0412{\scriptscriptstyle\pm 0.0037}\) & \(0.0540{\scriptscriptstyle\pm 0.0062}\) & \(0.0630{\scriptscriptstyle\pm 0.0080}\) & \(0.0699{\scriptscriptstyle\pm 0.0098}\) \\
 & EnKF & \(0.0170{\scriptscriptstyle\pm 0.0010}\) & \(0.0631{\scriptscriptstyle\pm 0.0039}\) & \(0.0883{\scriptscriptstyle\pm 0.0055}\) & \(0.1078{\scriptscriptstyle\pm 0.0067}\) & \(0.1242{\scriptscriptstyle\pm 0.0077}\) \\
 & EnKFS & \(0.0206{\scriptscriptstyle\pm 0.0025}\) & \(0.0554{\scriptscriptstyle\pm 0.0033}\) & \(0.0719{\scriptscriptstyle\pm 0.0044}\) & \(0.0839{\scriptscriptstyle\pm 0.0053}\) & \(0.0936{\scriptscriptstyle\pm 0.0061}\) \\
 & HubEnKF & \(0.0212{\scriptscriptstyle\pm 0.0023}\) & \(0.0610{\scriptscriptstyle\pm 0.0032}\) & \(0.0794{\scriptscriptstyle\pm 0.0044}\) & \(0.0926{\scriptscriptstyle\pm 0.0054}\) & \(0.1032{\scriptscriptstyle\pm 0.0063}\) \\
\addlinespace[1.5pt]
\textbf{Hopf} & \cellcolor[HTML]{EAF7F0}FilterFM & \cellcolor[HTML]{EAF7F0}\(\mathbf{0.0156{\scriptscriptstyle\pm 0.0009}}\) & \cellcolor[HTML]{EAF7F0}\(0.0431{\scriptscriptstyle\pm 0.0072}\) & \cellcolor[HTML]{EAF7F0}\(\mathbf{0.0551{\scriptscriptstyle\pm 0.0110}}\) & \cellcolor[HTML]{EAF7F0}\(\mathbf{0.0626{\scriptscriptstyle\pm 0.0137}}\) & \cellcolor[HTML]{EAF7F0}\(\mathbf{0.0680{\scriptscriptstyle\pm 0.0156}}\) \\
 & LLM-Filter & \(0.0176{\scriptscriptstyle\pm 0.0014}\) & \(\mathbf{0.0427{\scriptscriptstyle\pm 0.0046}}\) & \(0.0562{\scriptscriptstyle\pm 0.0071}\) & \(0.0662{\scriptscriptstyle\pm 0.0092}\) & \(0.0747{\scriptscriptstyle\pm 0.0111}\) \\
 & EnKF & \(0.0164{\scriptscriptstyle\pm 0.0009}\) & \(0.0632{\scriptscriptstyle\pm 0.0033}\) & \(0.0889{\scriptscriptstyle\pm 0.0048}\) & \(0.1087{\scriptscriptstyle\pm 0.0059}\) & \(0.1254{\scriptscriptstyle\pm 0.0069}\) \\
 & EnKFS & \(0.0212{\scriptscriptstyle\pm 0.0038}\) & \(0.0549{\scriptscriptstyle\pm 0.0040}\) & \(0.0715{\scriptscriptstyle\pm 0.0059}\) & \(0.0835{\scriptscriptstyle\pm 0.0074}\) & \(0.0932{\scriptscriptstyle\pm 0.0087}\) \\
 & HubEnKF & \(0.0216{\scriptscriptstyle\pm 0.0037}\) & \(0.0602{\scriptscriptstyle\pm 0.0040}\) & \(0.0787{\scriptscriptstyle\pm 0.0059}\) & \(0.0919{\scriptscriptstyle\pm 0.0074}\) & \(0.1023{\scriptscriptstyle\pm 0.0087}\) \\
\midrule
\textbf{Lorenz96} & \cellcolor[HTML]{EAF7F0}FilterFM & \cellcolor[HTML]{EAF7F0}\(\mathbf{1.170{\scriptscriptstyle\pm 0.026}}\) & \cellcolor[HTML]{EAF7F0}\(3.805{\scriptscriptstyle\pm 0.158}\) & \cellcolor[HTML]{EAF7F0}\(\mathbf{4.339{\scriptscriptstyle\pm 0.150}}\) & \cellcolor[HTML]{EAF7F0}\(\mathbf{4.546{\scriptscriptstyle\pm 0.153}}\) & \cellcolor[HTML]{EAF7F0}\(\mathbf{4.674{\scriptscriptstyle\pm 0.156}}\) \\
 & LLM-Filter & \(2.054{\scriptscriptstyle\pm 0.063}\) & \(\mathbf{3.688{\scriptscriptstyle\pm 0.130}}\) & \(4.376{\scriptscriptstyle\pm 0.159}\) & \(4.759{\scriptscriptstyle\pm 0.173}\) & \(5.011{\scriptscriptstyle\pm 0.180}\) \\
 & EnKF & \(5.311{\scriptscriptstyle\pm 0.243}\) & \(6.311{\scriptscriptstyle\pm 0.257}\) & \(6.842{\scriptscriptstyle\pm 0.245}\) & \(7.440{\scriptscriptstyle\pm 0.284}\) & \(8.105{\scriptscriptstyle\pm 0.444}\) \\
 & EnKFS & \(5.388{\scriptscriptstyle\pm 0.241}\) & \(5.944{\scriptscriptstyle\pm 0.268}\) & \(6.150{\scriptscriptstyle\pm 0.197}\) & \(6.389{\scriptscriptstyle\pm 0.212}\) & \(6.631{\scriptscriptstyle\pm 0.219}\) \\
 & HubEnKF & \(5.345{\scriptscriptstyle\pm 0.298}\) & \(6.003{\scriptscriptstyle\pm 0.258}\) & \(6.394{\scriptscriptstyle\pm 0.202}\) & \(6.753{\scriptscriptstyle\pm 0.219}\) & \(7.175{\scriptscriptstyle\pm 0.281}\) \\
\addlinespace[1.5pt]
\textbf{VL20} & \cellcolor[HTML]{EAF7F0}FilterFM & \cellcolor[HTML]{EAF7F0}\(\mathbf{0.531{\scriptscriptstyle\pm 0.010}}\) & \cellcolor[HTML]{EAF7F0}\(\mathbf{1.717{\scriptscriptstyle\pm 0.032}}\) & \cellcolor[HTML]{EAF7F0}\(\mathbf{2.141{\scriptscriptstyle\pm 0.041}}\) & \cellcolor[HTML]{EAF7F0}\(\mathbf{2.421{\scriptscriptstyle\pm 0.048}}\) & \cellcolor[HTML]{EAF7F0}\(\mathbf{2.644{\scriptscriptstyle\pm 0.052}}\) \\
 & LLM-Filter & \(2.469{\scriptscriptstyle\pm 0.088}\) & \(2.640{\scriptscriptstyle\pm 0.083}\) & \(2.881{\scriptscriptstyle\pm 0.086}\) & \(3.079{\scriptscriptstyle\pm 0.085}\) & \(3.249{\scriptscriptstyle\pm 0.084}\) \\
 & EnKF & \(1.482{\scriptscriptstyle\pm 0.294}\) & \(3.151{\scriptscriptstyle\pm 0.432}\) & \(3.973{\scriptscriptstyle\pm 0.409}\) & \(4.534{\scriptscriptstyle\pm 0.458}\) & \(5.033{\scriptscriptstyle\pm 0.477}\) \\
 & EnKFS & \(1.540{\scriptscriptstyle\pm 0.292}\) & \(2.756{\scriptscriptstyle\pm 0.388}\) & \(3.282{\scriptscriptstyle\pm 0.454}\) & \(3.686{\scriptscriptstyle\pm 0.573}\) & \(3.961{\scriptscriptstyle\pm 0.586}\) \\
 & HubEnKF & \(1.162{\scriptscriptstyle\pm 0.194}\) & \(2.422{\scriptscriptstyle\pm 0.475}\) & \(2.936{\scriptscriptstyle\pm 0.435}\) & \(3.328{\scriptscriptstyle\pm 0.440}\) & \(3.706{\scriptscriptstyle\pm 0.373}\) \\
\addlinespace[1.5pt]
\textbf{LorenzUV40} & \cellcolor[HTML]{EAF7F0}FilterFM & \cellcolor[HTML]{EAF7F0}\(\mathbf{0.122{\scriptscriptstyle\pm 0.006}}\) & \cellcolor[HTML]{EAF7F0}\(\mathbf{0.324{\scriptscriptstyle\pm 0.012}}\) & \cellcolor[HTML]{EAF7F0}\(\mathbf{0.414{\scriptscriptstyle\pm 0.013}}\) & \cellcolor[HTML]{EAF7F0}\(\mathbf{0.479{\scriptscriptstyle\pm 0.015}}\) & \cellcolor[HTML]{EAF7F0}\(\mathbf{0.532{\scriptscriptstyle\pm 0.018}}\) \\
 & LLM-Filter & \(0.254{\scriptscriptstyle\pm 0.039}\) & \(0.781{\scriptscriptstyle\pm 0.134}\) & \(1.209{\scriptscriptstyle\pm 0.171}\) & \(1.446{\scriptscriptstyle\pm 0.190}\) & \(1.609{\scriptscriptstyle\pm 0.221}\) \\
 & EnKF & \(0.321{\scriptscriptstyle\pm 0.036}\) & \(0.820{\scriptscriptstyle\pm 0.197}\) & \(1.324{\scriptscriptstyle\pm 0.321}\) & \(79.867{\scriptscriptstyle\pm 30.754}\) & \(98.758{\scriptscriptstyle\pm 0.434}\) \\
 & EnKFS & \(0.314{\scriptscriptstyle\pm 0.038}\) & \(0.510{\scriptscriptstyle\pm 0.071}\) & \(0.603{\scriptscriptstyle\pm 0.075}\) & \(1.975{\scriptscriptstyle\pm 1.189}\) & \(3.871{\scriptscriptstyle\pm 0.777}\) \\
 & HubEnKF & \(0.313{\scriptscriptstyle\pm 0.037}\) & \(0.514{\scriptscriptstyle\pm 0.067}\) & \(1.590{\scriptscriptstyle\pm 0.950}\) & \(3.825{\scriptscriptstyle\pm 1.190}\) & \(4.196{\scriptscriptstyle\pm 0.041}\) \\
\addlinespace[1.5pt]
\textbf{Vanderpol64} & \cellcolor[HTML]{EAF7F0}FilterFM & \cellcolor[HTML]{EAF7F0}\(\mathbf{0.068{\scriptscriptstyle\pm 0.002}}\) & \cellcolor[HTML]{EAF7F0}\(\mathbf{0.323{\scriptscriptstyle\pm 0.015}}\) & \cellcolor[HTML]{EAF7F0}\(0.511{\scriptscriptstyle\pm 0.017}\) & \cellcolor[HTML]{EAF7F0}\(0.610{\scriptscriptstyle\pm 0.018}\) & \cellcolor[HTML]{EAF7F0}\(0.672{\scriptscriptstyle\pm 0.019}\) \\
 & LLM-Filter & \(0.341{\scriptscriptstyle\pm 0.034}\) & \(0.403{\scriptscriptstyle\pm 0.032}\) & \(\mathbf{0.454{\scriptscriptstyle\pm 0.033}}\) & \(\mathbf{0.497{\scriptscriptstyle\pm 0.033}}\) & \(\mathbf{0.534{\scriptscriptstyle\pm 0.035}}\) \\
 & EnKF & \(0.135{\scriptscriptstyle\pm 0.014}\) & \(0.701{\scriptscriptstyle\pm 0.074}\) & \(0.966{\scriptscriptstyle\pm 0.080}\) & \(1.142{\scriptscriptstyle\pm 0.076}\) & \(1.263{\scriptscriptstyle\pm 0.072}\) \\
 & EnKFS & \(0.136{\scriptscriptstyle\pm 0.013}\) & \(0.522{\scriptscriptstyle\pm 0.040}\) & \(0.698{\scriptscriptstyle\pm 0.054}\) & \(0.818{\scriptscriptstyle\pm 0.066}\) & \(0.918{\scriptscriptstyle\pm 0.074}\) \\
 & HubEnKF & \(0.133{\scriptscriptstyle\pm 0.015}\) & \(0.518{\scriptscriptstyle\pm 0.039}\) & \(0.681{\scriptscriptstyle\pm 0.052}\) & \(0.816{\scriptscriptstyle\pm 0.072}\) & \(0.901{\scriptscriptstyle\pm 0.072}\) \\
\addlinespace[1.5pt]
\textbf{Rossler96} & \cellcolor[HTML]{EAF7F0}FilterFM & \cellcolor[HTML]{EAF7F0}\(\mathbf{0.058{\scriptscriptstyle\pm 0.001}}\) & \cellcolor[HTML]{EAF7F0}\(\mathbf{0.161{\scriptscriptstyle\pm 0.011}}\) & \cellcolor[HTML]{EAF7F0}\(\mathbf{0.185{\scriptscriptstyle\pm 0.016}}\) & \cellcolor[HTML]{EAF7F0}\(\mathbf{0.194{\scriptscriptstyle\pm 0.016}}\) & \cellcolor[HTML]{EAF7F0}\(\mathbf{0.198{\scriptscriptstyle\pm 0.014}}\) \\
 & LLM-Filter & \(0.246{\scriptscriptstyle\pm 0.023}\) & \(0.277{\scriptscriptstyle\pm 0.020}\) & \(0.303{\scriptscriptstyle\pm 0.019}\) & \(0.323{\scriptscriptstyle\pm 0.018}\) & \(0.341{\scriptscriptstyle\pm 0.018}\) \\
 & EnKF & \(0.182{\scriptscriptstyle\pm 0.009}\) & \(0.266{\scriptscriptstyle\pm 0.013}\) & \(0.333{\scriptscriptstyle\pm 0.017}\) & \(0.388{\scriptscriptstyle\pm 0.020}\) & \(0.436{\scriptscriptstyle\pm 0.023}\) \\
 & EnKFS & \(0.188{\scriptscriptstyle\pm 0.009}\) & \(0.268{\scriptscriptstyle\pm 0.015}\) & \(0.327{\scriptscriptstyle\pm 0.019}\) & \(0.375{\scriptscriptstyle\pm 0.023}\) & \(0.415{\scriptscriptstyle\pm 0.026}\) \\
 & HubEnKF & \(0.183{\scriptscriptstyle\pm 0.009}\) & \(0.269{\scriptscriptstyle\pm 0.015}\) & \(0.334{\scriptscriptstyle\pm 0.020}\) & \(0.387{\scriptscriptstyle\pm 0.024}\) & \(0.432{\scriptscriptstyle\pm 0.027}\) \\
\addlinespace[1.5pt]
\textbf{Cubic-Sensoring100} & \cellcolor[HTML]{EAF7F0}FilterFM & \cellcolor[HTML]{EAF7F0}\(\mathbf{0.141{\scriptscriptstyle\pm 0.005}}\) & \cellcolor[HTML]{EAF7F0}\(\mathbf{0.206{\scriptscriptstyle\pm 0.006}}\) & \cellcolor[HTML]{EAF7F0}\(\mathbf{0.229{\scriptscriptstyle\pm 0.007}}\) & \cellcolor[HTML]{EAF7F0}\(\mathbf{0.245{\scriptscriptstyle\pm 0.008}}\) & \cellcolor[HTML]{EAF7F0}\(\mathbf{0.260{\scriptscriptstyle\pm 0.010}}\) \\
 & LLM-Filter & \(0.728{\scriptscriptstyle\pm 0.069}\) & \(0.728{\scriptscriptstyle\pm 0.069}\) & \(0.728{\scriptscriptstyle\pm 0.069}\) & \(0.729{\scriptscriptstyle\pm 0.069}\) & \(0.729{\scriptscriptstyle\pm 0.069}\) \\
 & EnKF & \(0.150{\scriptscriptstyle\pm 0.004}\) & \(0.341{\scriptscriptstyle\pm 0.009}\) & \(0.431{\scriptscriptstyle\pm 0.011}\) & \(0.495{\scriptscriptstyle\pm 0.013}\) & \(0.547{\scriptscriptstyle\pm 0.013}\) \\
 & EnKFS & \(0.328{\scriptscriptstyle\pm 0.044}\) & \(0.457{\scriptscriptstyle\pm 0.069}\) & \(0.498{\scriptscriptstyle\pm 0.083}\) & \(0.507{\scriptscriptstyle\pm 0.068}\) & \(0.516{\scriptscriptstyle\pm 0.058}\) \\
 & HubEnKF & \(0.344{\scriptscriptstyle\pm 0.053}\) & \(0.519{\scriptscriptstyle\pm 0.067}\) & \(0.523{\scriptscriptstyle\pm 0.061}\) & \(0.533{\scriptscriptstyle\pm 0.078}\) & \(0.529{\scriptscriptstyle\pm 0.060}\) \\
\bottomrule
\end{tabular}%
}
\caption{Model-mismatch raw-space RMSE on the 20 held-out testing trajectories, reported as pooled RMSE ${\scriptscriptstyle\pm}$ the population standard deviation of trajectory-wise RMSE. FilterFM rows are shaded. Lower is better, and the lowest pooled RMSE for each dataset and OCER is bolded.}
\label{tab:model_mismatch_details}
\end{table}

\subsection{Additional Ablation and Scaling Results}
\label{app:additional_ablation_scaling}

We first examine which architectural components contribute to state estimation, then study the prior-loss weight, system diversity, model capacity, trajectory coverage, and patch length. The ablations share the main training corpus and optimization settings; the scaling experiments use the larger Panda-derived corpora described below.

\subsubsection{Architectural Ablations}
\label{app:architectural_ablations}

\begin{figure}[!ht]
\centering
\includegraphics[width=0.96\linewidth]{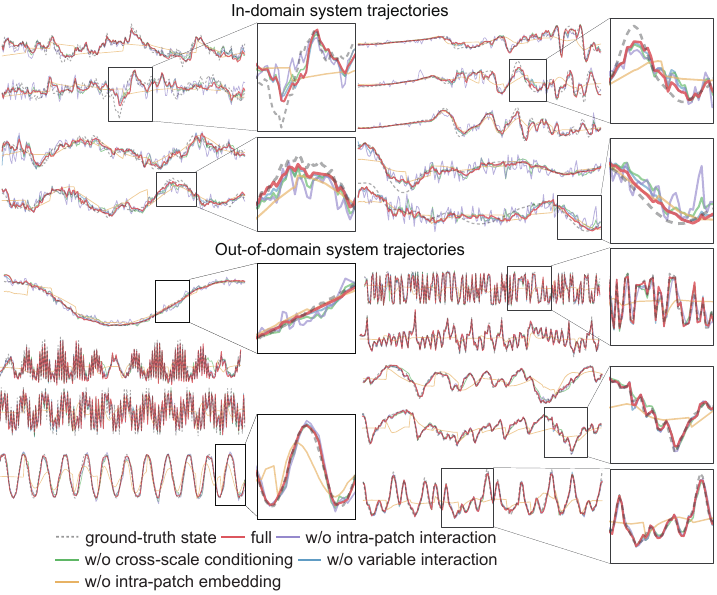}
\caption{Ablation trajectories. In-domain and zero-shot system trajectories across five architectural variants.}
\label{fig:ablation_trajectories}
\end{figure}

Figure~\ref{fig:ablation_trajectories} illustrates how the error differences in Figure~\ref{fig:ablation_dynamics} appear in individual trajectories. Removing intra-patch embeddings produces the largest median error increase on both in-domain and zero-shot systems. This variant estimates the current state from past patch representations without direct access to the current observation, which can leave substantial offsets and miss changes within a patch. Keeping the embeddings but removing intra-patch interaction gives the second-largest degradation: per-step observations remain available, but the model loses the local processing that combines them into a temporally coherent representation. The higher final training losses of these two variants indicate that their limitations are already visible when fitting the training tasks. These results suggest that Bayesian filtering accuracy depends strongly on incorporating current observations and that neglecting short-horizon temporal dependencies can substantially degrade performance.

Cross-scale conditioning uses historical patch summaries to help interpret noisy current observations, while variable interaction combines information across observation channels. Removing either module increases median RMSE on both in-domain and zero-shot systems, with a larger increase when cross-scale conditioning is removed. These increases are smaller than those caused by removing intra-patch embeddings or intra-patch interaction. This suggests that, for these filtering tasks, using recent observations and modeling short-horizon temporal dependencies contribute more to state-estimation accuracy than additional historical or cross-channel context.

\subsubsection{Prior-to-Posterior Updates and Prior-Loss Weight}
\label{app:prior_loss_weight}

Table~\ref{tab:prior_weight_details} compares direct posterior reconstruction with explicit prior-to-posterior updates. The direct variant reconstructs the posterior from current local memory without forming a prior. The In-domain raw column averages raw RMSE over the fourteen benchmark systems, with 20 test trajectories per system. High-dim raw reports the same metric for the nine systems with 40--100 state variables in Table~\ref{tab:selected14_system_configs}, using the same 20 test trajectories per system for a total of 180. The OOS raw column evaluates transfer on 8,207 held-out Panda trajectories using a 200-step window per trajectory.

Every tested prior-loss weight improves all three scores relative to the direct update. A weight of 1.0 reduces OOS RMSE from 1.1055 to 1.0660, a relative improvement of \(3.57\%\), and gives the best OOS result in the sweep. Increasing the weight to 1.8 gives the best in-domain scores but does not improve OOS accuracy. Our main model uses \(\lambda_{\rm prior}=1.0\) because this setting achieves the lowest OOS raw RMSE in the sweep. This comparison suggests that supervising a prediction from past evidence provides a useful basis for incorporating the current observation.

\begin{table}[!ht]
\centering
\footnotesize
\renewcommand{\arraystretch}{1.12}
\setlength{\tabcolsep}{8.0pt}
\begin{tabular}{@{}lcrrr@{}}
\toprule
\textbf{Update rule} & \(\boldsymbol{\lambda_{\rm prior}}\) & \textbf{In-domain raw} & \textbf{High-dim raw} & \textbf{OOS raw} \\
\midrule
Direct posterior & -- & 0.2351 & 0.2428 & 1.1055 \\
Prior-to-posterior & 0.05 & 0.2317 & 0.2413 & 1.0982 \\
Prior-to-posterior & 0.20 & 0.2334 & 0.2419 & 1.1021 \\
Prior-to-posterior & 0.50 & 0.2313 & 0.2403 & 1.0953 \\
Prior-to-posterior & 1.00 & 0.2317 & 0.2390 & \textbf{1.0660} \\
Prior-to-posterior & 1.50 & 0.2321 & 0.2396 & 1.0786 \\
Prior-to-posterior & 1.80 & \textbf{0.2296} & \textbf{0.2375} & 1.0664 \\
Prior-to-posterior & 2.00 & 0.2311 & 0.2396 & 1.0763 \\
\bottomrule
\end{tabular}
\caption{Prior-loss sweep and direct-posterior ablation. In-domain raw averages over all fourteen benchmark systems; High-dim raw averages over the subset with state dimension \(\ge 40\); OOS raw averages over zero-shot systems. Lower is better.}
\label{tab:prior_weight_details}
\end{table}

\subsubsection{Scaling the Diversity of Training Systems}

We use the Panda dataset\footnote{\begingroup\hypersetup{urlcolor=black}\url{https://huggingface.co/datasets/GilpinLab/scalinglaw}\endgroup} to construct eight training subsets with progressively broader coverage of dynamical systems. We generate observations by adding independent Gaussian noise to each state trajectory at every time step, following the protocol in Appendix~\ref{app:training_corpus}. Across these subsets, the nominal system count ranges from 164 to 20,980, while the initial-condition count (IC), denoting the target number of trajectories per system, decreases from 128 to 1 to keep the total trajectory budget approximately matched. We independently initialize FilterFM for each subset and train it from scratch using the same architecture and 10-epoch schedule.

Table~\ref{tab:ic_scaling_details} lists the nominal system counts and the actual training-trajectory counts, which range from 19,101 to 20,702. The original dataset contains trajectories of 4,096 time steps, which we slice into 200-step windows for training and evaluation. Each trajectory yields 21 windows, including an end-aligned final window that overlaps the preceding window by 104 steps. Evaluation uses the same 172,599 windows from 8,219 held-out Panda trajectories, and Figure~\ref{fig:scaling}B summarizes normalized RMSE across these windows. The empirical fits are
\[
\begin{aligned}
\mathrm{RMSE}_{\rm med}(N_{\rm sys})
&=0.1961+1.0957N_{\rm sys}^{-0.5496},\; R^2=0.9976,\\
\mathrm{RMSE}_{\rm mean}(N_{\rm sys})
&=0.2346+1.7555N_{\rm sys}^{-0.6227},\; R^2=0.9977.
\end{aligned}
\]
With approximately matched training-trajectory counts, broader dynamical coverage lowers both mean and median OOS RMSE. The gains diminish at larger system counts.

\begin{table}[!ht]
\centering
\scriptsize
\setlength{\tabcolsep}{7.0pt}
\resizebox{0.88\linewidth}{!}{%
\begin{tabular}{@{}rrrrr@{}}
\toprule
\textbf{IC} & \(\boldsymbol{N_{\rm sys}}\) & \textbf{Train trajectories} & \textbf{OOS normalized median} & \textbf{OOS normalized mean} \\
\midrule
128 & 164 & 19,101 & 0.2624 & 0.3086 \\
64 & 328 & 19,806 & 0.2417 & 0.2804 \\
32 & 656 & 19,672 & 0.2270 & 0.2661 \\
16 & 1312 & 19,534 & 0.2174 & 0.2554 \\
8 & 2623 & 19,369 & 0.2102 & 0.2479 \\
4 & 5245 & 19,357 & 0.2050 & 0.2426 \\
2 & 10490 & 19,349 & 0.2051 & 0.2417 \\
1 & 20980 & 20,702 & \textbf{0.1996} & \textbf{0.2366} \\
\bottomrule
\end{tabular}
}
\caption{System-diversity scaling details. IC is the target number of initial conditions per system, and \(N_{\rm sys}\) is the nominal system count of the corresponding training subset. The actual training-trajectory counts reflect data filtering and validation allocation. Values are OOS normalized RMSE statistics over 200-step windows sliced from 4096-step held-out trajectories. Lower is better.}
\label{tab:ic_scaling_details}
\end{table}

\subsubsection{Scaling Model Capacity}

To examine how model capacity affects generalization with the training data held fixed, we train FilterFM models with six different parameter counts on an expanded Panda-derived corpus\footnote{\begingroup\hypersetup{urlcolor=black}\url{https://huggingface.co/datasets/GilpinLab/skew-mixedp-ic16}\endgroup} containing 301,873 training trajectories from 20,471 systems, with 347 additional trajectories reserved for validation. The systems have three to six state variables and up to sixteen initial conditions each. We construct filtering tasks by adding independent Gaussian observation noise at each time step of the 4,096-step state trajectories and slicing the paired sequences into 200-step windows. All models share a 10-epoch training schedule, optimization settings, quantile levels \(\mathcal{Q}=\{0.1,0.5,0.9\}\), and prior-loss weight \(\lambda_{\mathrm{prior}}=1.0\).

Table~\ref{tab:model_parameter_scaling_architecture} specifies the model widths and depths. The 2.89M--34.54M models increase the hidden width \(d\) and FFN hidden width while keeping block counts fixed and using four heads for the encoder's temporal self-attention and self-attention among prototypes. The 51.21M model retains the widths of the 34.54M model, uses six heads for these self-attention operations, and adds one patch-summary temporal encoding layer, one patch-level block, one intra-patch block, and one prior-to-posterior state-reconstruction block.

We evaluate each model's epoch-10 checkpoint on the same 172,599 windows from 8,219 held-out trajectories. For each trajectory, raw RMSE is computed over all time steps and valid state coordinates across its windows, then summarized by the mean and median across trajectories. Table~\ref{tab:model_parameter_scaling_results} and Figure~\ref{fig:scaling}C show that increasing capacity from 2.89M to 51.21M parameters lowers training loss, in-domain validation RMSE, and OOS error. The OOS mean decreases from 0.8600 to 0.8104 and the median from 0.2189 to 0.2104, indicating better generalization to unseen systems.

\begin{table}[!ht]
\centering
\scriptsize
\renewcommand{\arraystretch}{1.12}
\setlength{\tabcolsep}{3.0pt}
\resizebox{0.82\linewidth}{!}{%
\begin{tabular}{@{}lrrrr@{}}
\toprule
\textbf{Parameters} & \shortstack{\textbf{Training loss}\\\textbf{at epoch 10}} & \shortstack{\textbf{In-domain validation}\\\textbf{raw RMSE}} & \multicolumn{2}{c}{\textbf{OOS raw RMSE}} \\
\cmidrule(lr){4-5}
 & & & \textbf{Mean} & \textbf{Median} \\
\midrule
2.89M  & 0.148945 & 0.284954 & 0.860030 & 0.218894 \\
5.09M  & 0.143718 & 0.277112 & 0.847594 & 0.216160 \\
11.36M & 0.135006 & 0.266257 & 0.833914 & 0.213774 \\
20.11M & 0.129503 & 0.263754 & 0.829695 & 0.212346 \\
34.54M & 0.125071 & 0.257718 & 0.817275 & 0.210798 \\
51.21M & \textbf{0.119767} & \textbf{0.254317} & \textbf{0.810427} & \textbf{0.210380} \\
\bottomrule
\end{tabular}%
}
\caption{Model-parameter scaling training and raw-space evaluation results. All results use the same data, training budget, and epoch-10 checkpoint selection and are from single-seed runs. Lower is better; the best value in each metric column is bolded.}
\label{tab:model_parameter_scaling_results}
\end{table}

\begin{table}[!ht]
\centering
\scriptsize
\renewcommand{\arraystretch}{1.0}
\setlength{\tabcolsep}{3.0pt}
\setlength{\abovecaptionskip}{4pt}
\newcommand{\architectureheader}[1]{\begin{tabular}[t]{@{}c@{}}#1\end{tabular}}
\resizebox{\linewidth}{!}{%
\begin{tabular}{@{}lrrrrrrr@{}}
\toprule
\architectureheader{\textbf{Parameters}} & \architectureheader{\textbf{Hidden}\\\textbf{width} \(\boldsymbol{d}\)} & \architectureheader{\textbf{FFN}\\\textbf{hidden width}} & \architectureheader{\textbf{Self-attention}\\\textbf{heads}} & \architectureheader{\textbf{Patch-summary}\\\textbf{temporal encoding}\\\textbf{layers} \(\boldsymbol{J}_{\rm enc}\)} & \architectureheader{\textbf{Patch-level}\\\textbf{blocks}} & \architectureheader{\textbf{Intra-patch}\\\textbf{blocks}} & \architectureheader{\textbf{Prior-to-posterior}\\\textbf{state-reconstruction}\\\textbf{blocks}} \\
\midrule
2,888,007  & 96  & 384  & 4 & 2 & 3 & 2 & 1 \\
5,091,751  & 128 & 512  & 4 & 2 & 3 & 2 & 1 \\
11,360,871 & 192 & 768  & 4 & 2 & 3 & 2 & 1 \\
20,112,167 & 256 & 1024 & 4 & 2 & 3 & 2 & 1 \\
34,541,847 & 336 & 1344 & 4 & 2 & 3 & 2 & 1 \\
51,205,431 & 336 & 1344 & 6 & 3 & 4 & 3 & 2 \\
\bottomrule
\end{tabular}%
}
\caption{Architecture configurations for model-capacity scaling. Patch-summary temporal encoding layers process observation tokens before pooling, as described in Appendix~\ref{app:method_patch_summaries}. Patch-level blocks combine causal temporal interaction with prototype-based variable interaction; intra-patch blocks additionally apply cross-scale conditioning between these operations, as detailed in Appendices~\ref{app:method_temporal_variable}--\ref{app:method_cross_scale}. Prior-to-posterior state-reconstruction blocks follow Section~\ref{sec:method_belief} and Appendix~\ref{app:method_state_reconstruction}. The listed head count applies to temporal self-attention for patch-summary construction, patch-level and intra-patch temporal self-attention, and self-attention among prototypes at both levels. Gated reassembly and the Retrieve operators use single-head cross-attention.}
\label{tab:model_parameter_scaling_architecture}
\end{table}

\subsubsection{Scaling Training-Trajectory Coverage}
\label{app:trajectory_scaling}
To study whether denser trajectory coverage improves generalization at fixed system diversity, we keep the 20.11M-parameter FilterFM and the same 20,471 training systems while varying the maximum number of trajectories per system over \(\{1,2,4,8,12,16\}\). This produces six datasets containing 20,471 to 301,873 training trajectories. Each configuration is trained independently for 10 epochs across three seeds, with the same validation set, optimization settings, and evaluation protocol as the model-capacity scaling experiment. Table~\ref{tab:trajectory_count_scaling} reports the mean and sample standard deviation of each RMSE statistic across seeds.

\begin{figure}[!ht]
\centering
\includegraphics[width=0.715\linewidth]{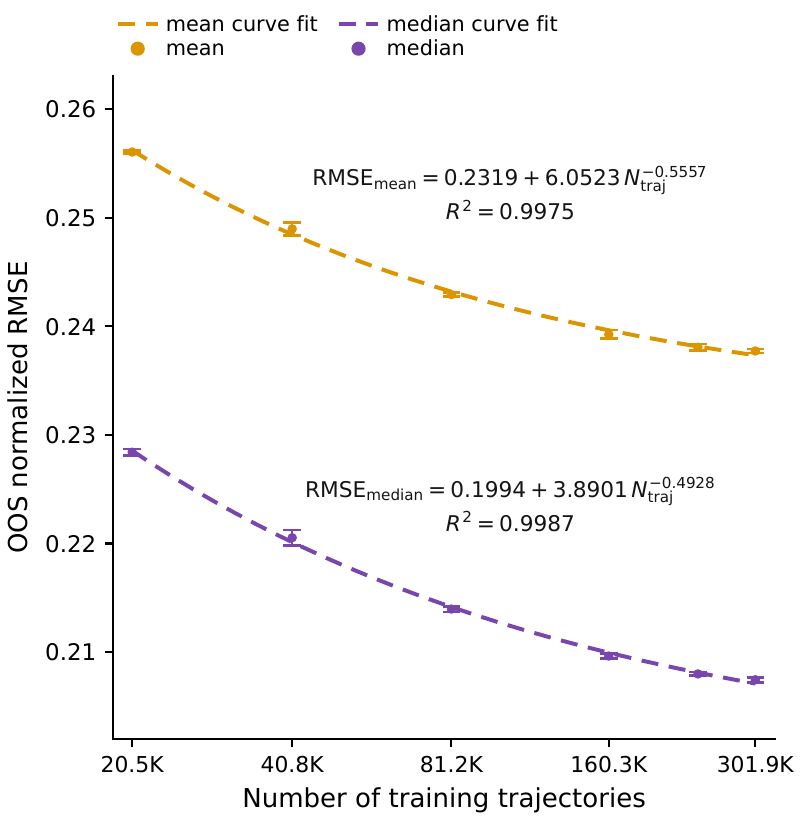}
\caption{Training-trajectory scaling at fixed model capacity and system diversity. Points show OOS trajectory-level mean or median RMSE averaged over three seeds, with vertical error bars indicating one sample standard deviation. Dashed curves show power-law-with-floor fits.}
\label{fig:trajectory_count_scaling}
\end{figure}

\begin{table}[!ht]
\centering
\footnotesize
\renewcommand{\arraystretch}{1.13}
\setlength{\tabcolsep}{4.0pt}
\begin{tabular}{@{}crcccc@{}}
\toprule
\textbf{Maximum} & \multicolumn{1}{c}{\textbf{Training}} & \multicolumn{2}{c}{\textbf{OOS normalized RMSE}} & \multicolumn{2}{c}{\textbf{OOS raw RMSE}} \\
\cmidrule(lr){3-4}\cmidrule(l){5-6}
\textbf{trajectories} & \multicolumn{1}{c}{\textbf{trajectories}} & \textbf{Mean} & \textbf{Median} & \textbf{Mean} & \textbf{Median} \\
\textbf{per system} & & & & & \\
\midrule
1 & 20,471 & \(0.25605{\scriptscriptstyle\pm 0.00016}\) & \(0.22841{\scriptscriptstyle\pm 0.00029}\) & \(0.90963{\scriptscriptstyle\pm 0.00142}\) & \(0.23163{\scriptscriptstyle\pm 0.00017}\) \\
2 & 40,845 & \(0.24897{\scriptscriptstyle\pm 0.00061}\) & \(0.22054{\scriptscriptstyle\pm 0.00072}\) & \(0.87511{\scriptscriptstyle\pm 0.00531}\) & \(0.22380{\scriptscriptstyle\pm 0.00081}\) \\
4 & 81,239 & \(0.24292{\scriptscriptstyle\pm 0.00019}\) & \(0.21397{\scriptscriptstyle\pm 0.00026}\) & \(0.84550{\scriptscriptstyle\pm 0.00407}\) & \(0.21733{\scriptscriptstyle\pm 0.00036}\) \\
8 & 160,304 & \(0.23925{\scriptscriptstyle\pm 0.00040}\) & \(0.20964{\scriptscriptstyle\pm 0.00021}\) & \(0.83187{\scriptscriptstyle\pm 0.00150}\) & \(0.21345{\scriptscriptstyle\pm 0.00036}\) \\
12 & 235,557 & \(0.23807{\scriptscriptstyle\pm 0.00032}\) & \(0.20799{\scriptscriptstyle\pm 0.00016}\) & \(\mathbf{0.82480{\scriptscriptstyle\pm 0.00333}}\) & \(0.21252{\scriptscriptstyle\pm 0.00026}\) \\
16 & 301,873 & \(\mathbf{0.23773{\scriptscriptstyle\pm 0.00020}}\) & \(\mathbf{0.20745{\scriptscriptstyle\pm 0.00023}}\) & \(0.82621{\scriptscriptstyle\pm 0.00731}\) & \(\mathbf{0.21234{\scriptscriptstyle\pm 0.00032}}\) \\
\bottomrule
\end{tabular}
\caption{Training-trajectory scaling across three seeds. Entries report the mean ${\scriptscriptstyle\pm}$ sample standard deviation of each OOS trajectory statistic. All runs use the same 20,471 training systems and 20.11M-parameter model. Lower is better; the best average in each metric column is bolded.}
\label{tab:trajectory_count_scaling}
\end{table}

Figure~\ref{fig:trajectory_count_scaling} and Table~\ref{tab:trajectory_count_scaling} show that increasing trajectory coverage improves OOS estimation when the set of training systems is fixed. Across the six dataset sizes, the seed-averaged normalized mean and median RMSE decrease monotonically from 0.25605 to 0.23773 and from 0.22841 to 0.20745, respectively, corresponding to relative reductions of \(7.16\%\) and \(9.17\%\), with diminishing gains at larger sizes.

\subsubsection{Patch-Length Sensitivity}

We vary the temporal patch length \(P\) while keeping the 200-step window, training corpus, all other architectural hyperparameters, quantile objective, prior-loss weight (\(\lambda_{\rm prior}=1.0\)), and 20-epoch optimization budget fixed. Each result uses the checkpoint selected by validation raw RMSE.

\begin{table}[!ht]
\centering
\footnotesize
\renewcommand{\arraystretch}{1.10}
\setlength{\tabcolsep}{5.0pt}
\begin{tabular}{@{}crrrrr@{}}
\toprule
\textbf{Patch length} \(\boldsymbol{P}\) & \textbf{Patches / window} & \multicolumn{2}{c}{\textbf{In-domain RMSE}} & \multicolumn{2}{c}{\textbf{OOS RMSE}} \\
\cmidrule(lr){3-4}\cmidrule(l){5-6}
& & \textbf{Raw} & \textbf{Normalized} & \textbf{Raw} & \textbf{Normalized} \\
\midrule
1  & 200 & 0.2669 & 0.22673 & 1.1814 & 0.22043 \\
5  & 40  & \textbf{0.2278} & 0.20818 & 1.0674 & \textbf{0.20794} \\
10 & 20  & 0.2281 & \textbf{0.20760} & \textbf{1.0548} & 0.20802 \\
\rowcolor{black!6}
20 & 10  & 0.2317 & 0.20948 & 1.0660 & 0.20961 \\
40 & 5   & 0.2329 & 0.20762 & 1.0825 & 0.21087 \\
\bottomrule
\end{tabular}
\setlength{\abovecaptionskip}{4pt}
\caption{Patch-length sensitivity of FilterFM. In-domain values are trajectory-level means over the 280 test trajectories from the 14 main benchmark systems; OOS values are trajectory-level means over 8,207 held-out zero-shot trajectories. All runs use a single seed, predict the \(0.1/0.5/0.9\) quantiles, and use one prior-to-posterior state-reconstruction block. The shaded row is the \(P=20\) setting used in the main experiments. Lower is better.}
\label{tab:patch_length_sensitivity}
\end{table}

Relative to the best result at \(P>1\) for each metric, \(P=1\) increases in-domain raw and normalized RMSE by \(17.2\%\) and \(9.2\%\), and OOS raw and normalized RMSE by \(12.0\%\) and \(6.0\%\), respectively. This trend is consistent with the gains from increasing input patch length reported for the time-series foundation model TimesFM~\citep{das2024decoder}. At \(P=1\), patch-level attention still models dependencies across the observed history, but temporal attention in patch-summary construction and intra-patch processing reduces to single-token operations. Short-term variations and longer-range dependencies therefore rely on the same patch-level temporal attention.

OOS errors rise at both ends of the tested patch-length range. Our main model uses \(P=20\).

\begingroup
\raggedbottom
\section{Additional Method Details}
\label{app:method_details}

We detail the implementation of our model, FilterFM, including observation encoding, prototype-based variable interaction, prior-to-posterior state reconstruction, and the training objective. We follow the notation in Section~\ref{sec:method}, with hidden width \(d\), and omit block indices, projection biases, dropout, and pre-layer normalization where they do not affect the explanation.

\subsection{Representing Individual Observations}
\label{app:method_observation_tokens}
The encoder begins with one token for each observation channel and time step. For \(t=pL_p+r\), with patch index \(p\ge0\) and within-patch position \(r\in\{1,\ldots,L_p\}\), we form
\[
\left\{\begin{aligned}
&\mathbf{c}_{p,r,m}=[\tilde{y}_{t,m};a_m;\rho_r;\tau_t]\in\mathbb{R}^{4},\\
&\mathbf{o}_{p,r,m}={\rm ObsEmb}(\mathbf{c}_{p,r,m})+\mathbf{e}^y_m+\mathbf{e}^{\rm patch}_p+\mathbf{e}^{\rm pos}_r\in\mathbb{R}^{d}.
\end{aligned}\right.
\]
Here \(\tilde{y}_{t,m}\in\mathbb{R}\) is the normalized observation, and \(a_m\in\{0,1\}\) indicates whether observation channel \(m\) is valid for the current system. The scalars \(\rho_r=(r-1)/\max(L_p-1,1)\) and \(\tau_t=(t-1)/\max(T-1,1)\) lie in \([0,1]\) and encode relative positions within the patch and the full window. The shared MLP \({\rm ObsEmb}:\mathbb{R}^{4}\to\mathbb{R}^{d}\) embeds these features. The learned vectors \(\mathbf{e}^y_m,\mathbf{e}^{\rm patch}_p,\mathbf{e}^{\rm pos}_r\in\mathbb{R}^{d}\) encode observation-channel identity, patch index, and within-patch position, respectively.

\subsection{Summarizing Completed Patches}
\label{app:method_patch_summaries}
Before forming patch summaries, causal temporal encoding processes each channel's observation tokens within the patch. Let \(J_{\rm enc}\) be the number of patch-summary temporal encoding layers and set \(\mathbf{o}^{(0)}_{p,r,m}=\mathbf{o}_{p,r,m}\). For \(j=1,\ldots,J_{\rm enc}\), the updates and final encoded token are
\[
\left\{\begin{aligned}
&\mathbf{h}^{(j)}_{p,r,m}=\mathbf{o}^{(j-1)}_{p,r,m}+{\rm SelfAttn}_j(\mathbf{o}^{(j-1)}_{p,1:r,m}),\\
&\mathbf{o}^{(j)}_{p,r,m}=\mathbf{h}^{(j)}_{p,r,m}+{\rm FFN}_j(\mathbf{h}^{(j)}_{p,r,m}),\\
&\tilde{\mathbf{o}}_{p,r,m}=\mathbf{o}^{(J_{\rm enc})}_{p,r,m}\in\mathbb{R}^{d}.
\end{aligned}\right.
\]
Here \(\mathbf{h}^{(j)}_{p,r,m},\mathbf{o}^{(j)}_{p,r,m}\in\mathbb{R}^{d}\), and attention is causal within each patch. Our main model uses \(J_{\rm enc}=2\).

We then combine three components into the completed-patch summary \(\mathbf{s}_{p,m}\):
\[
\left\{\begin{aligned}
&\mathbf{s}^{\rm raw}_{p,m}=\mathbf{W}_{\rm lin}\,\mathrm{vec}(\mathbf{c}_{p,1:L_p,m}),\\
&\pi_{p,r,m}={\rm softmax}_r\!\left(\frac{\mathbf{q}_{\rm pool}^{\top}\mathbf{W}_{\rm pool,k}\tilde{\mathbf{o}}_{p,r,m}}{\sqrt{d}}\right),\\
&\mathbf{s}^{\rm pool}_{p,m}={\rm AttentivePool}(\tilde{\mathbf{o}}_{p,1:L_p,m})
=\sum_{r=1}^{L_p}\pi_{p,r,m}\mathbf{W}_{\rm pool,v}\tilde{\mathbf{o}}_{p,r,m},\\
&\mathbf{s}^{\rm last}_{p,m}=\mathbf{W}_{\rm last}\tilde{\mathbf{o}}_{p,L_p,m},\\
&\mathbf{s}_{p,m}={\rm LayerNorm}\big(\mathbf{s}^{\rm raw}_{p,m}+\mathbf{s}^{\rm pool}_{p,m}+\mathbf{s}^{\rm last}_{p,m}\big).
\end{aligned}\right.
\]
The operator \(\mathrm{vec}\) concatenates the four-dimensional vectors \(\mathbf{c}_{p,1,m},\ldots,\mathbf{c}_{p,L_p,m}\) in temporal order into a vector in \(\mathbb{R}^{4L_p}\). The learned projections \(\mathbf{W}_{\rm lin}\in\mathbb{R}^{d\times4L_p}\) and \(\mathbf{W}_{\rm last}\in\mathbb{R}^{d\times d}\) contain \(4L_pd\) and \(d^2\) weights, respectively, each with an additional \(d\)-dimensional bias omitted above. At \(L_p=20\) and \(d=256\), these are 20,736 and 65,792 parameters including biases.

Attentive pooling uses a learned query \(\mathbf{q}_{\rm pool}\in\mathbb{R}^{d}\) and learned key and value projections \(\mathbf{W}_{\rm pool,k},\mathbf{W}_{\rm pool,v}\in\mathbb{R}^{d\times d}\). Its scalar weights \(\pi_{p,r,m}\in[0,1]\) sum to one over the \(L_p\) positions. Thus, \(\mathbf{s}^{\rm raw}_{p,m}\) preserves position-specific raw features through concatenation, \(\mathbf{s}^{\rm pool}_{p,m}\) weights encoded positions by their learned relevance, and \(\mathbf{s}^{\rm last}_{p,m}\) projects the final token, whose causal encoding covers the completed patch. All three vectors and their normalized sum \(\mathbf{s}_{p,m}\) belong to \(\mathbb{R}^{d}\).

\subsection{Temporal and Variable Interaction}
\label{app:method_temporal_variable}

The encoder combines causal temporal interaction with prototype-based variable interaction at both levels. Following Section~\ref{sec:method_encoder}, \(\ell\in\{\mathrm{patch},\mathrm{intra}\}\) denotes the encoding level, with \(s=p\) for patches and \(s=(p,r)\) for intra-patch positions. Block indices are omitted, and the symbols refer to the block under consideration, with independently learned parameters across blocks. For channel \(m\), the column vectors \(\mathbf{u}^\ell_{s,m},\bar{\mathbf{u}}^\ell_{s,m},\mathbf{u}^{\ell,{\rm out}}_{s,m}\in\mathbb{R}^{d}\) denote its input, temporal output, and full block output, respectively. Their transposes form the \(m\)-th rows of \(\mathbf{U}^\ell_s,\bar{\mathbf{U}}^\ell_s,\mathbf{U}^{\ell,{\rm out}}_s\in\mathbb{R}^{M\times d}\). Temporal interaction follows Eq.~(\ref{eq:method_temporal_block}): patch-level attention accesses summaries up to patch \(p\), while intra-patch attention accesses positions up to \(r\) within the current patch.

Prototype-based variable interaction exchanges information across observation channels through \(C\) shared slots. Its input \(\mathbf{v}^\ell_{s,m}\in\mathbb{R}^{d}\) is the representation of observation channel \(m\) in \(\mathbf{V}^\ell_s\in\mathbb{R}^{M\times d}\). At the patch level, \(\mathbf{V}^{\rm patch}_p=\bar{\mathbf{U}}^{\rm patch}_p\). At the intra-patch level, the temporal output first incorporates historical context through the cross-scale conditioning described in Appendix~\ref{app:method_cross_scale}. The learned matrices \(\mathbf{A}^\ell_q,\mathbf{A}^\ell_v\in\mathbb{R}^{d\times d}\) project these representations to queries and values, while the rows \(\mathbf{k}^{\ell,\pm}_c\in\mathbb{R}^{d}\) of \(\mathbf{K}^\ell_\pm\in\mathbb{R}^{C\times d}\) provide positive and negative attention keys:
\[
\left\{\begin{aligned}
&\alpha^{\ell,\pm}_{s,m,c}={\rm softmax}_c\!\left(\frac{(\mathbf{A}^\ell_q\mathbf{v}^{\ell}_{s,m})^\top\mathbf{k}^{\ell,\pm}_c}{\sqrt{d}}\right),\\
&\mathbf{p}^\ell_{s,c}=\frac{1}{\sum_{m'=1}^{M}a_{m'}}
\sum_{m'=1}^{M}a_{m'}\big(\alpha^{\ell,+}_{s,m',c}-\lambda_\ell\alpha^{\ell,-}_{s,m',c}\big)
\mathbf{A}^\ell_v\mathbf{v}^{\ell}_{s,m'}.
\end{aligned}\right.
\]
The query and value projections \(\mathbf{A}^\ell_q\) and \(\mathbf{A}^\ell_v\) are learned independently in each block and are separate from the temporal-attention, prototype-reassembly, and \({\rm Retrieve}\) projections. Within each block, positive and negative attention share these query and value projections but use distinct keys \(\mathbf{K}^\ell_+\) and \(\mathbf{K}^\ell_-\). The scalar weights \(\alpha^{\ell,+}_{s,m,c},\alpha^{\ell,-}_{s,m,c}\in[0,1]\) each sum to one over \(c\), and \(\lambda_\ell\in\mathbb{R}\) controls the contribution of negative attention, with \(\lambda_\ell=0.5\) in our main model. The keys \(\mathbf{K}^\ell_\pm\) are learned parameters shared across input sequences; the prototypes \(\mathbf{p}^\ell_{s,c}\in\mathbb{R}^{d}\) are computed from the current representations \(\mathbf{v}^\ell_{s,m}\). The indicators \(a_m\) exclude invalid channels, and division by \(\sum_m a_m\) averages over valid channels. When all \(M\) channels are valid, this is exactly the aggregation in Eq.~(\ref{eq:method_diffattn}).

After obtaining \(\mathbf{p}^\ell_{s,c}\), we stack the \(C\) vectors as rows of \(\mathbf{P}^\ell_s\in\mathbb{R}^{C\times d}\). Self-attention across these prototypes, followed by a residual FFN, gives
\[
\left\{\begin{aligned}
&\bar{\mathbf{P}}^\ell_s=\mathbf{P}^\ell_s+{\rm SelfAttn}_\ell(\mathbf{P}^\ell_s),\\
&\tilde{\mathbf{P}}^\ell_s=\bar{\mathbf{P}}^\ell_s+{\rm FFN}_\ell(\bar{\mathbf{P}}^\ell_s),
\end{aligned}\right.
\]
where \(\bar{\mathbf{P}}^\ell_s,\tilde{\mathbf{P}}^\ell_s\in\mathbb{R}^{C\times d}\). These operations mix prototypes at the same temporal position \(s\). Writing the rows of \(\tilde{\mathbf{P}}^\ell_s\) as \(\tilde{\mathbf{p}}^\ell_{s,c}\in\mathbb{R}^{d}\), gated reassembly in Eq.~(\ref{eq:method_reassemble}) updates each variable as
\[
\left\{\begin{aligned}
&\beta^\ell_{s,m,c}={\rm softmax}_c\!\left(\frac{(\mathbf{W}^{\ell,\mathrm{proto}}_r\mathbf{v}^{\ell}_{s,m})^\top(\mathbf{W}^{\ell,\mathrm{proto}}_i\tilde{\mathbf{p}}^{\ell}_{s,c})}{\sqrt{d}}\right),\\
&\mathbf{r}^\ell_{s,m}=\sum_{c=1}^{C}\beta^\ell_{s,m,c}\mathbf{W}^{\ell,\mathrm{proto}}_s\tilde{\mathbf{p}}^{\ell}_{s,c},\\
&\mathbf{g}^\ell_{s,m}=\sigma\!\left({\rm MLP}_\ell([\mathbf{v}^{\ell}_{s,m};\mathbf{r}^\ell_{s,m}])\right),\\
&\widehat{\mathbf{v}}^\ell_{s,m}=\mathbf{v}^{\ell}_{s,m}+\mathbf{g}^\ell_{s,m}\odot\mathbf{W}^{\ell,\mathrm{proto}}_o\mathbf{r}^\ell_{s,m},\\
&\mathbf{u}^{\ell,{\rm out}}_{s,m}=\widehat{\mathbf{v}}^\ell_{s,m}+{\rm FFN}_\ell(\widehat{\mathbf{v}}^\ell_{s,m}).
\end{aligned}\right.
\]
The learned projections \(\mathbf{W}^{\ell,\mathrm{proto}}_r,\mathbf{W}^{\ell,\mathrm{proto}}_i,\mathbf{W}^{\ell,\mathrm{proto}}_s,\mathbf{W}^{\ell,\mathrm{proto}}_o\in\mathbb{R}^{d\times d}\) belong to the prototype-reassembly router, with independent parameters in each block. For each variable, \(\beta^\ell_{s,m,c}\in[0,1]\) normalizes over \(C\) prototypes to form \(\mathbf{r}^\ell_{s,m}\in\mathbb{R}^{d}\), and \(\mathbf{g}^\ell_{s,m}\in(0,1)^d\) gates its contribution to \(\widehat{\mathbf{v}}^\ell_{s,m}\in\mathbb{R}^{d}\). Stacking these retrieved vectors, gates, and updated representations as rows recovers \(\mathbf{R}^\ell_s,\mathbf{G}^\ell_s,\widehat{\mathbf{V}}^\ell_s\) in Eq.~(\ref{eq:method_reassemble}).

\subsection{Cross-Scale Conditioning and Local Memory}
\label{app:method_cross_scale}

Patch-level temporal and variable interaction transform \(\mathbf{s}_{p,m}\) into \(\mathbf{H}^{\rm raw}_{p,m}\in\mathbb{R}^{d}\). The intra-patch encoder receives \(\mathbf{H}_{p,m}=\mathbf{H}^{\rm raw}_{p-1,m}\) for \(p\ge1\), initialized with the learned vector \(\mathbf{H}_{0,m}=\mathbf{h}^{\rm bos}_m\in\mathbb{R}^{d}\). \emph{This one-patch delay prevents future leakage: time steps within patch \(p\) cannot access its completed summary, which includes observations later in that patch.}

After intra-patch temporal interaction, each variable representation reads the delayed patch memory. For a query \(\mathbf{q}\in\mathbb{R}^{d}\) and memory \(\mathbf{B}\in\mathbb{R}^{M\times d}\), whose \(m\)-th row is written as the column vector \(\mathbf{B}_m\in\mathbb{R}^{d}\), the retrieval operator in Eq.~(\ref{eq:method_retrieve}) is
\[
\left\{\begin{aligned}
&\gamma_m={\rm softmax}_{m:a_m=1}\!\left(\frac{(\mathbf{W}^{\mathrm{ret}}_r\mathbf{q})^\top(\mathbf{W}^{\mathrm{ret}}_i\mathbf{B}_m)}{\sqrt{d}}\right),\\
&\mathbf{r}=\sum_{m:a_m=1}\gamma_m\mathbf{W}^{\mathrm{ret}}_s\mathbf{B}_m,\\
&\mathbf{g}=\sigma({\rm MLP}([\mathbf{q};\mathbf{r}])),\\
&\tilde{\mathbf{q}}=\mathbf{q}+\mathbf{g}\odot\mathbf{W}^{\mathrm{ret}}_o\mathbf{r},\\
&{\rm Retrieve}(\mathbf{q},\mathbf{B})=\tilde{\mathbf{q}}+{\rm FFN}(\tilde{\mathbf{q}}).
\end{aligned}\right.
\]
The learned matrices \(\mathbf{W}^{\mathrm{ret}}_r,\mathbf{W}^{\mathrm{ret}}_i,\mathbf{W}^{\mathrm{ret}}_s,\mathbf{W}^{\mathrm{ret}}_o\in\mathbb{R}^{d\times d}\) project the query, memory keys, memory values, and retrieved vector, respectively. These parameters are independent of the prototype-reassembly parameters in Appendix~\ref{app:method_temporal_variable}. The scalar weights \(\gamma_m\in[0,1]\) normalize only over valid channels with \(a_m=1\). Both the retrieved vector \(\mathbf{r}\) and updated query \(\tilde{\mathbf{q}}\) lie in \(\mathbb{R}^{d}\). The gate MLP maps \([\mathbf{q};\mathbf{r}]\in\mathbb{R}^{2d}\) to \(\mathbf{g}\in(0,1)^d\), controlling the memory contribution to each query feature.

For cross-scale conditioning, the query is the temporal output \(\bar{\mathbf{u}}^{\rm intra}_{p,r,m}\), and the memory contains the delayed summaries \(\mathbf{H}_{p,1:M}\). This produces the same history-conditioned variable representation as Eq.~(\ref{eq:method_topdown_read}):
\[
\mathbf{v}^{\rm intra}_{p,r,m}={\rm Retrieve}\!\left(\bar{\mathbf{u}}^{\rm intra}_{p,r,m},\mathbf{H}_{p,1:M}\right)\in\mathbb{R}^{d}.
\]
Stacking \(\mathbf{v}^{\rm intra}_{p,r,m}\) over channels gives \(\mathbf{V}^{\rm intra}_{p,r}\in\mathbb{R}^{M\times d}\), which undergoes prototype-based variable interaction through Eqs.~(\ref{eq:method_diffattn})--(\ref{eq:method_reassemble}). Each subsequent intra-patch block uses the preceding block's \(\mathbf{U}^{{\rm intra},{\rm out}}_{p,r}\) as its input \(\mathbf{U}^{\rm intra}_{p,r}\) and repeats temporal interaction, history retrieval, and variable interaction. For the final block, \(\mathbf{L}_{t,m}=\mathbf{u}^{{\rm intra},{\rm out}}_{(p,r),m}\in\mathbb{R}^{d}\) is the local memory used for posterior reconstruction; its one-step shift \(\mathbf{L}^-_{t,m}=\mathbf{L}_{t-1,m}\), with \(\mathbf{L}_{0,m}=\mathbf{0}\), is used for prior prediction.

\subsection{State Queries and State Reconstruction}
\label{app:method_state_reconstruction}

At each time step, the encoder provides \(M\) observation-channel memory vectors \(\mathbf{L}_{t,m}\). To estimate the \(N\) latent state components, the reconstruction module assigns a learned query to each state channel \(n\) and uses it to retrieve information from all valid observation channels. This permits \(N\) and \(M\) to differ. For \(t=pL_p+r\), the initial query is
\[
\mathbf{q}_{t,n}=\mathbf{e}^x_n+\mathbf{e}^{\rm patch}_p+\mathbf{e}^{\rm pos}_r\in\mathbb{R}^{d}.
\]
Here \(\mathbf{e}^x_n\in\mathbb{R}^{d}\) is the sum of the learned embedding for state channel \(n\) and a learned vector shared by all state queries. The patch-index embedding \(\mathbf{e}^{\rm patch}_p\) and within-patch-position embedding \(\mathbf{e}^{\rm pos}_r\) are shared with the observation encoder. The query \(\mathbf{q}_{t,n}\) contains no observation or target state value; it obtains evidence through the following two retrieval operations.

For prior prediction, \(\mathbf{q}_{t,n}\) queries shifted memory \(\mathbf{L}^-_{t,1:M}=\mathbf{L}_{t-1,1:M}\) to form the prior latent representation \(\mathbf{z}^-_{t,n}\in\mathbb{R}^{d}\) from observations before \(t\). This prior then queries current memory \(\mathbf{L}_{t,1:M}\) to obtain current evidence \(\mathbf{e}_{t,n}\in\mathbb{R}^{d}\) for the posterior update at \(t\):
\[
\left\{\begin{aligned}
&\mathbf{z}^-_{t,n}={\rm Retrieve}(\mathbf{q}_{t,n},\mathbf{L}^-_{t,1:M}),\\
&\mathbf{e}_{t,n}={\rm Retrieve}(\mathbf{z}^-_{t,n},\mathbf{L}_{t,1:M}).
\end{aligned}\right.
\]
The two retrieval operations use Eq.~(\ref{eq:method_retrieve}) with separate projections, gates, and FFNs, independent of those used for cross-scale conditioning in Appendix~\ref{app:method_cross_scale}. The posterior update then concatenates the prior, current evidence, and their difference into a vector in \(\mathbb{R}^{3d}\), as described in Section~\ref{sec:method_belief}:
\[
\left\{\begin{aligned}
&\mathbf{g}_{t,n}=\sigma\!\left({\rm MLP}_{\rm gate}([\mathbf{z}^-_{t,n};\mathbf{e}_{t,n};\mathbf{e}_{t,n}-\mathbf{z}^-_{t,n}])\right),\\
&\boldsymbol{\Delta}_{t,n}={\rm MLP}_{\rm update}([\mathbf{z}^-_{t,n};\mathbf{e}_{t,n};\mathbf{e}_{t,n}-\mathbf{z}^-_{t,n}]),\\
&\mathbf{z}^+_{t,n}=\mathbf{z}^-_{t,n}+\mathbf{g}_{t,n}\odot\boldsymbol{\Delta}_{t,n}.
\end{aligned}\right.
\]
The difference \(\mathbf{e}_{t,n}-\mathbf{z}^-_{t,n}\) compares current evidence with the prior in the learned feature space. The two MLPs map \(\mathbb{R}^{3d}\) to \(\mathbb{R}^{d}\), producing the sigmoid gate \(\mathbf{g}_{t,n}\in(0,1)^d\) and correction \(\boldsymbol{\Delta}_{t,n}\in\mathbb{R}^{d}\) that form the posterior latent representation \(\mathbf{z}^+_{t,n}\in\mathbb{R}^{d}\). Learned quantile projections \(W^+_\tau\in\mathbb{R}^{1\times d}\) yield scalar normalized state quantiles, with the median mapped back to the original state scale:
\[
\left\{\begin{aligned}
&Q^+_{t,n,\tau}=W^+_\tau\mathbf{z}^+_{t,n}\in\mathbb{R},\quad \tau\in\mathcal Q=\{0.1,0.5,0.9\},\\
&\hat{x}_{t,n}=\sigma^x_n Q^+_{t,n,0.5}+\mu^x_n\in\mathbb{R}.
\end{aligned}\right.
\]

\subsection{Exact Training Loss}
\label{app:method_training_loss}

Training supervises the state quantiles in the normalized coordinates defined in Section~\ref{sec:method_encoder}. Let \(\Omega_b\) contain the valid time--state pairs \((t,n)\) in sample \(b\), and let \(\mathcal Q=\{0.1,0.5,0.9\}\). The posterior pinball loss is defined below, with the same form applied to prior quantiles \(Q^-\) for \(\mathcal L_{\rm prior}\):
\begin{equation}
\left\{\begin{aligned}
&\mathcal{L}_{\rm post}=\frac{1}{B}\sum_{b=1}^{B}\frac{1}{|\Omega_b|\,|\mathcal Q|}
\sum_{(t,n)\in\Omega_b}\sum_{\tau\in\mathcal Q}
\rho_\tau\!\left(\tilde{x}_{b,t,n}-Q^+_{b,t,n,\tau}\right),\\
&\rho_\tau(e)=\max\{\tau e,(\tau-1)e\}.
\end{aligned}\right.
\end{equation}

The median Huber term provides an additional point-estimation objective. With transition scale \(\delta>0\), the implementation uses the smooth-L1 form
\[
\left\{\begin{aligned}
&\mathcal L_{\rm huber}=\frac{1}{B}\sum_{b=1}^{B}\frac{1}{|\Omega_b|}
\sum_{(t,n)\in\Omega_b}\ell_\delta\!\left(\tilde{x}_{b,t,n}-Q^+_{b,t,n,0.5}\right),\\
&\ell_\delta(e)=\begin{cases}
e^2/(2\delta), & |e|<\delta,\\
|e|-\delta/2, & |e|\ge\delta.
\end{cases}
\end{aligned}\right.
\]
This term is quadratic for small errors and linear for large errors, retaining sensitivity to local estimation accuracy without the quadratic growth of MSE on large residuals.

Prototype-key regularization acts on the learned positive and negative key matrices \(\mathbf K^\ell_\pm\) in Eq.~(\ref{eq:method_diffattn}). Here \(j=1,\ldots,J\) indexes all patch-level and intra-patch blocks, whose individual indices were omitted in that equation. For block \(j\), \(\bar{\mathbf K}_{j,+},\bar{\mathbf K}_{j,-}\in\mathbb R^{C\times d}\) denote its key matrices after normalizing each row to unit \(\ell_2\) norm. The main model has \(J=5\): three patch-level blocks and two intra-patch blocks. The regularizer is
\[
\left\{\begin{aligned}
&\mathcal R_j=\frac{1}{C^2}\left\|\bar{\mathbf K}_{j,+}\bar{\mathbf K}_{j,-}^{\top}\right\|_F^2
+\frac{1}{2C^2}\sum_{\eta\in\{+,-\}}
\left\|\bar{\mathbf K}_{j,\eta}\bar{\mathbf K}_{j,\eta}^{\top}-\mathbf I_C\right\|_F^2,\\
&\mathcal L_{\rm orth}=\frac{1}{J}\sum_{j=1}^{J}\mathcal R_j.
\end{aligned}\right.
\]
Here \(\|\cdot\|_F\) is the Frobenius norm. The first term penalizes alignment across the two key sets, while the remaining terms encourage distinct directions among keys within each set. The full objective combines posterior estimation, prior supervision, median accuracy, and key diversity:
\[
\mathcal L=\mathcal L_{\rm post}
+\lambda_{\rm prior}\mathcal L_{\rm prior}
+\lambda_{\rm huber}\mathcal L_{\rm huber}
+\lambda_{\rm orth}\mathcal L_{\rm orth}.
\]
The coefficients set the relative contributions of the auxiliary terms, with \(\lambda_{\rm prior}\) studied explicitly in Table~\ref{tab:prior_weight_details}. Joint prior and posterior supervision encourages the model to learn both a useful prediction from history and its refinement after receiving the current observation.

\subsection{Main Model Training Configuration}
\label{app:main_hyperparameters}

Table~\ref{tab:main_hyperparameters} lists the hyperparameters used to train the main FilterFM model on the 15,120 trajectories described in Appendix~\ref{app:training_corpus}. Training starts from random initialization, and the checkpoint with the lowest validation raw RMSE is used for both in-domain and zero-shot evaluation.

\begin{table}[H]
\centering
\small
\renewcommand{\arraystretch}{1.00}
\setlength{\tabcolsep}{3pt}
\begin{minipage}[t]{0.49\linewidth}
\vspace{0pt}
\begin{tabular}{@{}>{\raggedright\arraybackslash}p{0.65\linewidth}>{\raggedright\arraybackslash}p{\dimexpr0.35\linewidth-6pt\relax}@{}}
\toprule[1pt]
\textbf{Hyperparameter} & \textbf{Setting} \\
\midrule
\multicolumn{2}{@{}l}{\textbf{Architecture and input representation}} \\
Window length \(T\) & 200 \\
Patch length \(L_p\) & 20 \\
Maximum observation / state dimension & 100 / 100 \\
Hidden width \(d\) & 256 \\
Self-attention heads & 4 \\
FFN hidden width & 1,024 \\
Patch-summary temporal encoding layers \(J_{\rm enc}\) & 2 \\
Patch-level blocks & 3 \\
Intra-patch blocks & 2 \\
Prototype self-attention layers per block & 1 \\
Prior-to-posterior state-reconstruction blocks & 1 \\
Prototype slots \(C\) & 8 \\
Negative-attention coefficient \(\lambda_\ell\) & 0.5 \\
Dropout probability & 0.1 \\
Activation & GELU \\
Gate bias initialization & \(-2.0\) \\
Observation and state normalization & Per-channel instance normalization \\
\bottomrule[1pt]
\end{tabular}
\end{minipage}\hfill
\begin{minipage}[t]{0.49\linewidth}
\vspace{0pt}
\begin{tabular}{@{}>{\raggedright\arraybackslash}p{0.59\linewidth}>{\raggedright\arraybackslash}p{\dimexpr0.41\linewidth-6pt\relax}@{}}
\toprule[1pt]
\textbf{Hyperparameter} & \textbf{Setting} \\
\midrule
\multicolumn{2}{@{}l}{\textbf{Training objective}} \\
Quantile levels \(\mathcal{Q}\) & \(\{0.1,0.5,0.9\}\) \\
Prior-loss weight \(\lambda_{\mathrm{prior}}\) & 1.0 \\
Median Huber weight \(\lambda_{\mathrm{huber}}\) & 0.5 \\
Huber transition scale \(\delta\) & 0.5 \\
Prototype-key regularization weight \(\lambda_{\mathrm{orth}}\) & \(10^{-3}\) \\
\midrule
\multicolumn{2}{@{}l}{\textbf{Optimization and checkpoint selection}} \\
Optimizer & AdamW \\
AdamW coefficients \((\beta_1,\beta_2)\) & \((0.9,0.999)\) \\
AdamW numerical stabilizer \(\epsilon_{\mathrm{opt}}\) & \(10^{-8}\) \\
Initial learning rate & \(10^{-4}\) \\
Learning-rate schedule & Cosine decay to \(10^{-8}\) over 20 epochs \\
Weight decay & \(10^{-5}\) \\
Gradient \(\ell_2\)-norm clipping threshold & 1.0 \\
Batch size & 2 trajectories \\
Training epochs & 20 \\
Early-stopping patience & 15 epochs \\
Mixed-precision training & bfloat16 \\
Checkpoint selection criteria & Validation raw RMSE \\
\bottomrule[1pt]
\end{tabular}
\end{minipage}
\caption{Architecture and training hyperparameters of the main FilterFM model.}
\label{tab:main_hyperparameters}
\end{table}

\clearpage
\subsection{Interpreting Patch-Level Interaction}
\label{app:patch_attention_interpretation}
Figure~\ref{fig:patch_attention_visualization} shows all four temporal-attention heads in the third patch-level block of the checkpoint evaluated in Table~\ref{tab:panda_foundation_selected14_online}. Each system uses its first test trajectory and first observation channel, with heads displayed separately rather than averaged. Each heatmap row shows attention over the available patch history, with the causal mask restricting the first row to P0. For Cubic-Sensoring100, head 1 assigns its largest weight to P0 for queries P1--P5, whereas heads 2 and 3 emphasize P3 for queries P4--P7. In Duffing64, at P9, head 2 places its largest weight on P5 and head 3 on P6. These patterns show how different heads access historical summaries to complement intra-patch processing of current observations.

\begingroup
\setlength{\intextsep}{2pt}
\begin{figure}[H]
\centering
\includegraphics[width=\linewidth,page=3]{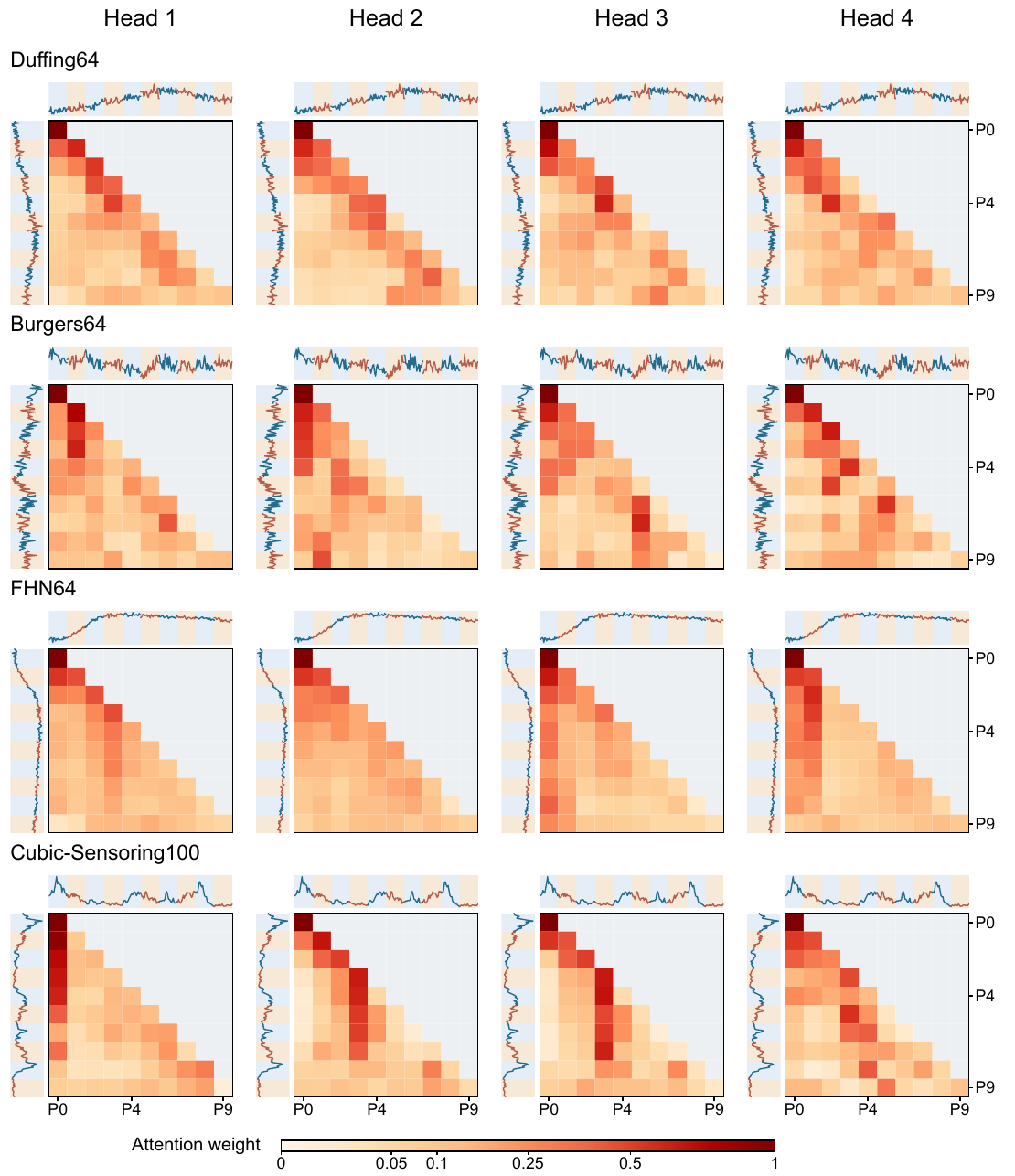}
\caption{Patch-level temporal attention from all four heads of the third patch-level block in the Table~\ref{tab:panda_foundation_selected14_online} checkpoint. Each system occupies one row and heads 1--4 occupy the columns. Within each heatmap, rows are query patches and columns are key patches; gray cells are causally masked. The marginal curves show the same unsmoothed observation channel in both directions, with alternating colors marking 20-step patches. All panels share a square-root color scale over weights from 0 to 1.}
\label{fig:patch_attention_visualization}
\end{figure}
\endgroup

\clearpage
\begin{figure}[p]
\centering
\includegraphics[width=\linewidth,page=1]{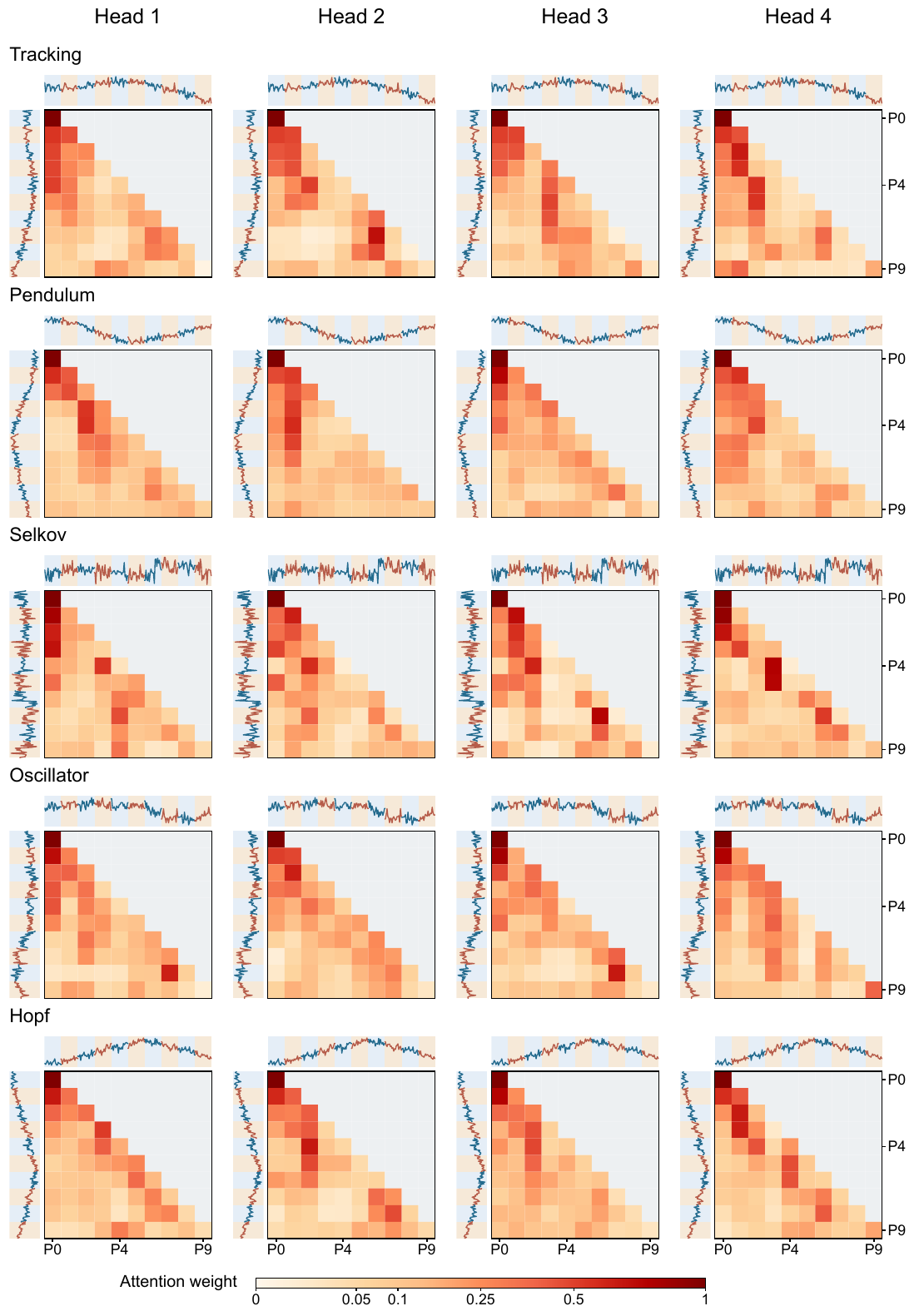}
\par\vspace{\abovecaptionskip}
\noindent Figure~\ref*{fig:patch_attention_visualization} (continued): All four heads for Tracking, Pendulum, Selkov, Oscillator, and Hopf, using the same checkpoint, trajectory and channel selection, and color scale.
\end{figure}

\begin{figure}[p]
\centering
\includegraphics[width=\linewidth,page=2]{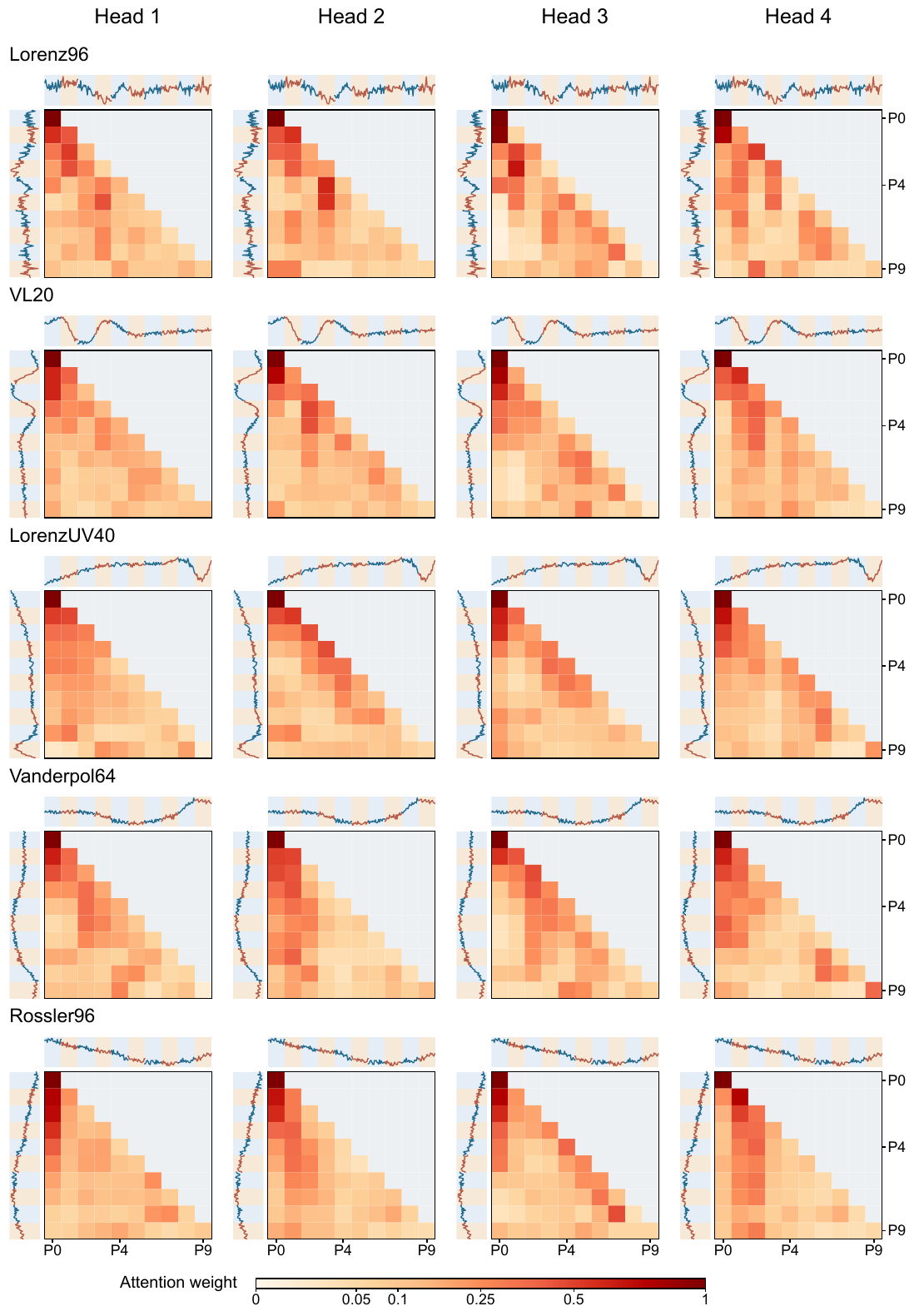}
\par\vspace{\abovecaptionskip}
\noindent Figure~\ref*{fig:patch_attention_visualization} (continued): All four heads for Lorenz96, VL20, LorenzUV40, Vanderpol64, and Rossler96, using the same checkpoint, trajectory and channel selection, and color scale.
\end{figure}

\clearpage
\subsection{Interpreting Prototype-Based Variable Interaction}
\label{app:prototype_writing_interpretation}
Figure~\ref{fig:prototype_writing} shows the coefficients used to aggregate observation-channel representations into prototypes in the third patch-level block of the Table~\ref{tab:panda_foundation_selected14_online} checkpoint. We use the first test trajectory of four systems and retain all observation channels.

In LorenzUV40, slot 1 assigns larger coefficients, on average, to the first eight slow-variable channels than to the remaining channels at all three displayed patches. In Cubic-Sensoring100, the mean coefficient in slot 5 decreases from P0 to P8, while the mean coefficient in slot 4 increases. These patterns show how fixed prototype keys produce input-dependent aggregation, reusing eight slots across systems with different observation dimensions.

\begin{figure}[H]
\centering
\includegraphics[width=\linewidth]{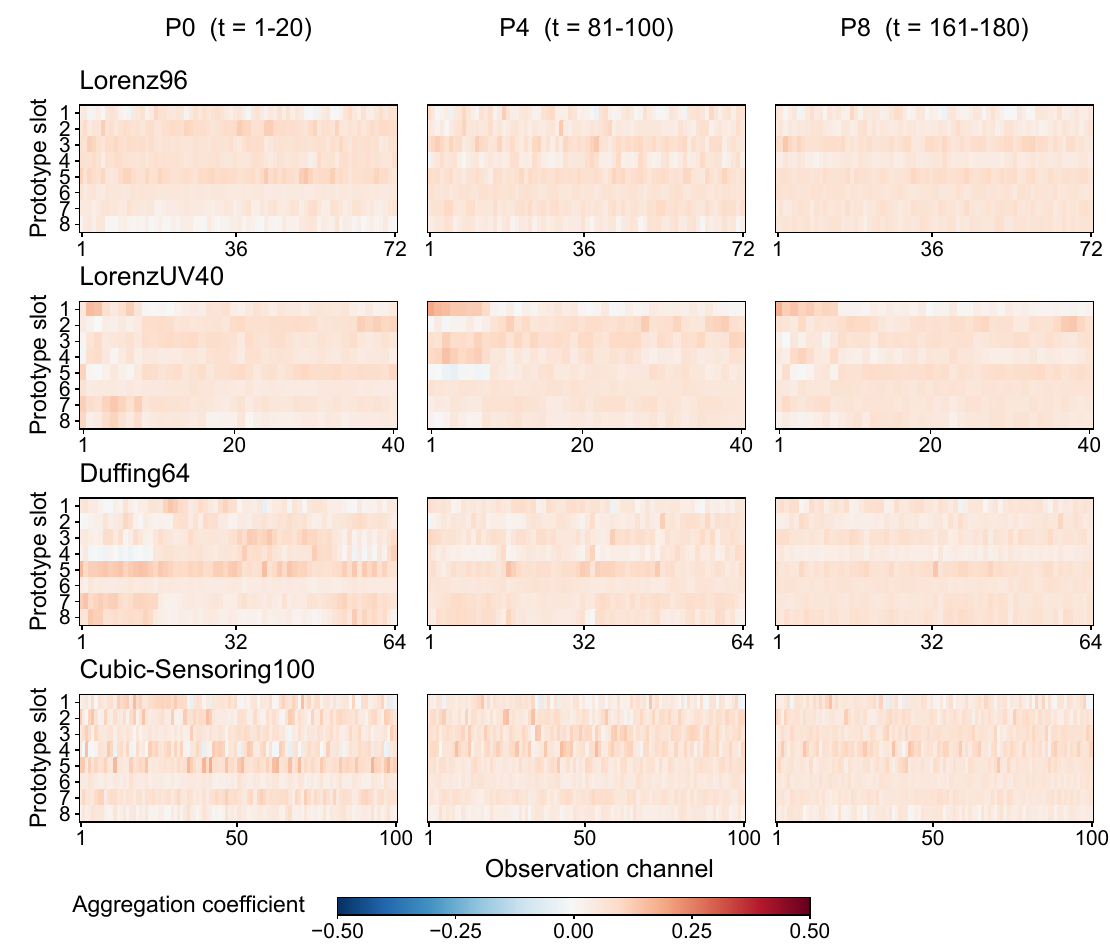}
\caption{Coefficients \(\alpha^{\ell,+}_{p,m,c}-0.5\alpha^{\ell,-}_{p,m,c}\) for forming prototypes in Eq.~(\ref{eq:method_diffattn}), before the common \(1/M\) factor. Rows are prototype slots and columns are observation channels. Panels share a linear color scale: red indicates positive coefficients and blue negative coefficients.}
\label{fig:prototype_writing}
\end{figure}

\endgroup 

\clearpage
\section{More State Estimation Showcase}
\label{app:state_estimation_showcase}

\subsection{In-Domain State Estimation}
\label{app:indomain_trajectory_showcase}

\begin{figure}[H]
\centering
\includegraphics[width=\linewidth,page=1]{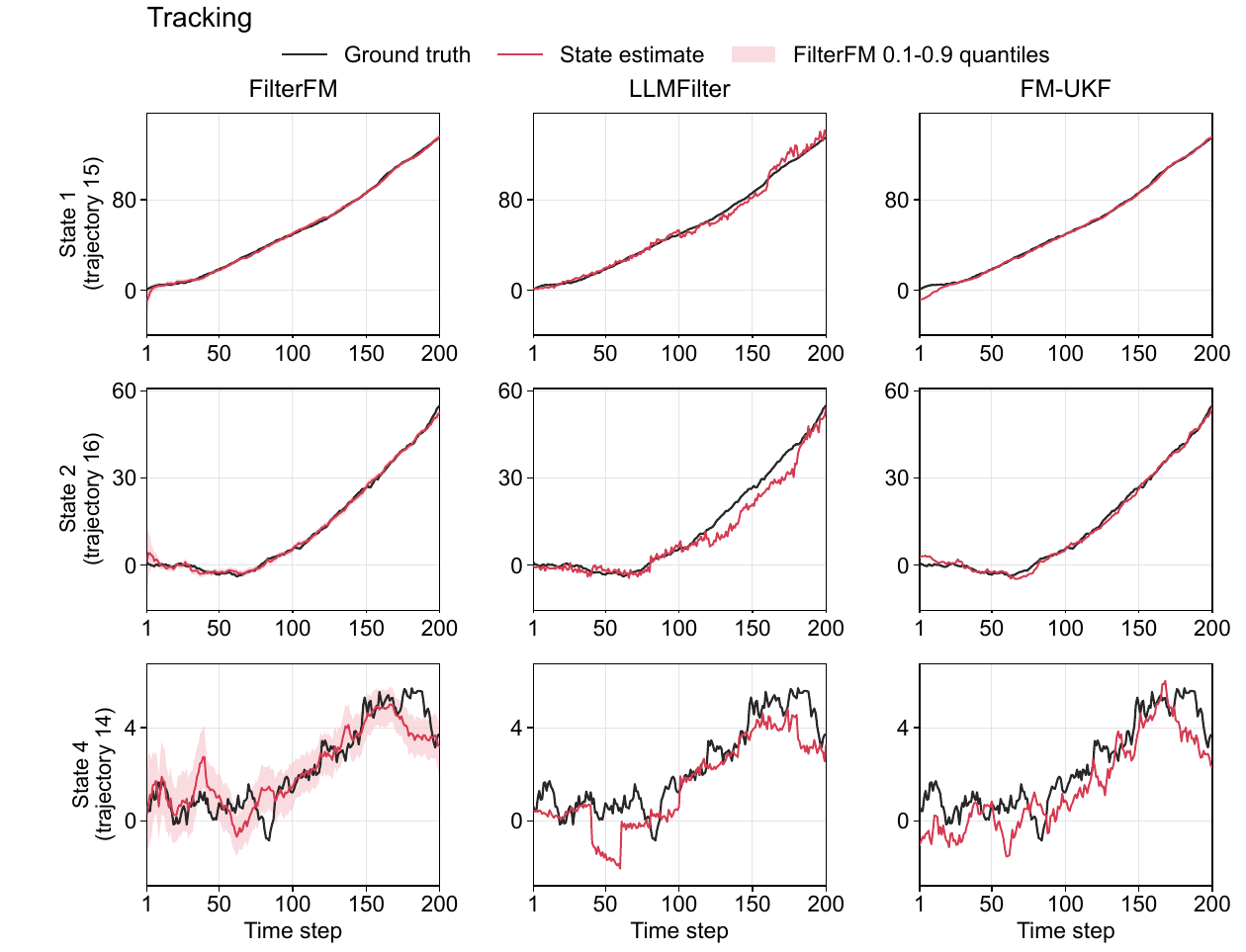}
\caption{Additional in-domain state-estimation examples on Tracking.}
\label{fig:additional_state_estimation_showcase_tracking}
\end{figure}

\begin{figure}[p]
\centering
\includegraphics[width=\linewidth,page=2]{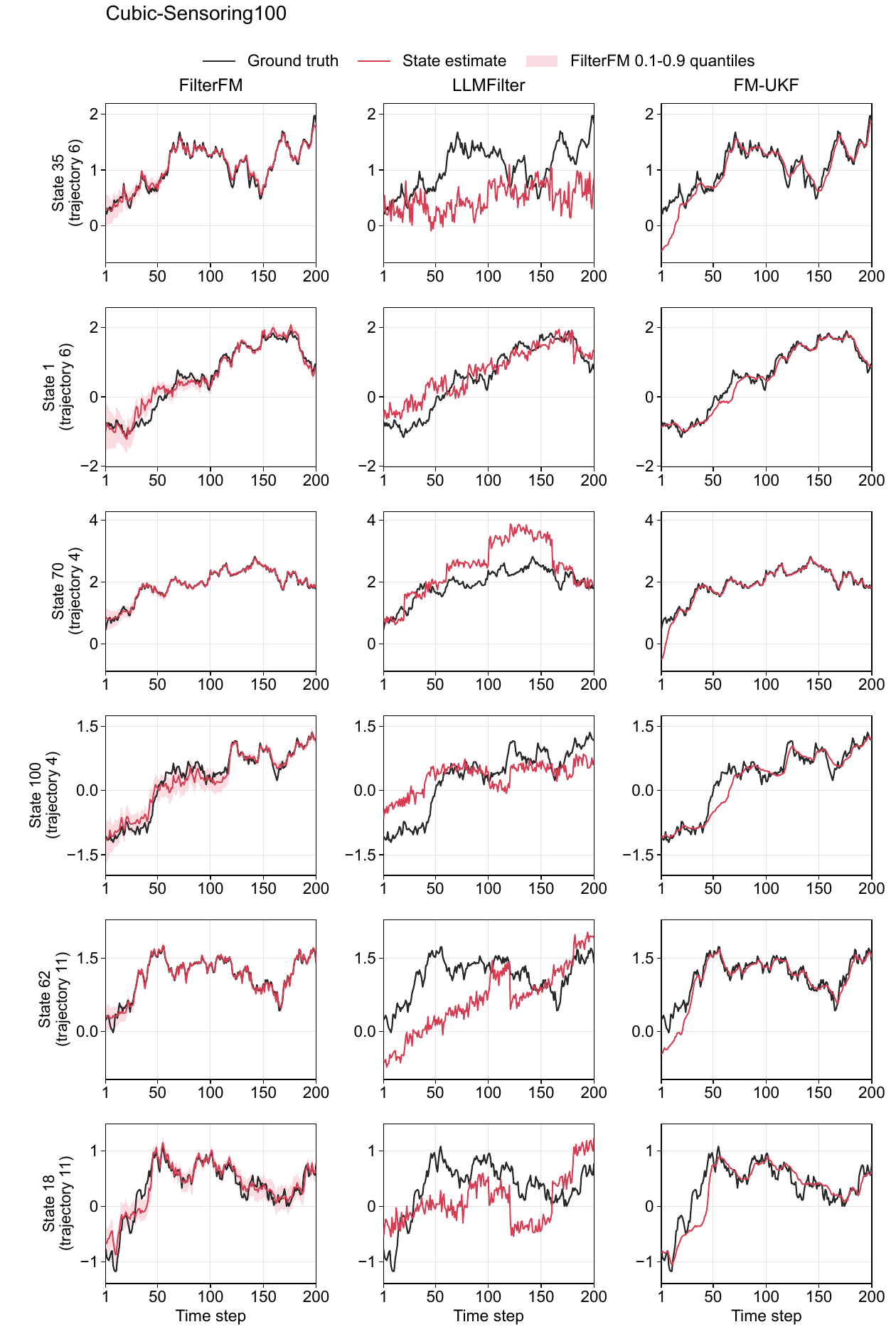}
\caption{In-domain state estimation on Cubic-Sensoring100: six state channels from three held-out trajectories, with the same axis limits across methods in each row.}
\label{fig:additional_state_estimation_showcase_cubic}
\end{figure}

\begin{figure}[p]
\centering
\setlength{\parskip}{0pt}
\includegraphics[width=0.97\linewidth,page=3]{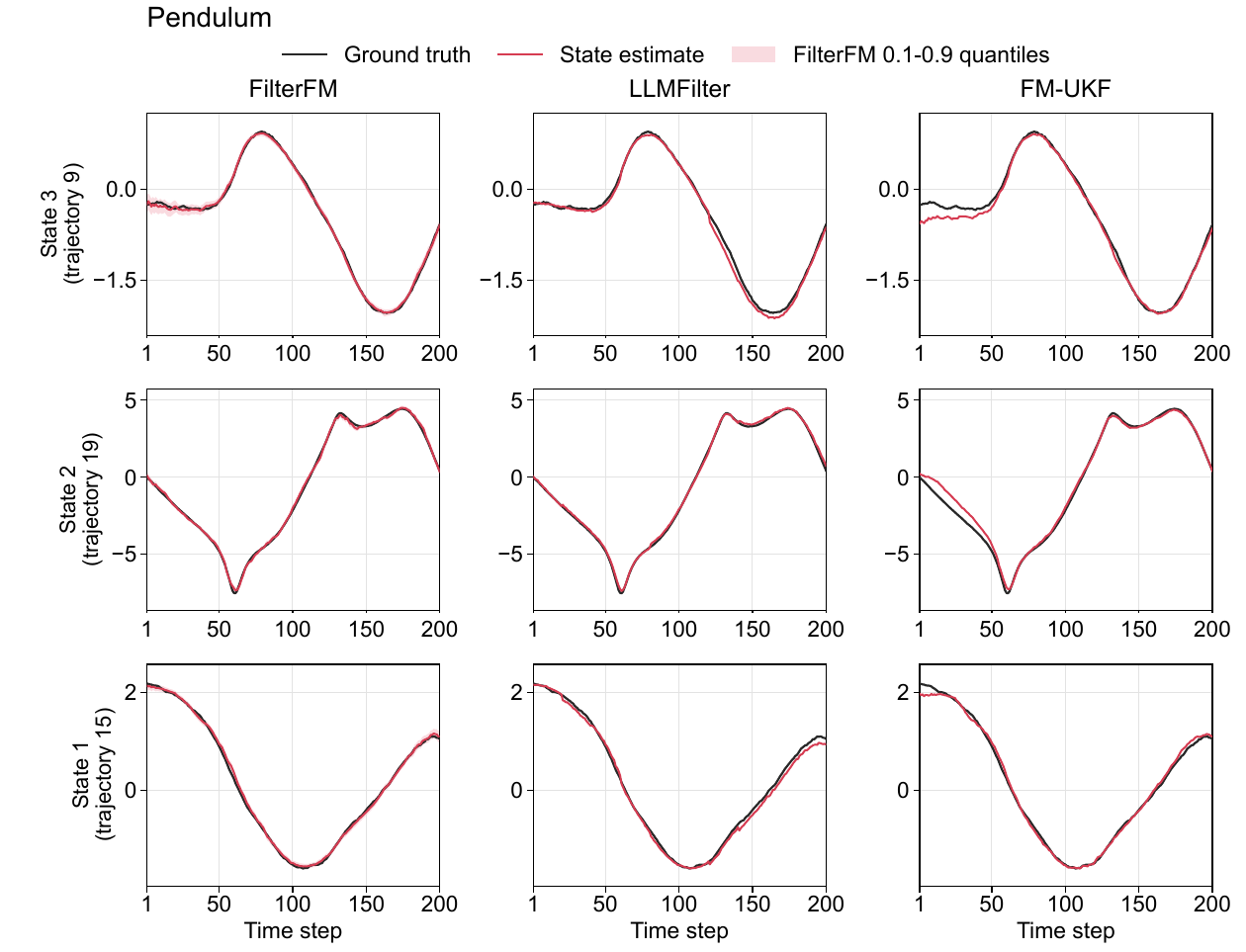}
\caption{Additional in-domain state-estimation examples on Pendulum.}
\label{fig:additional_state_estimation_showcase_pendulum}
\par\vspace{8pt}
\includegraphics[width=0.97\linewidth,page=4]{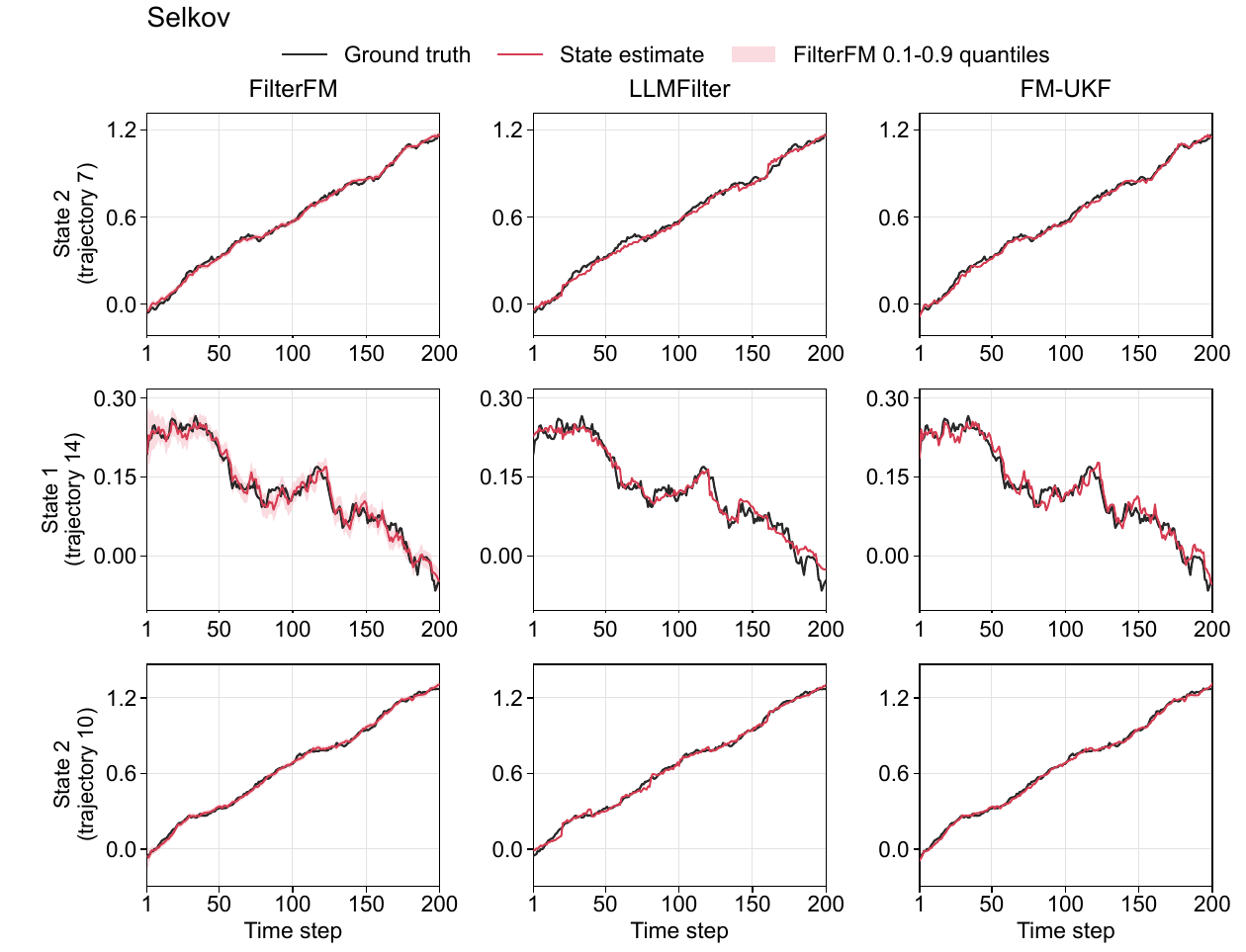}
\caption{Additional in-domain state-estimation examples on Selkov.}
\label{fig:additional_state_estimation_showcase_selkov}
\end{figure}

\begin{figure}[p]
\centering
\setlength{\parskip}{0pt}
\includegraphics[width=0.97\linewidth,page=5]{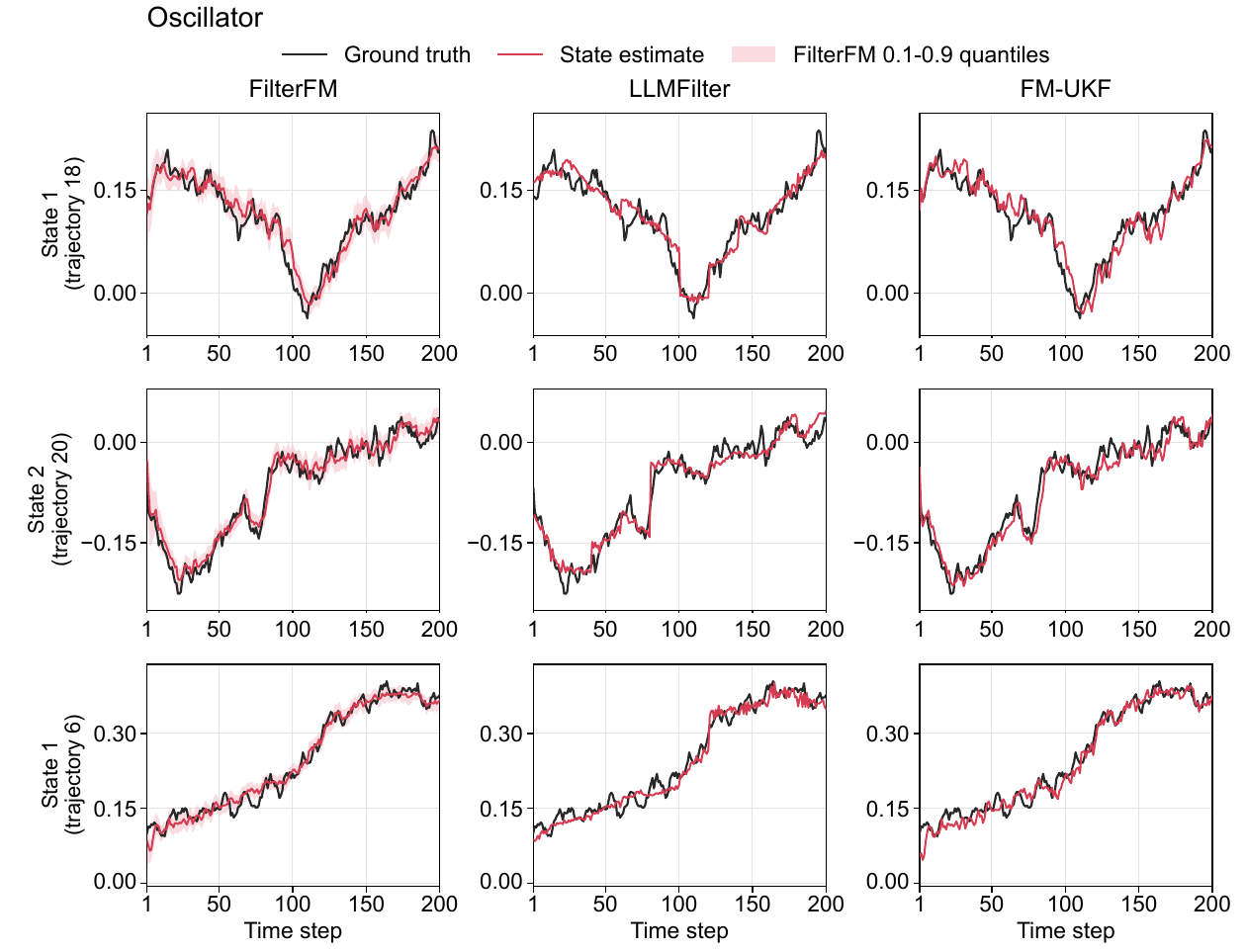}
\caption{Additional in-domain state-estimation examples on Oscillator.}
\label{fig:additional_state_estimation_showcase_oscillator}
\par\vspace{8pt}
\includegraphics[width=0.97\linewidth,page=6]{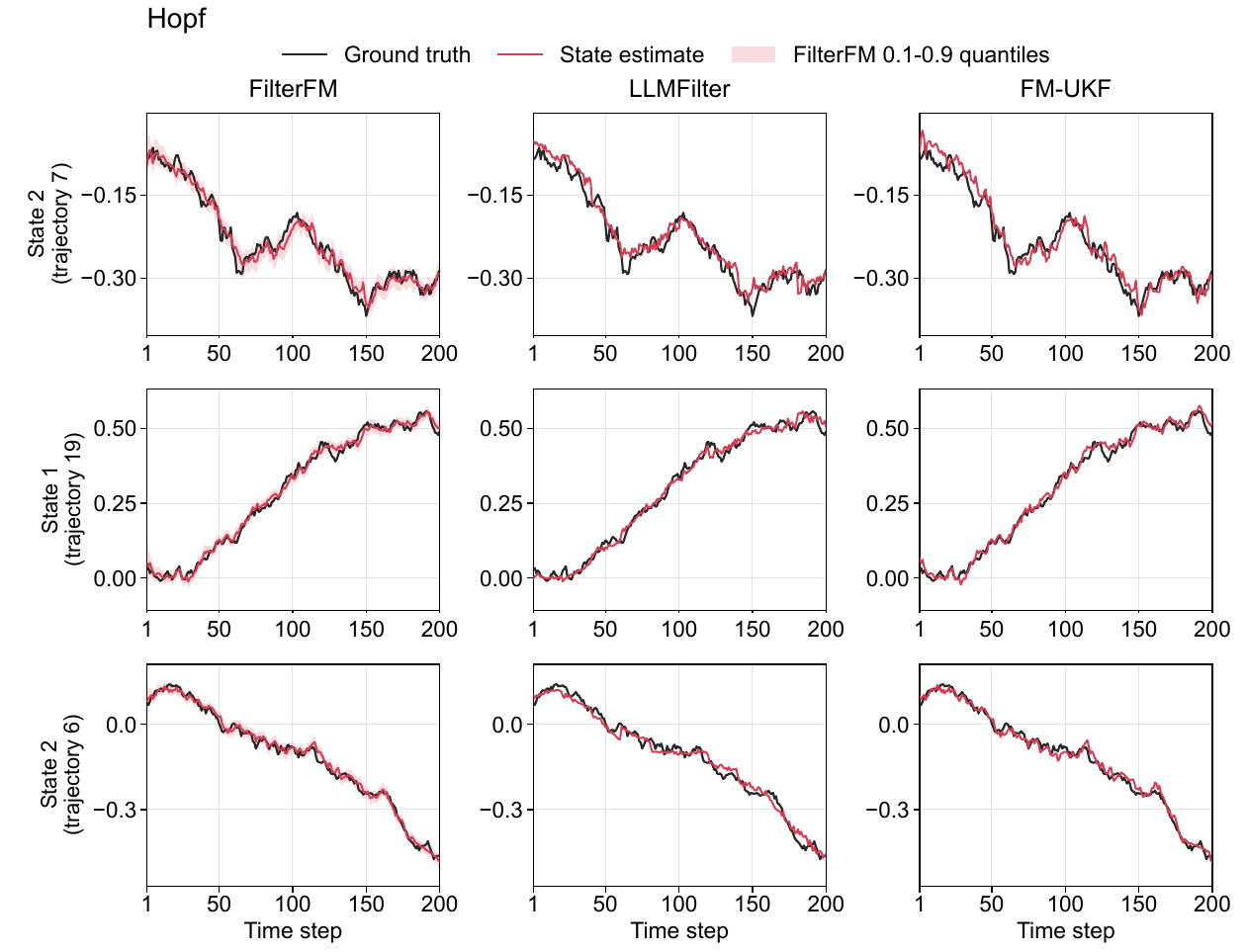}
\caption{Additional in-domain state-estimation examples on Hopf.}
\label{fig:additional_state_estimation_showcase_hopf}
\end{figure}

\begin{figure}[p]
\centering
\includegraphics[width=\linewidth,page=7]{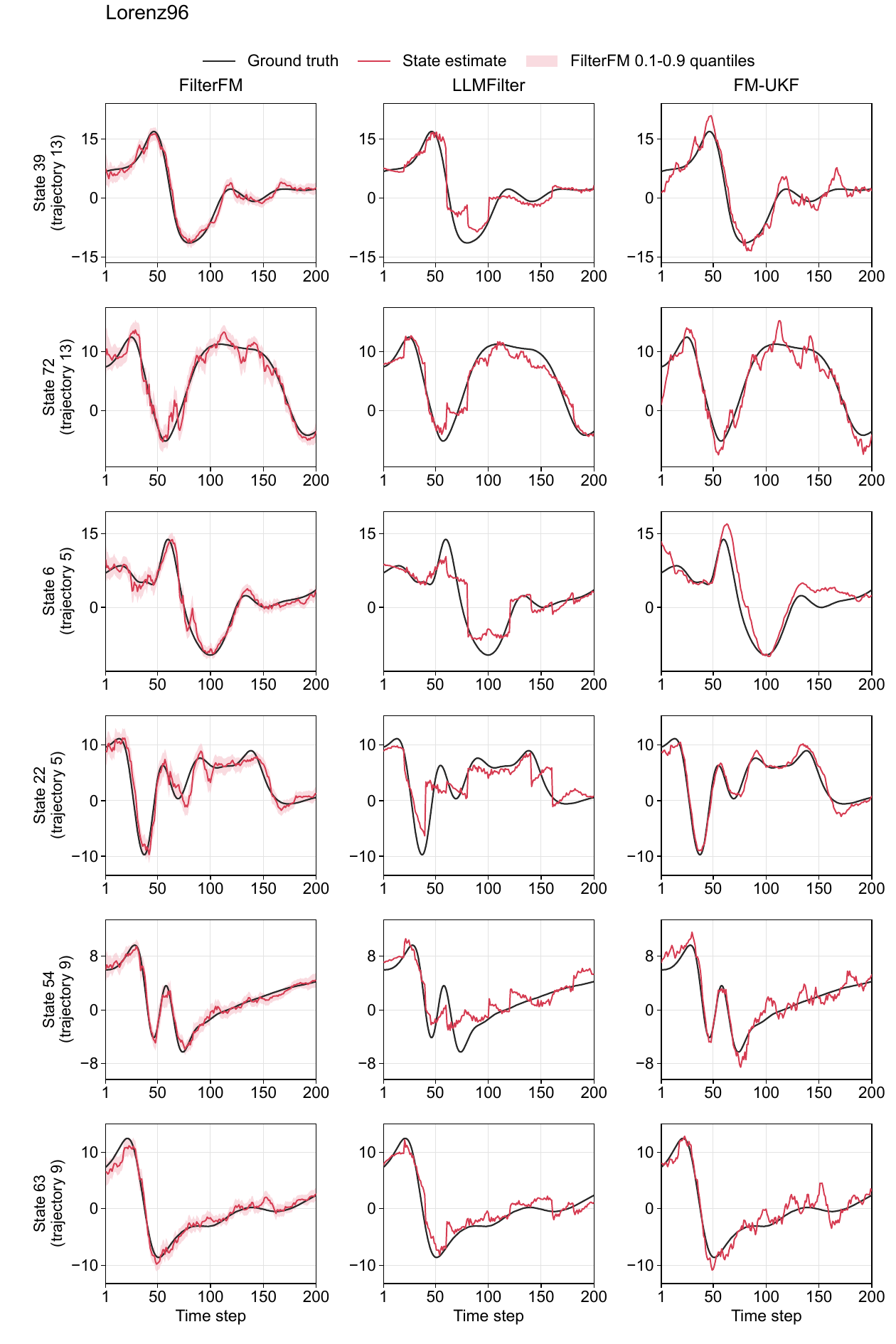}
\caption{In-domain state estimation on Lorenz96: six state channels from three held-out trajectories, with the same axis limits across methods in each row.}
\label{fig:additional_state_estimation_showcase_lorenz96}
\end{figure}

\begin{figure}[p]
\centering
\includegraphics[width=\linewidth,page=8]{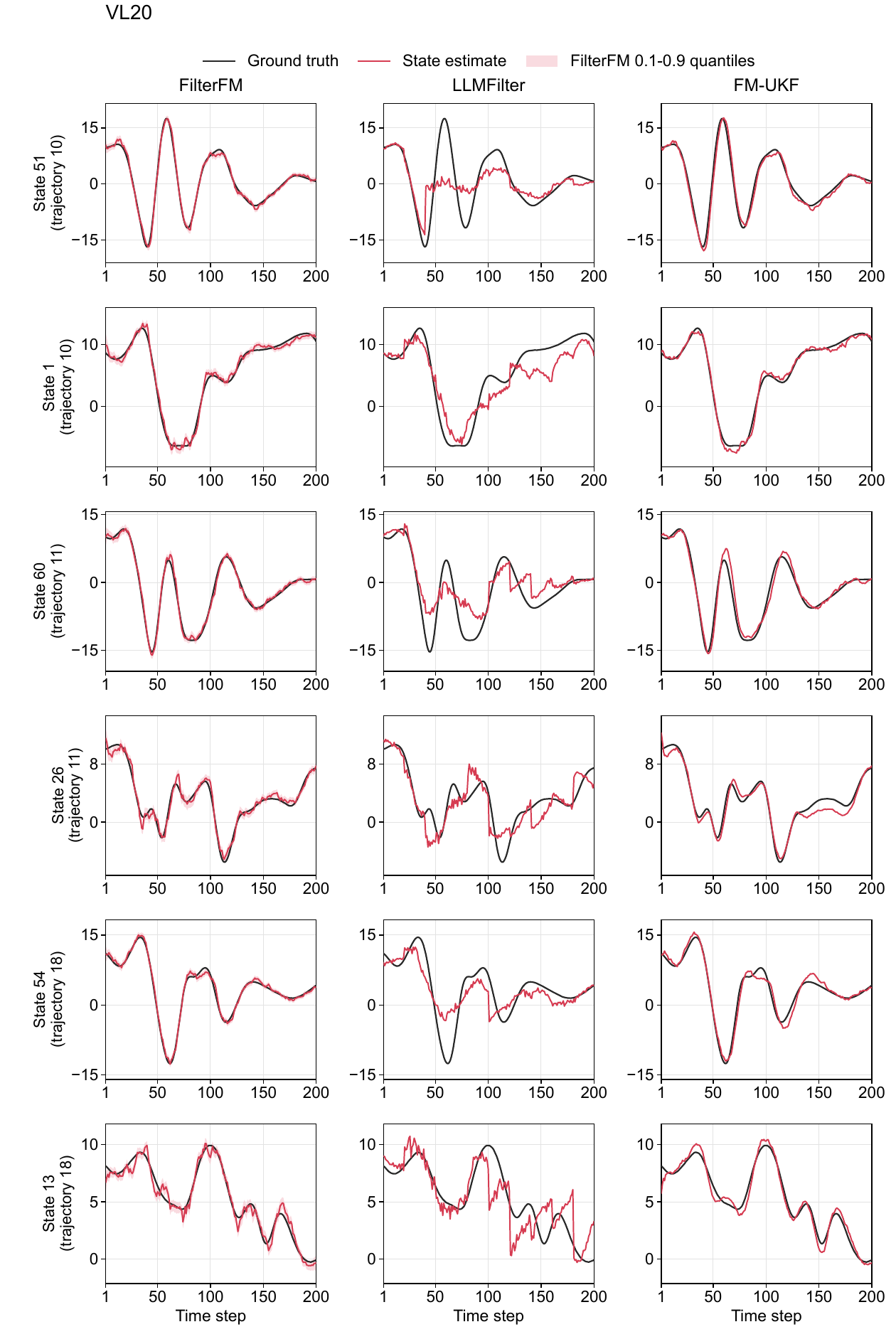}
\caption{In-domain state estimation on VL20: six state channels from three held-out trajectories, with the same axis limits across methods in each row.}
\label{fig:additional_state_estimation_showcase_vl20}
\end{figure}

\begin{figure}[p]
\centering
\includegraphics[width=\linewidth,page=9]{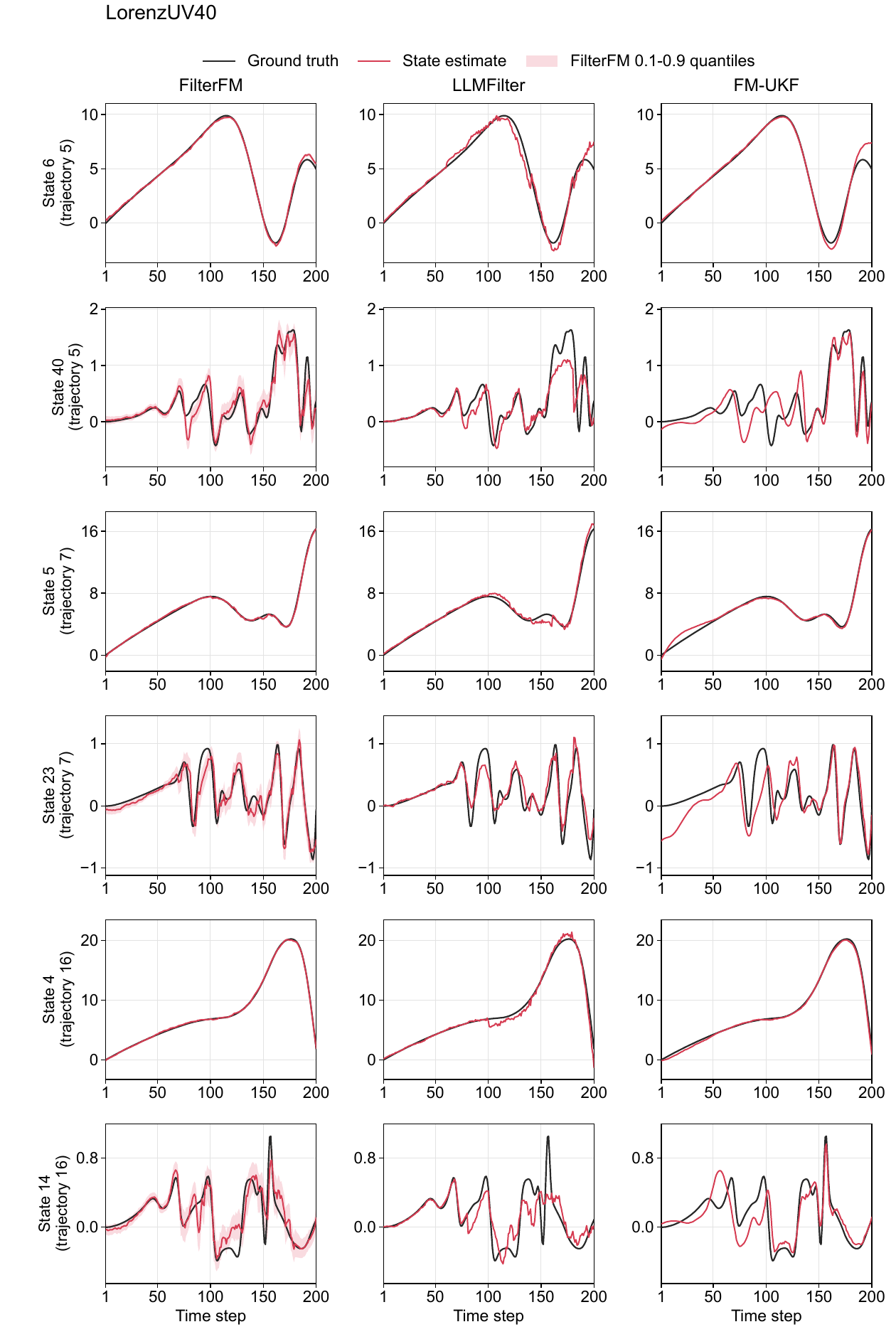}
\caption{In-domain state estimation on LorenzUV40: six state channels from three held-out trajectories, with the same axis limits across methods in each row.}
\label{fig:additional_state_estimation_showcase_lorenzuv40}
\end{figure}

\begin{figure}[p]
\centering
\includegraphics[width=\linewidth,page=10]{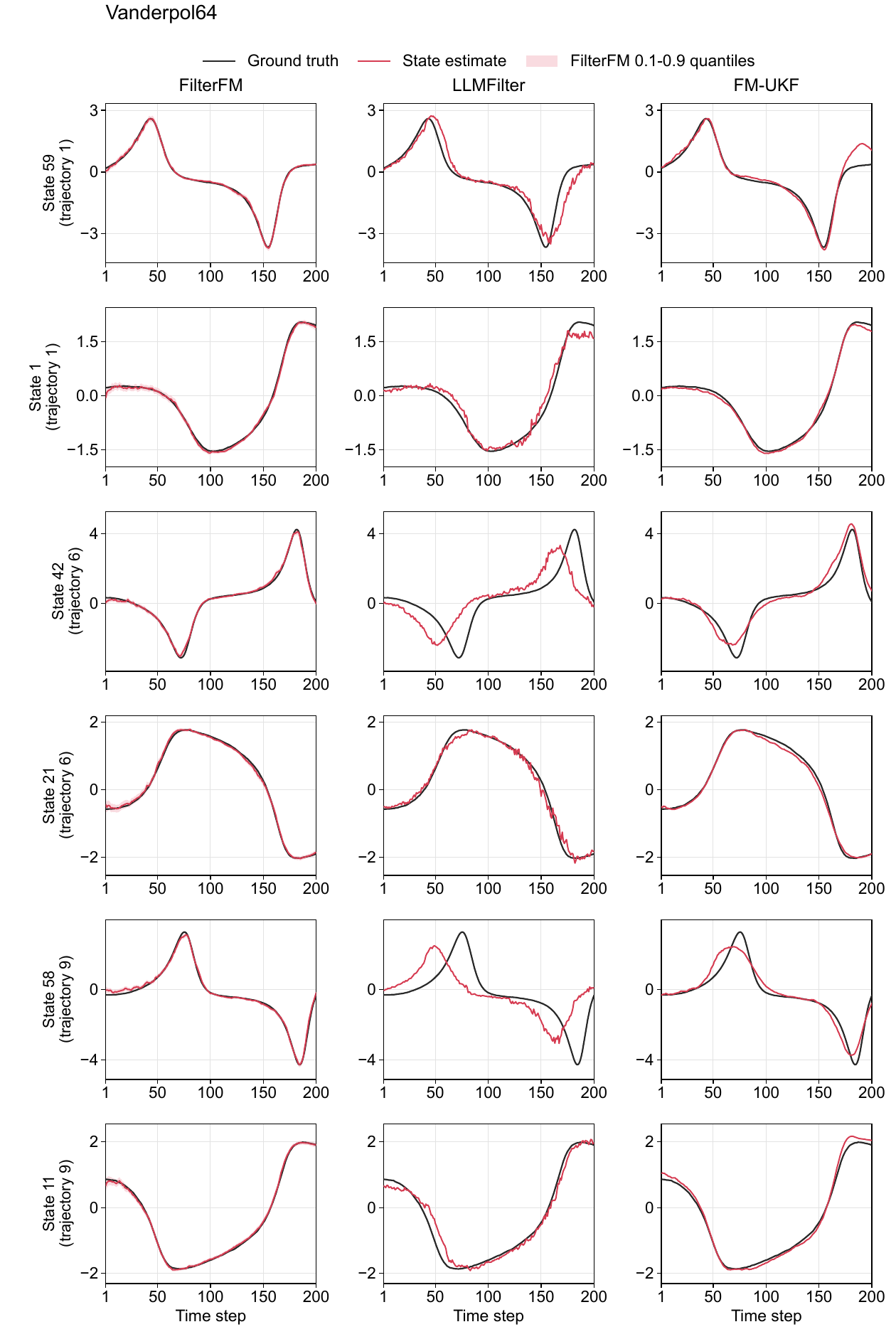}
\caption{In-domain state estimation on Vanderpol64: six state channels from three held-out trajectories, with the same axis limits across methods in each row.}
\label{fig:additional_state_estimation_showcase_vanderpol64}
\end{figure}

\begin{figure}[p]
\centering
\includegraphics[width=\linewidth,page=11]{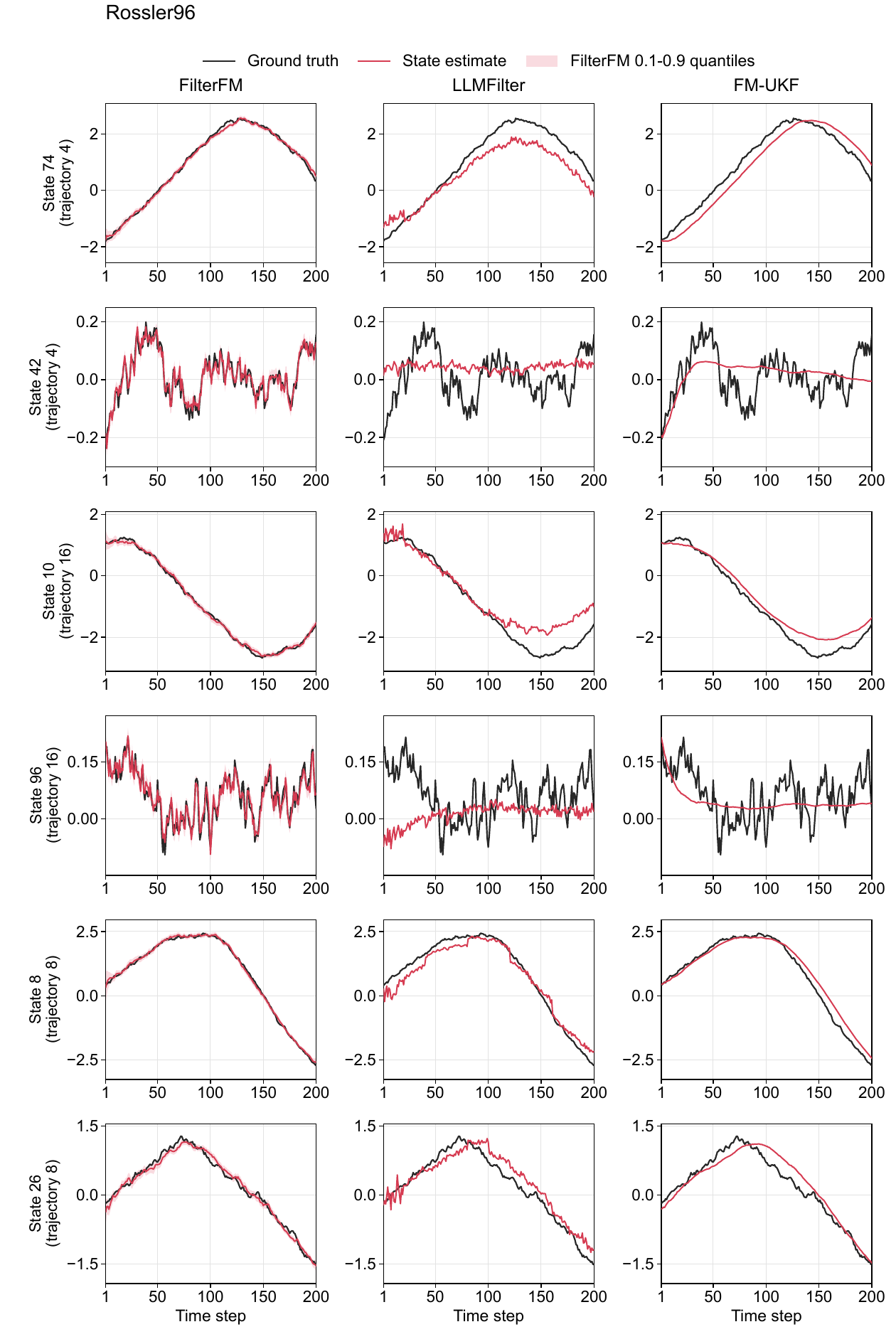}
\caption{In-domain state estimation on Rossler96: six state channels from three held-out trajectories, with the same axis limits across methods in each row.}
\label{fig:additional_state_estimation_showcase_rossler96}
\end{figure}

\begin{figure}[p]
\centering
\includegraphics[width=\linewidth,page=12]{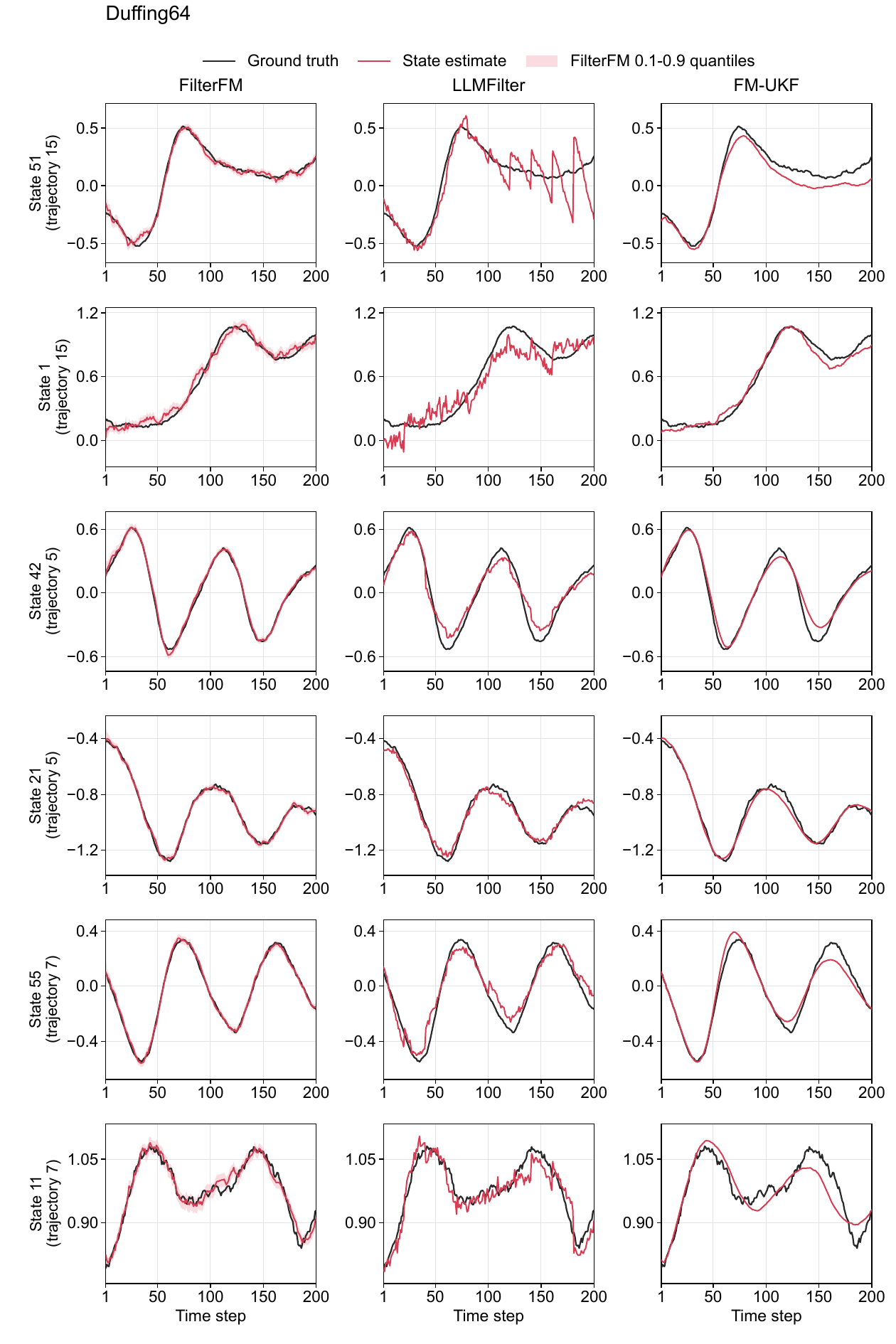}
\caption{In-domain state estimation on Duffing64: six state channels from three held-out trajectories, with the same axis limits across methods in each row.}
\label{fig:additional_state_estimation_showcase_duffing64}
\end{figure}

\begin{figure}[p]
\centering
\includegraphics[width=\linewidth,page=13]{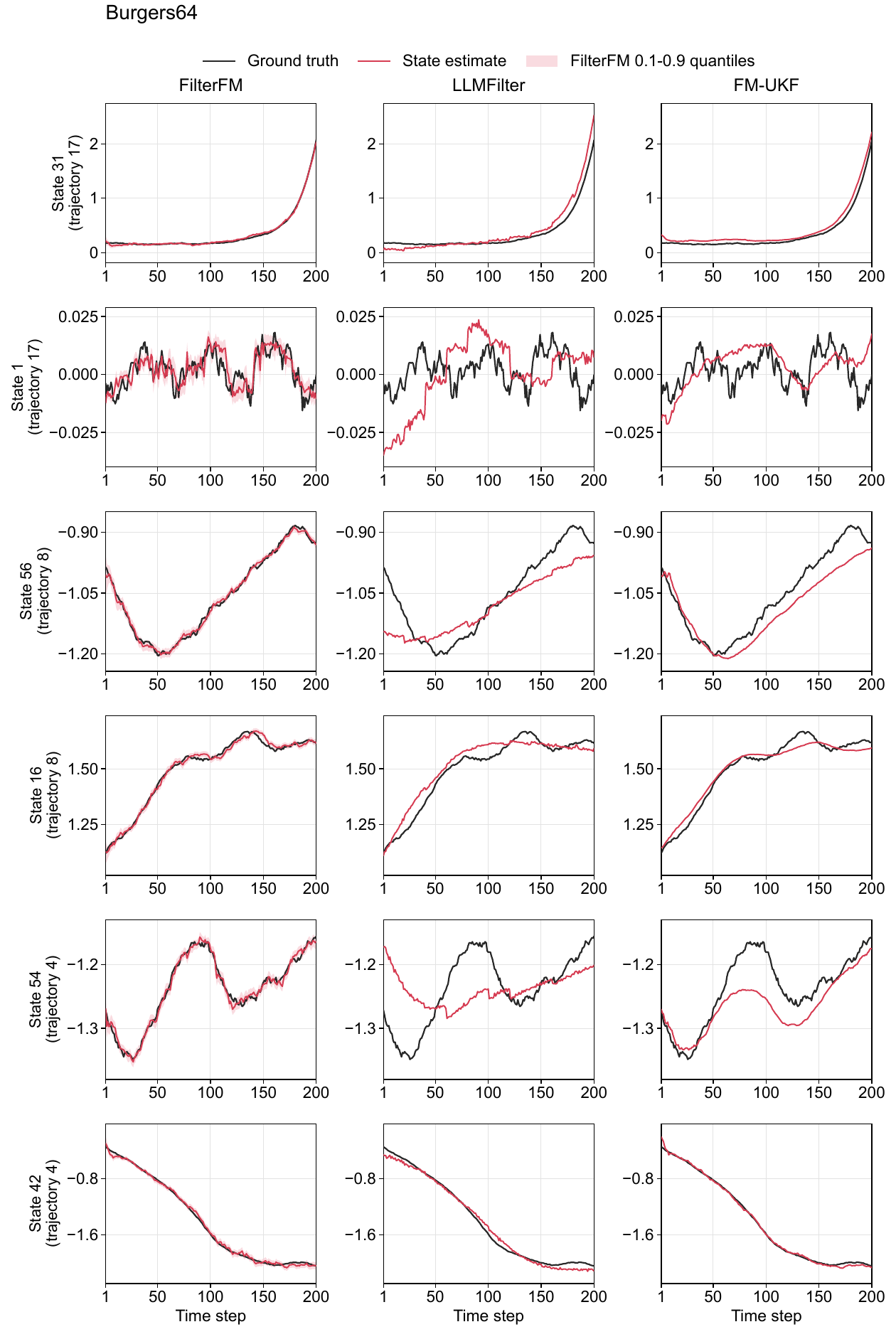}
\caption{In-domain state estimation on Burgers64: six state channels from three held-out trajectories, with the same axis limits across methods in each row.}
\label{fig:additional_state_estimation_showcase_burgers64}
\end{figure}

\begin{figure}[p]
\centering
\includegraphics[width=\linewidth,page=14]{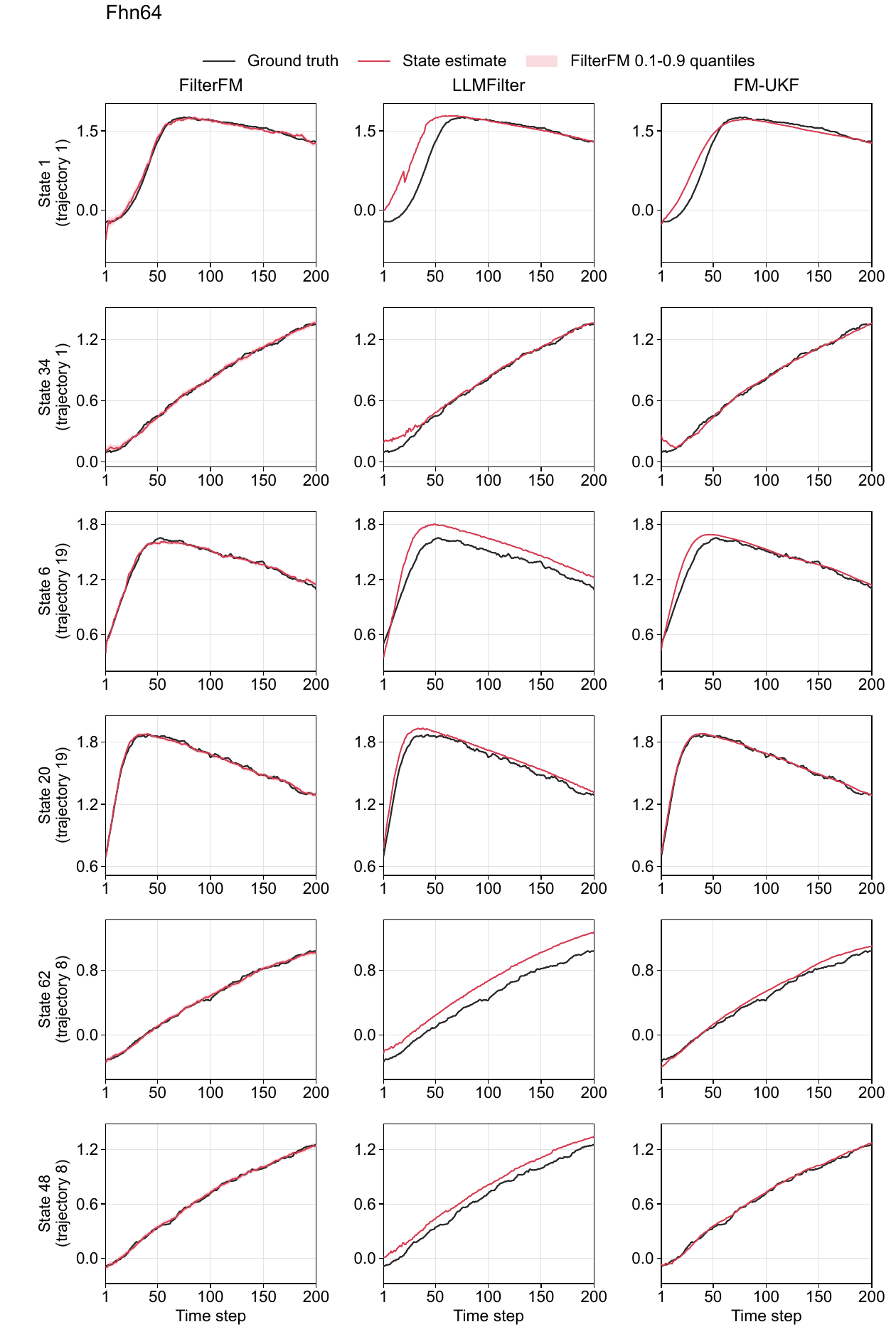}
\caption{In-domain state estimation on Fhn64: six state channels from three held-out trajectories, with the same axis limits across methods in each row.}
\label{fig:additional_state_estimation_showcase_fhn64}
\end{figure}

Figures~\ref{fig:additional_state_estimation_showcase_tracking}--\ref{fig:additional_state_estimation_showcase_fhn64} compare FilterFM, LLMFilter, and FM-UKF across all fourteen benchmark systems. Each high-dimensional system is shown through six distinct state channels from three held-out trajectories; the additional channels are spread across the state-index range without using prediction errors for selection. Each row shows the same trajectory and channel for all three methods with shared axis limits, and the FilterFM panel includes its predicted $0.1$--$0.9$ quantile interval. The low-dimensional examples retain three trajectory--channel pairs per system.

\clearpage
\subsection{Zero-Shot State Estimation}
\label{app:oos_trajectory_showcase}

These Panda systems are generated by skew-product coupling between driver and response dynamics~\citep{lai2026panda}. The 457 dynamical systems used for zero-shot evaluation are absent from the Panda training sources, including those used for high-dimensional composites. We exhaustively checked all 200-step state windows and verified that no exact duplicates exist between the evaluation set and the low-dimensional Panda or fourteen benchmark datasets.

\begin{figure}[H]
\centering
\includegraphics[width=\linewidth,page=1]{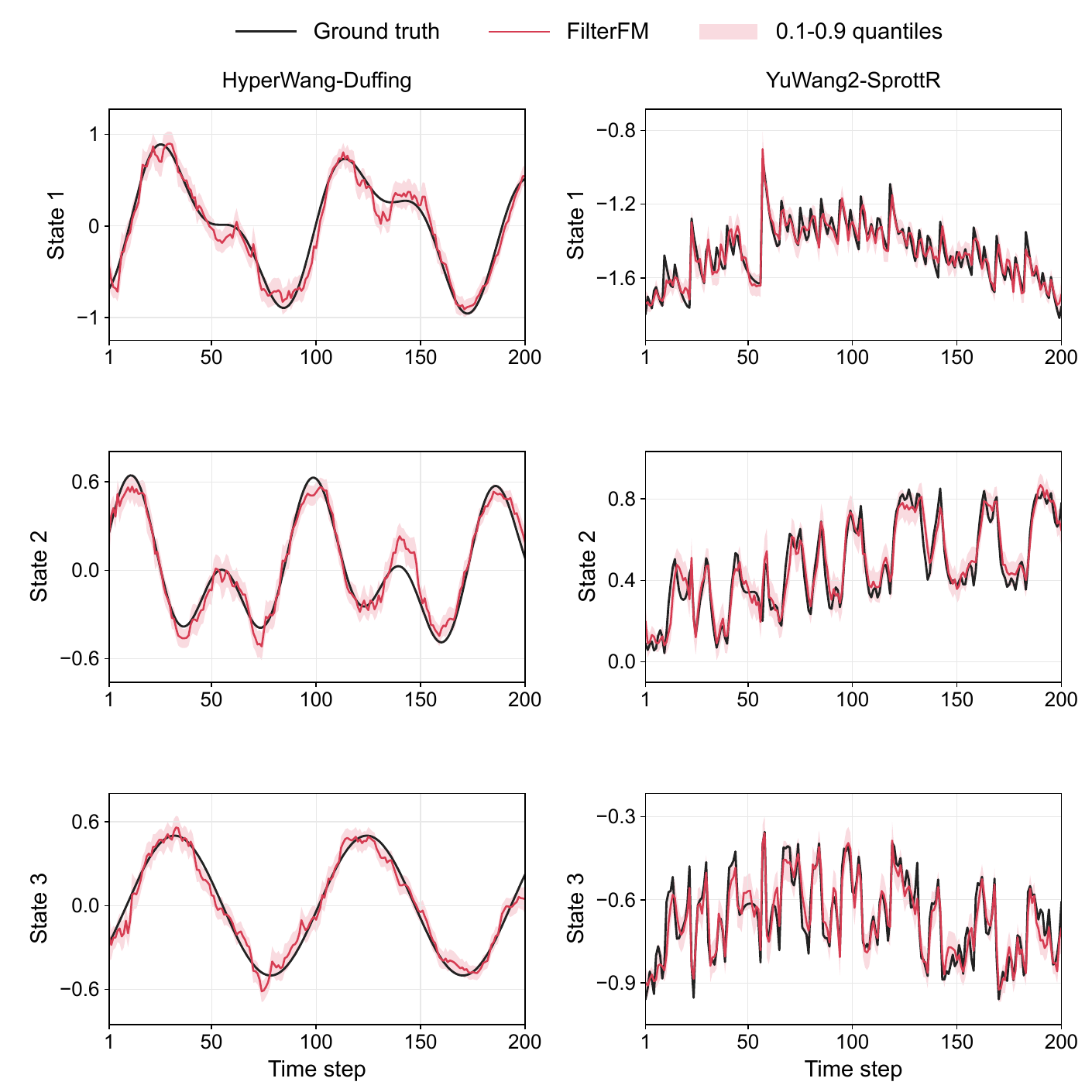}
\caption{Zero-shot state estimation on two three-dimensional held-out systems. Each column shows one system, with all three state channels arranged from top to bottom. Black and red curves denote the ground truth and FilterFM posterior median, respectively; light-red bands indicate the $0.1$--$0.9$ quantile intervals. All examples use the main checkpoint without fine-tuning.}
\label{fig:oos_heldout_trajectories_1}
\end{figure}

\begin{figure}[H]
\centering
\includegraphics[width=\linewidth,page=2]{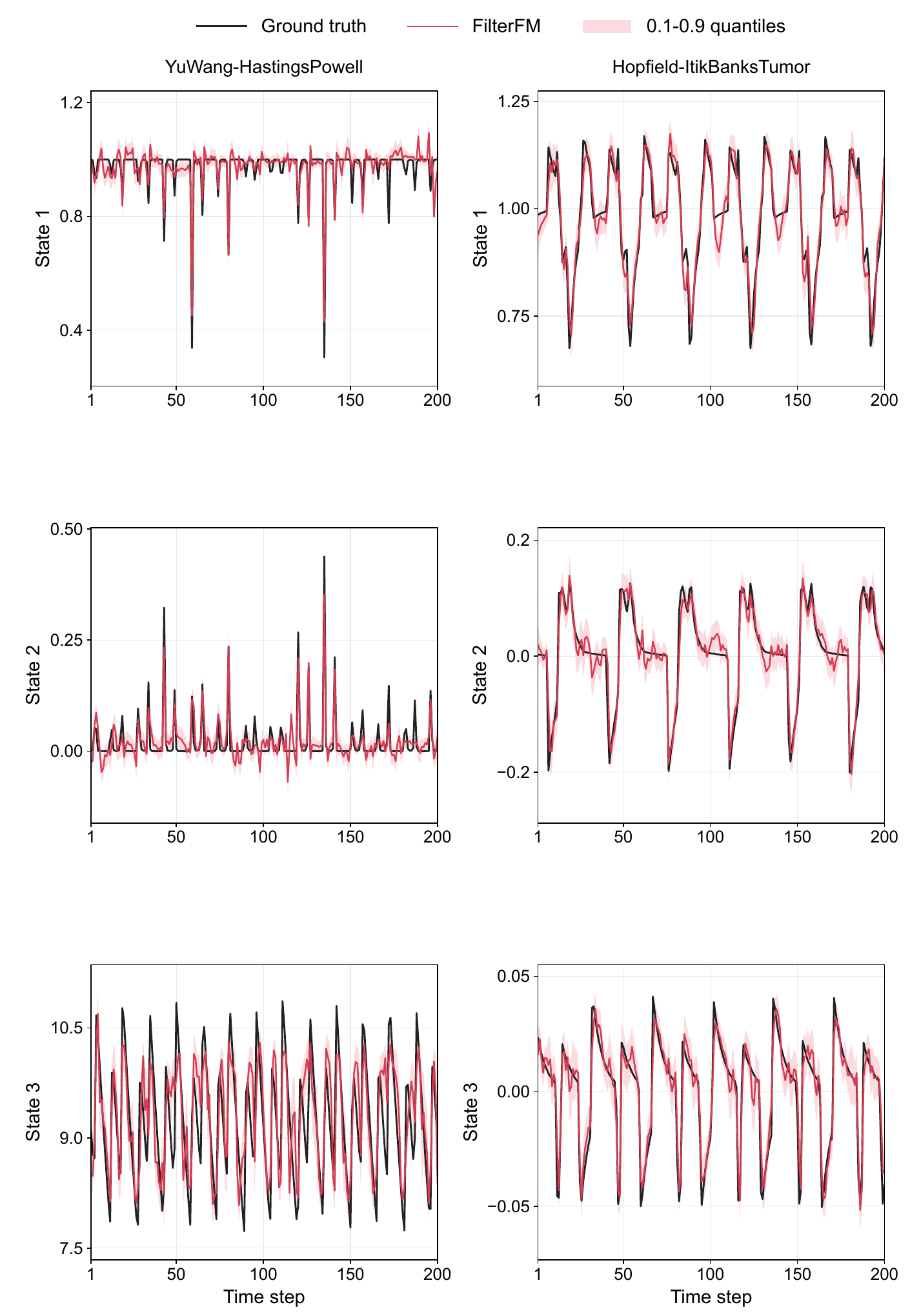}
\caption{Zero-shot state estimation on two additional three-dimensional held-out systems. Each column shows one system and all three state channels, using the same checkpoint and plotting conventions as Figure~\ref{fig:oos_heldout_trajectories_1}.}
\label{fig:oos_heldout_trajectories_2}
\end{figure}

\begin{figure}[H]
\centering
\includegraphics[width=\linewidth,page=3]{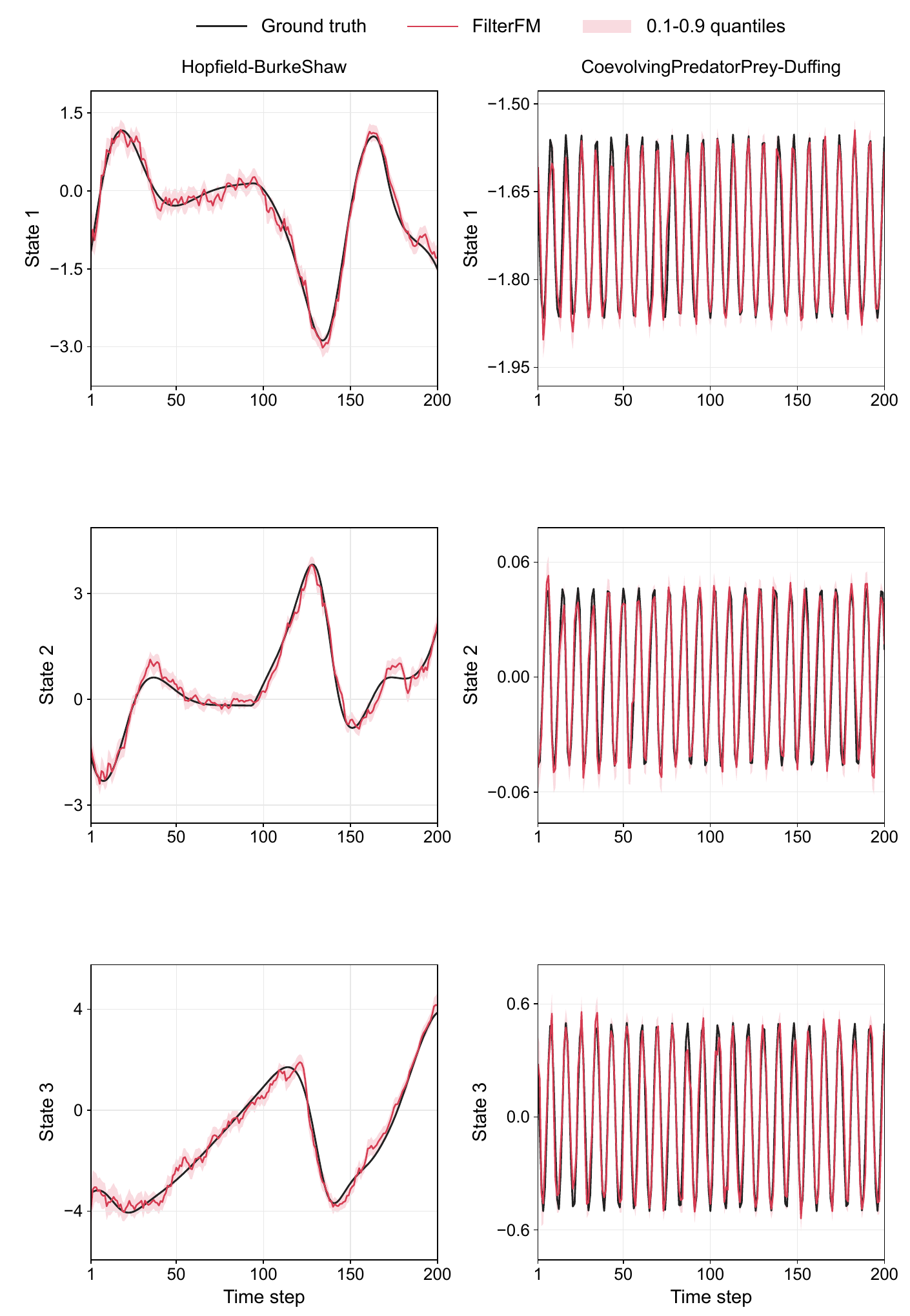}
\caption{Zero-shot state estimation on two additional three-dimensional held-out systems. Each column shows one system and all three state channels, using the same checkpoint and plotting conventions as Figure~\ref{fig:oos_heldout_trajectories_1}.}
\label{fig:oos_heldout_trajectories_3}
\end{figure}

\begin{figure}[H]
\centering
\includegraphics[width=\linewidth,page=4]{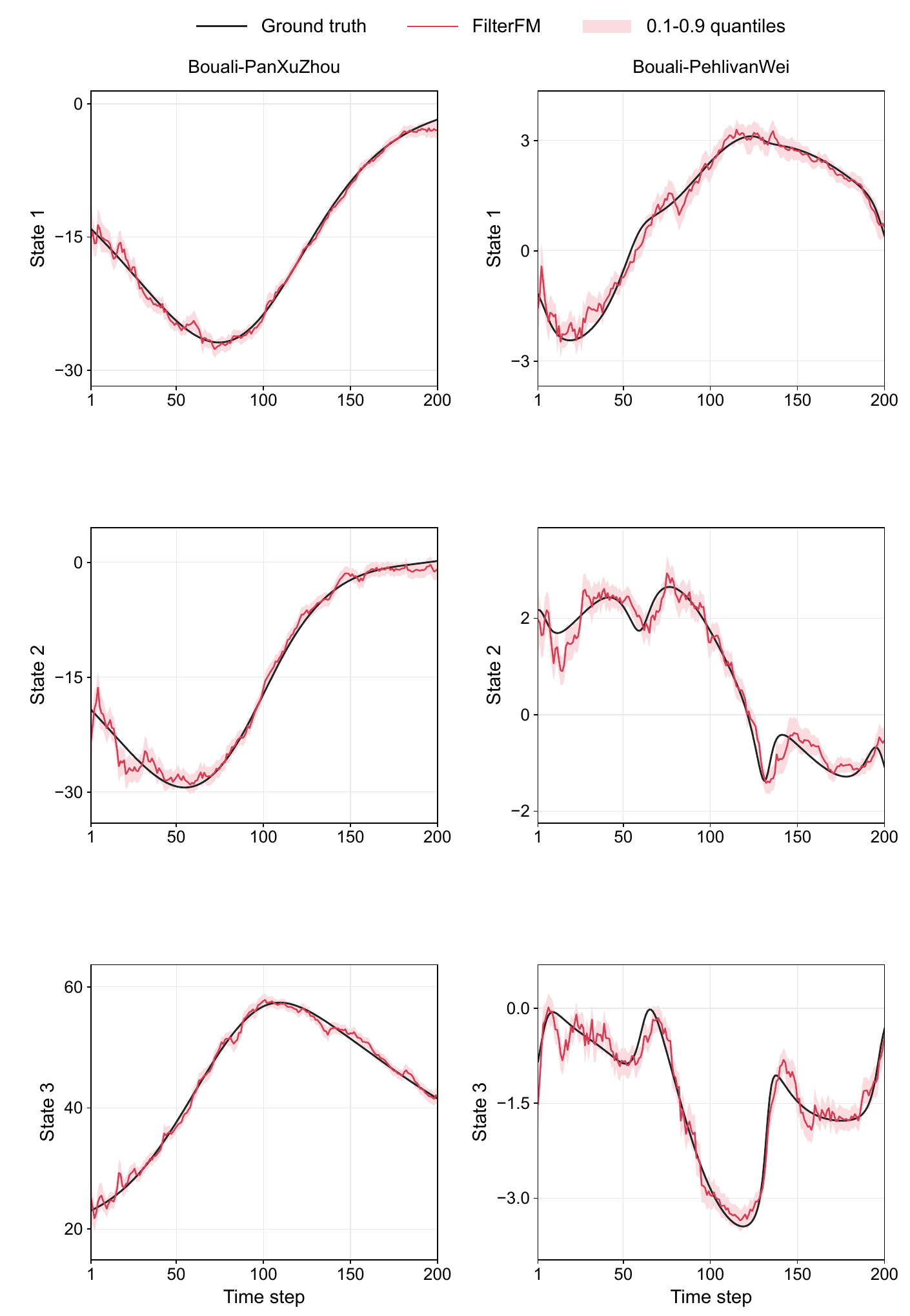}
\caption{Zero-shot state estimation on two additional three-dimensional held-out systems. Each column shows one system and all three state channels, using the same checkpoint and plotting conventions as Figure~\ref{fig:oos_heldout_trajectories_1}.}
\label{fig:oos_heldout_trajectories_4}
\end{figure}

\begin{figure}[H]
\centering
\includegraphics[width=\linewidth,page=5]{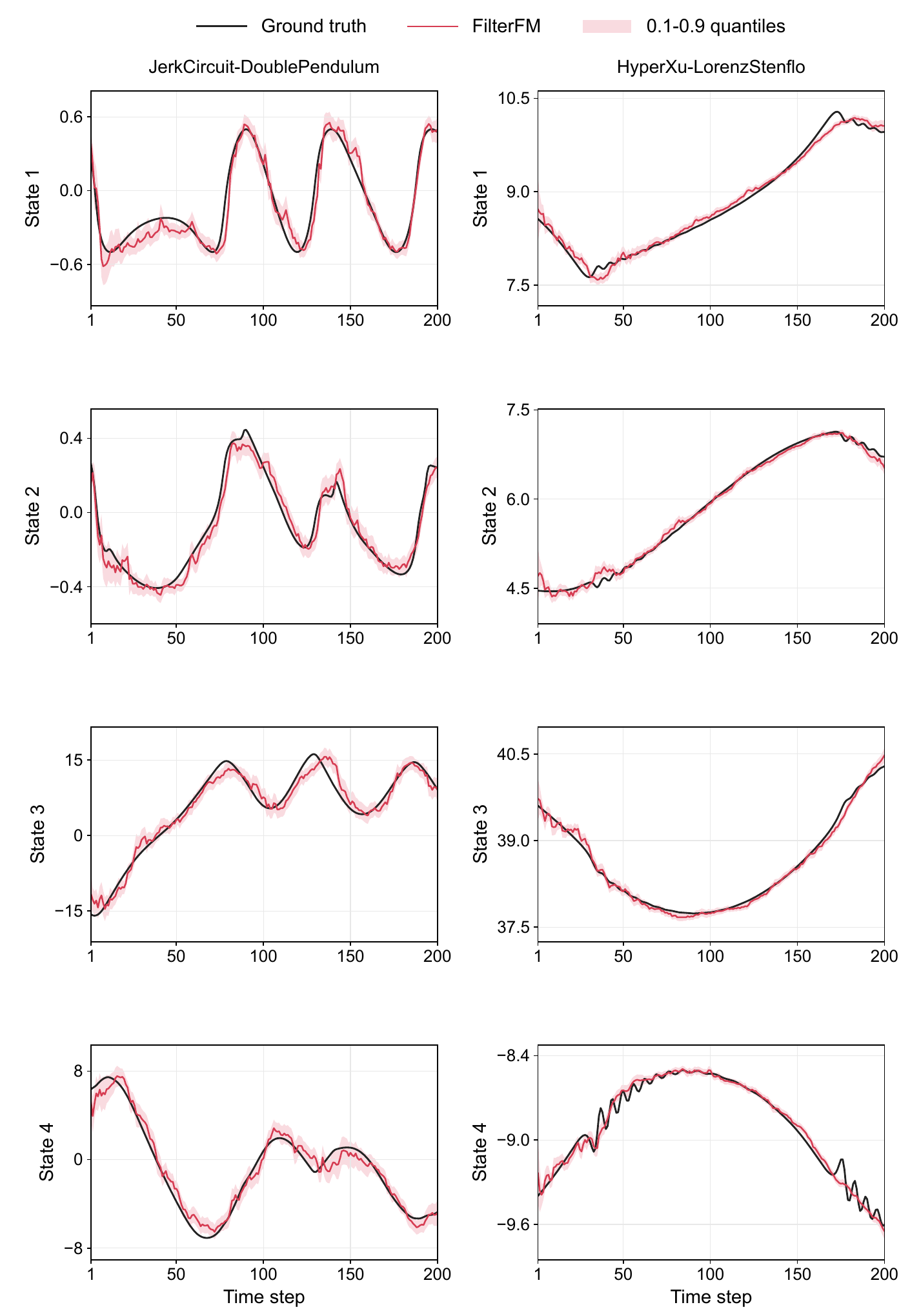}
\caption{Zero-shot state estimation on two additional four-dimensional held-out systems. Each column shows one system and all four state channels, using the same checkpoint and plotting conventions as Figure~\ref{fig:oos_heldout_trajectories_1}.}
\label{fig:oos_heldout_trajectories_5}
\end{figure}

\begin{figure}[H]
\centering
\includegraphics[width=\linewidth,page=6]{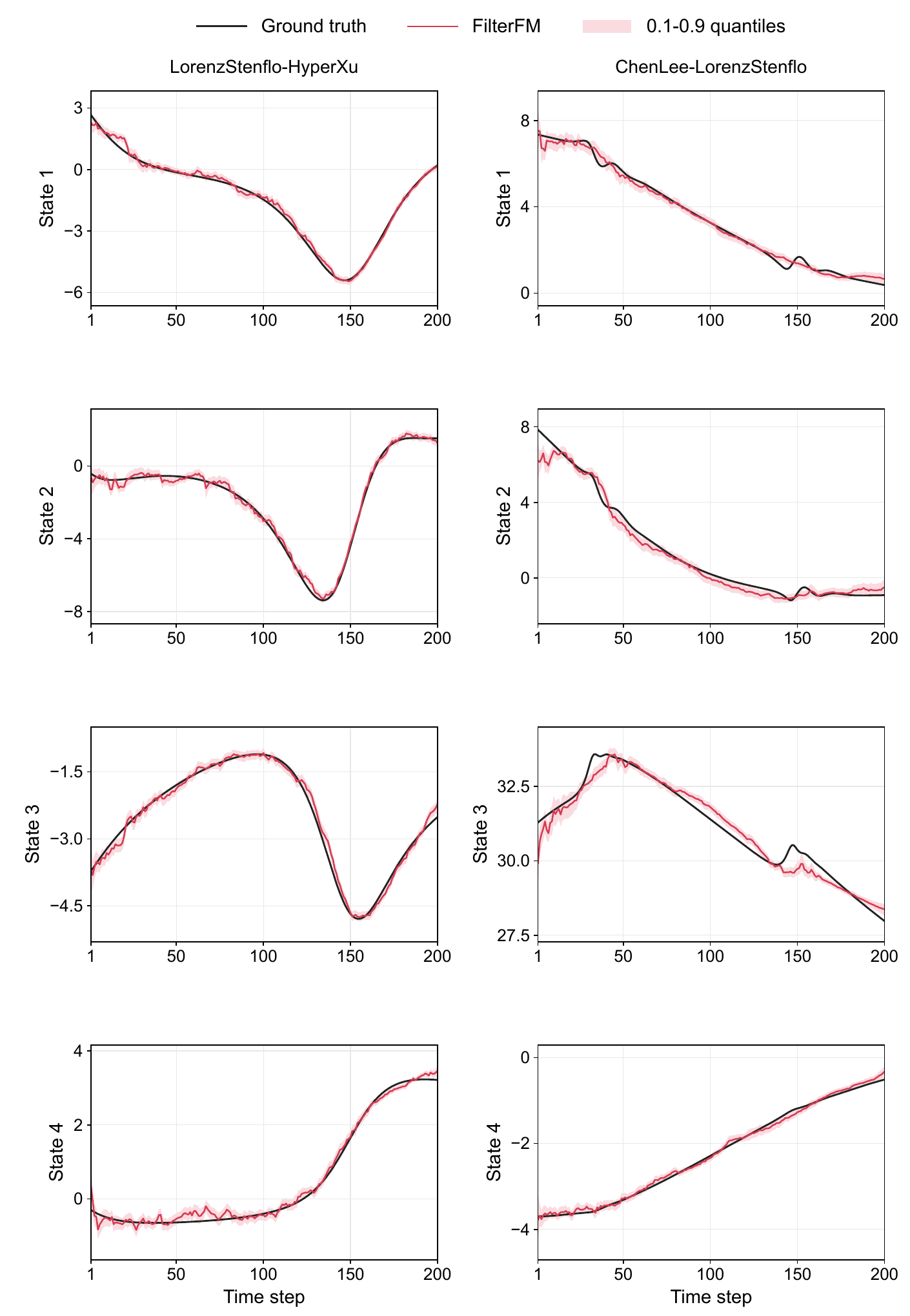}
\caption{Zero-shot state estimation on two additional four-dimensional held-out systems. Each column shows one system and all four state channels, using the same checkpoint and plotting conventions as Figure~\ref{fig:oos_heldout_trajectories_1}.}
\label{fig:oos_heldout_trajectories_6}
\end{figure}

\begin{figure}[H]
\centering
\includegraphics[width=\linewidth,page=7]{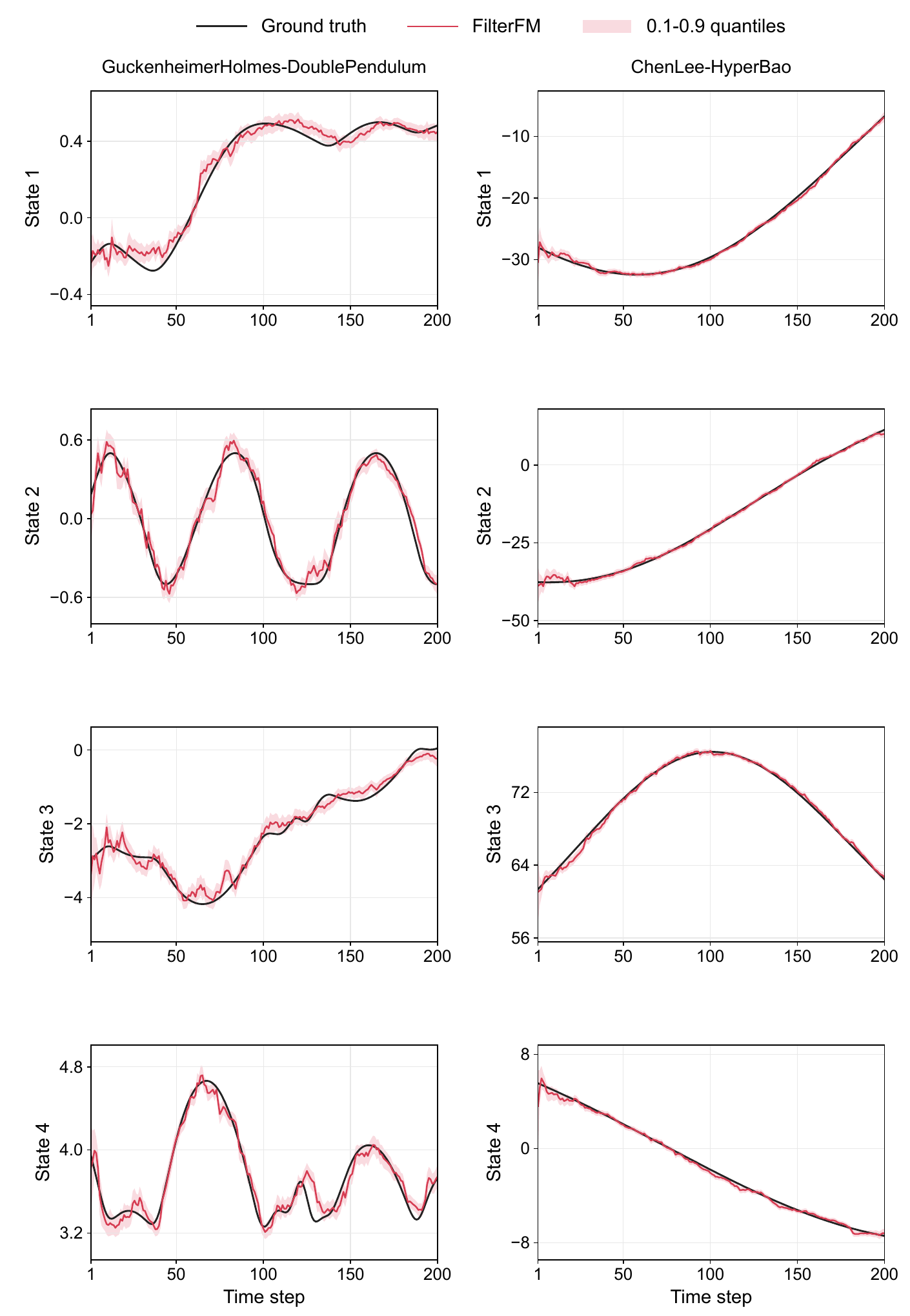}
\caption{Zero-shot state estimation on two additional four-dimensional held-out systems. Each column shows one system and all four state channels, using the same checkpoint and plotting conventions as Figure~\ref{fig:oos_heldout_trajectories_1}.}
\label{fig:oos_heldout_trajectories_7}
\end{figure}

\begin{figure}[H]
\centering
\includegraphics[width=\linewidth,page=8]{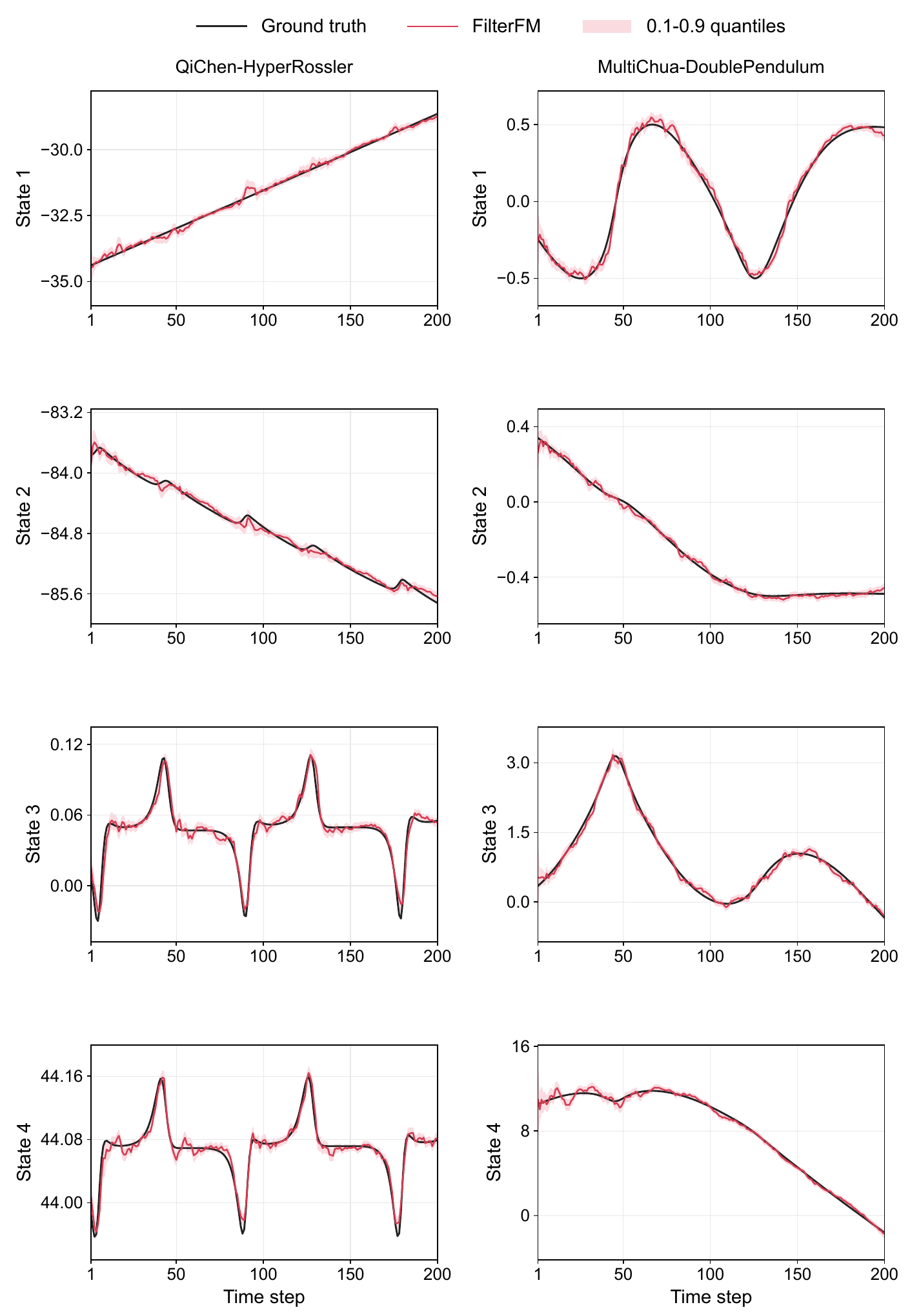}
\caption{Zero-shot state estimation on two additional four-dimensional held-out systems. Each column shows one system and all four state channels, using the same checkpoint and plotting conventions as Figure~\ref{fig:oos_heldout_trajectories_1}.}
\label{fig:oos_heldout_trajectories_8}
\end{figure}

\begin{figure}[H]
\centering
\includegraphics[width=\linewidth,page=9]{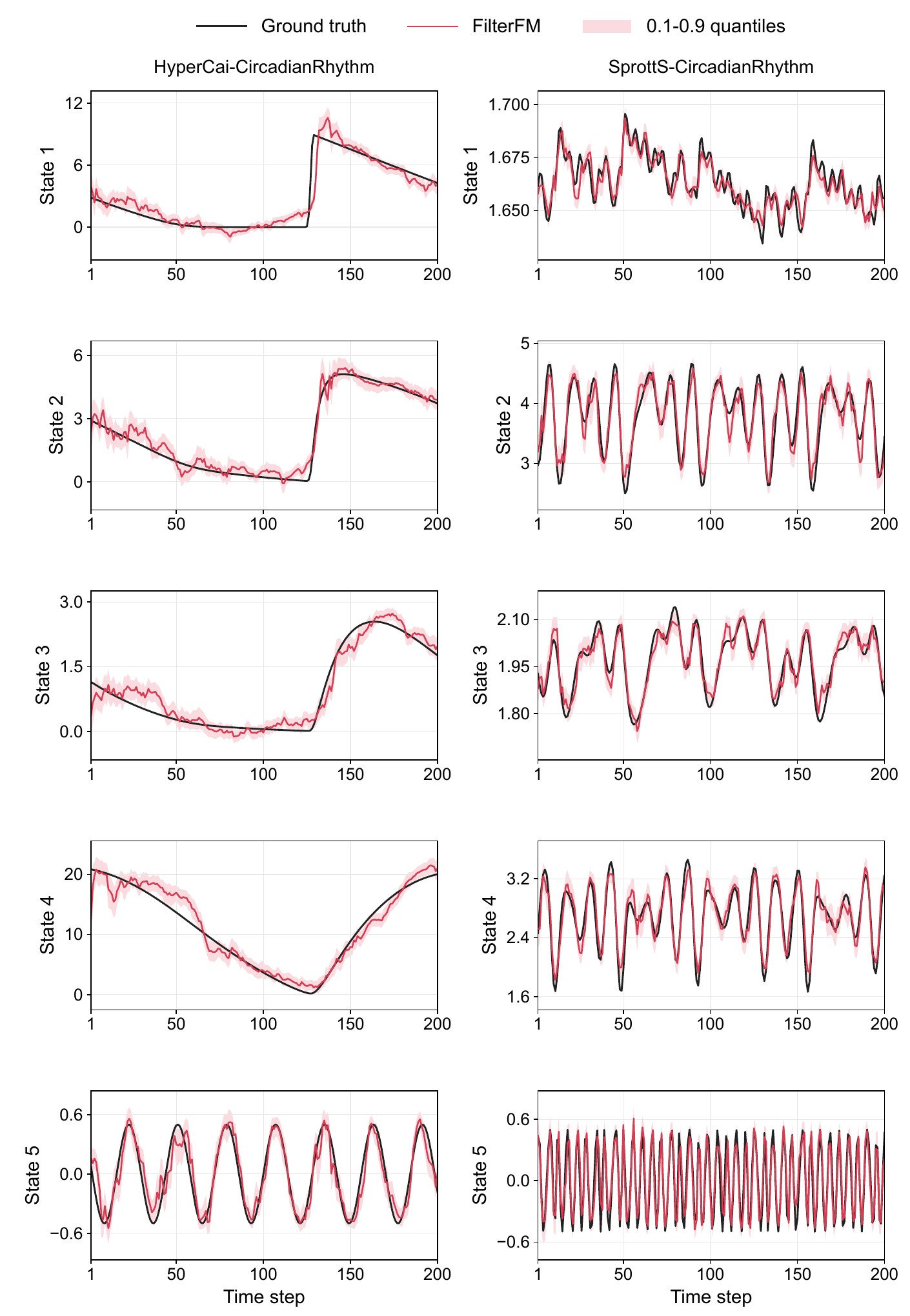}
\caption{Zero-shot state estimation on two additional five-dimensional held-out systems. Each column shows one system and all five state channels, using the same checkpoint and plotting conventions as Figure~\ref{fig:oos_heldout_trajectories_1}.}
\label{fig:oos_heldout_trajectories_9}
\end{figure}

\begin{figure}[H]
\centering
\includegraphics[width=\linewidth,page=10]{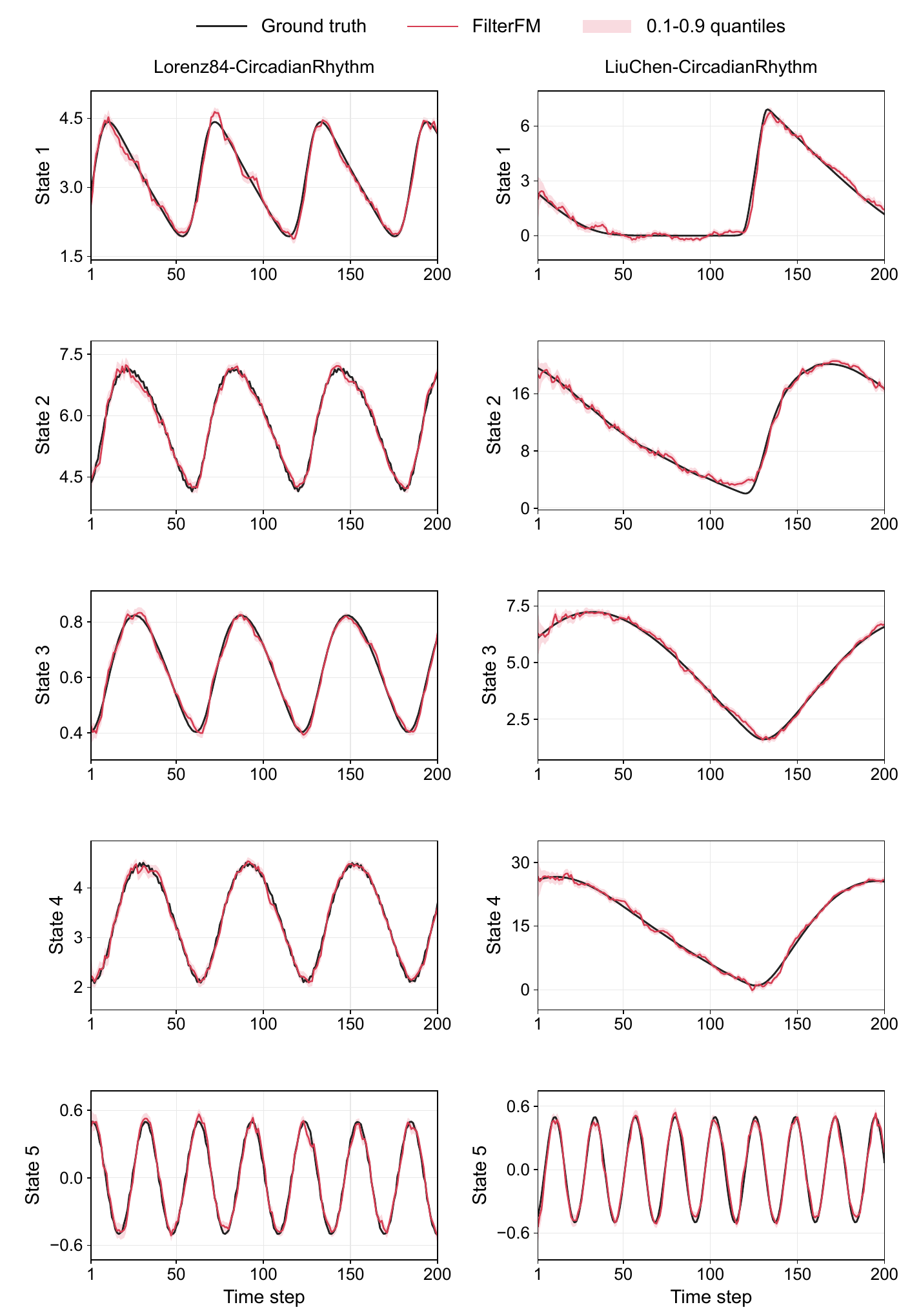}
\caption{Zero-shot state estimation on two additional five-dimensional held-out systems. Each column shows one system and all five state channels, using the same checkpoint and plotting conventions as Figure~\ref{fig:oos_heldout_trajectories_1}.}
\label{fig:oos_heldout_trajectories_10}
\end{figure}

\begin{figure}[H]
\centering
\includegraphics[width=\linewidth,page=11]{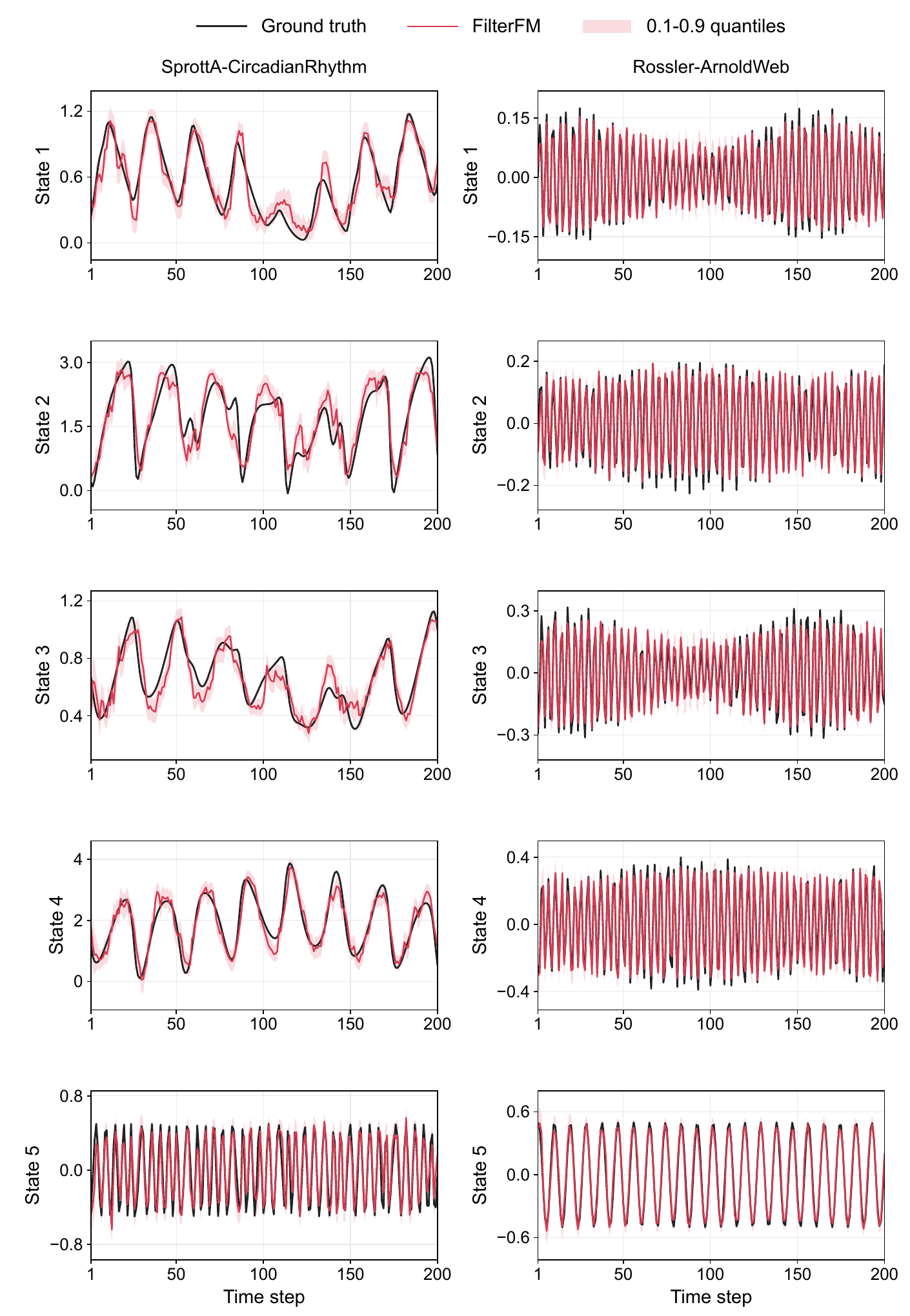}
\caption{Zero-shot state estimation on two additional five-dimensional held-out systems. Each column shows one system and all five state channels, using the same checkpoint and plotting conventions as Figure~\ref{fig:oos_heldout_trajectories_1}.}
\label{fig:oos_heldout_trajectories_11}
\end{figure}

\begin{figure}[H]
\centering
\includegraphics[width=\linewidth,page=12]{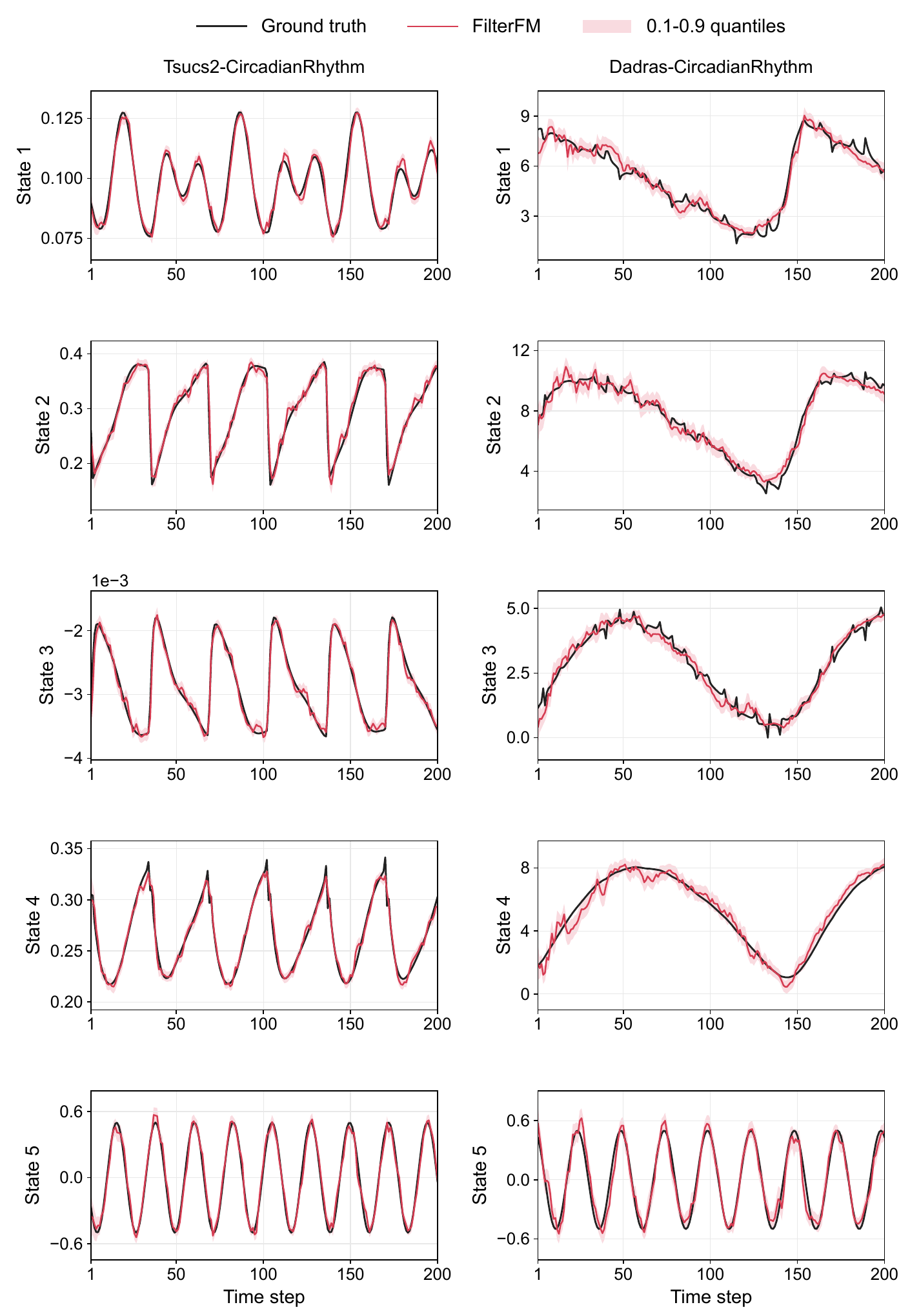}
\caption{Zero-shot state estimation on two additional five-dimensional held-out systems. Each column shows one system and all five state channels, using the same checkpoint and plotting conventions as Figure~\ref{fig:oos_heldout_trajectories_1}.}
\label{fig:oos_heldout_trajectories_12}
\end{figure}

\begin{figure}[H]
\centering
\includegraphics[width=\linewidth,page=13]{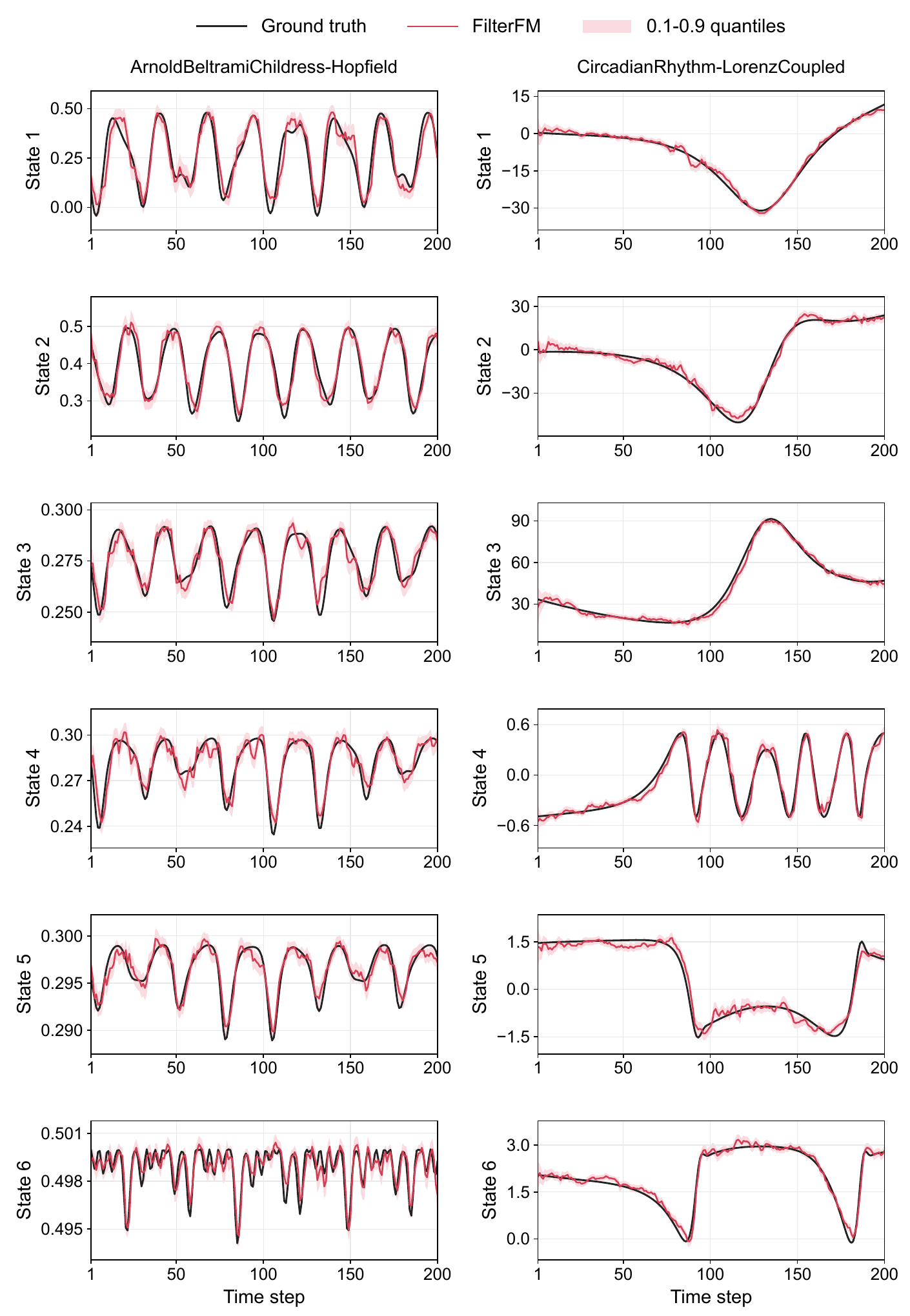}
\caption{Zero-shot state estimation on two additional six-dimensional held-out systems. Each column shows one system and all six state channels, using the same checkpoint and plotting conventions as Figure~\ref{fig:oos_heldout_trajectories_1}.}
\label{fig:oos_heldout_trajectories_13}
\end{figure}

\begin{figure}[H]
\centering
\includegraphics[width=\linewidth,page=14]{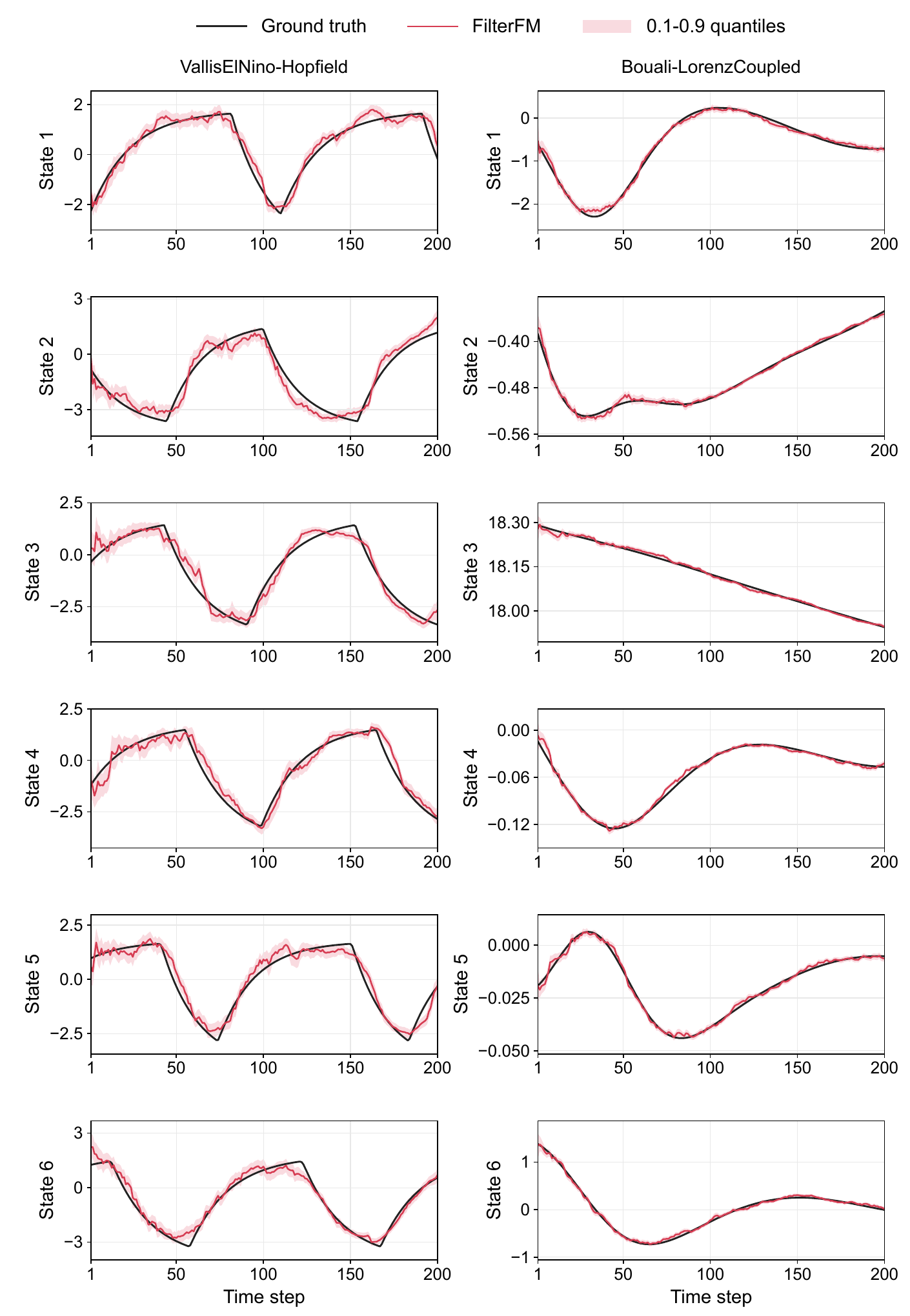}
\caption{Zero-shot state estimation on two additional six-dimensional held-out systems. Each column shows one system and all six state channels, using the same checkpoint and plotting conventions as Figure~\ref{fig:oos_heldout_trajectories_1}.}
\label{fig:oos_heldout_trajectories_14}
\end{figure}

\begin{figure}[H]
\centering
\includegraphics[width=\linewidth,page=15]{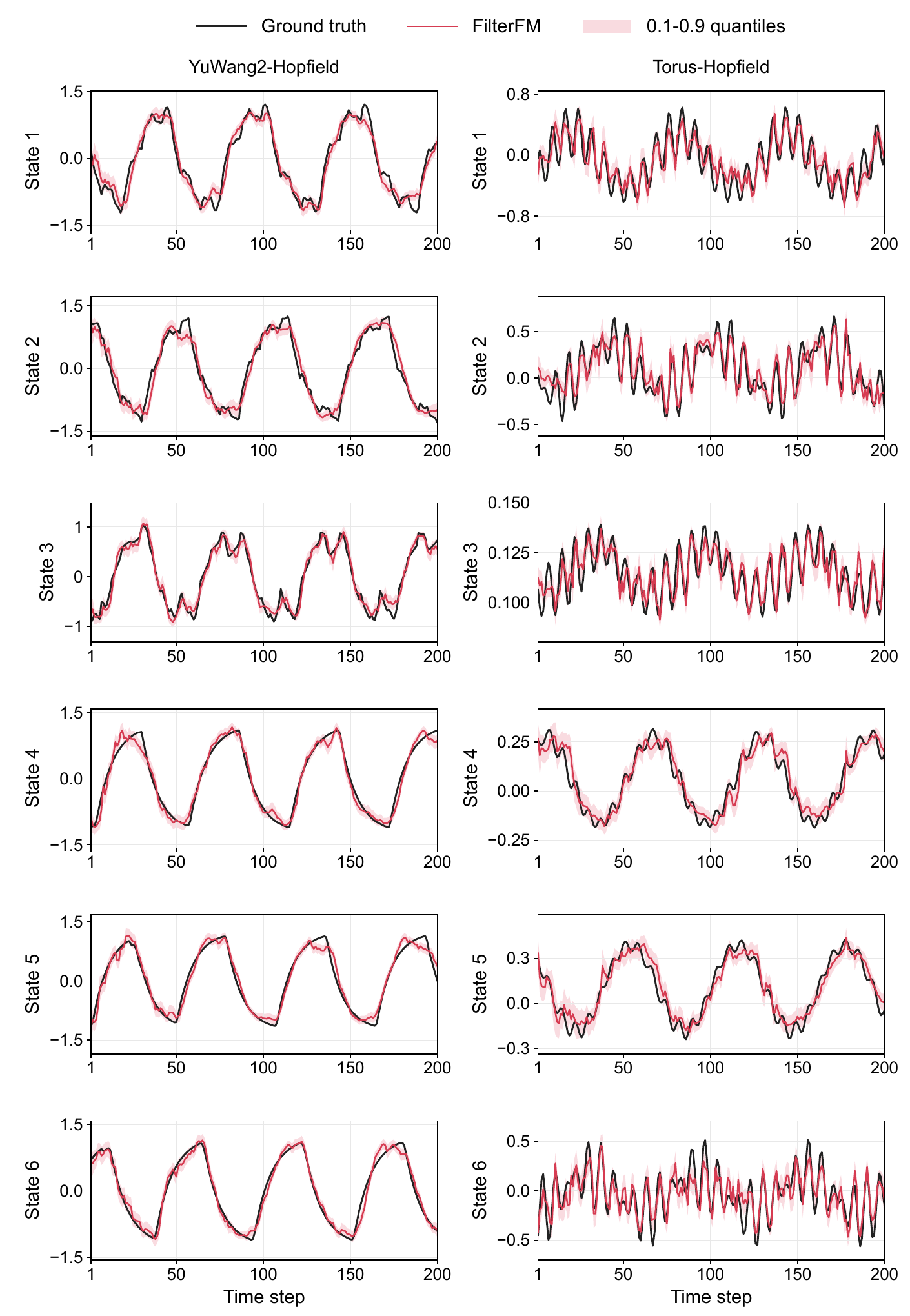}
\caption{Zero-shot state estimation on two additional six-dimensional held-out systems. Each column shows one system and all six state channels, using the same checkpoint and plotting conventions as Figure~\ref{fig:oos_heldout_trajectories_1}.}
\label{fig:oos_heldout_trajectories_15}
\end{figure}

\begin{figure}[H]
\centering
\includegraphics[width=\linewidth,page=16]{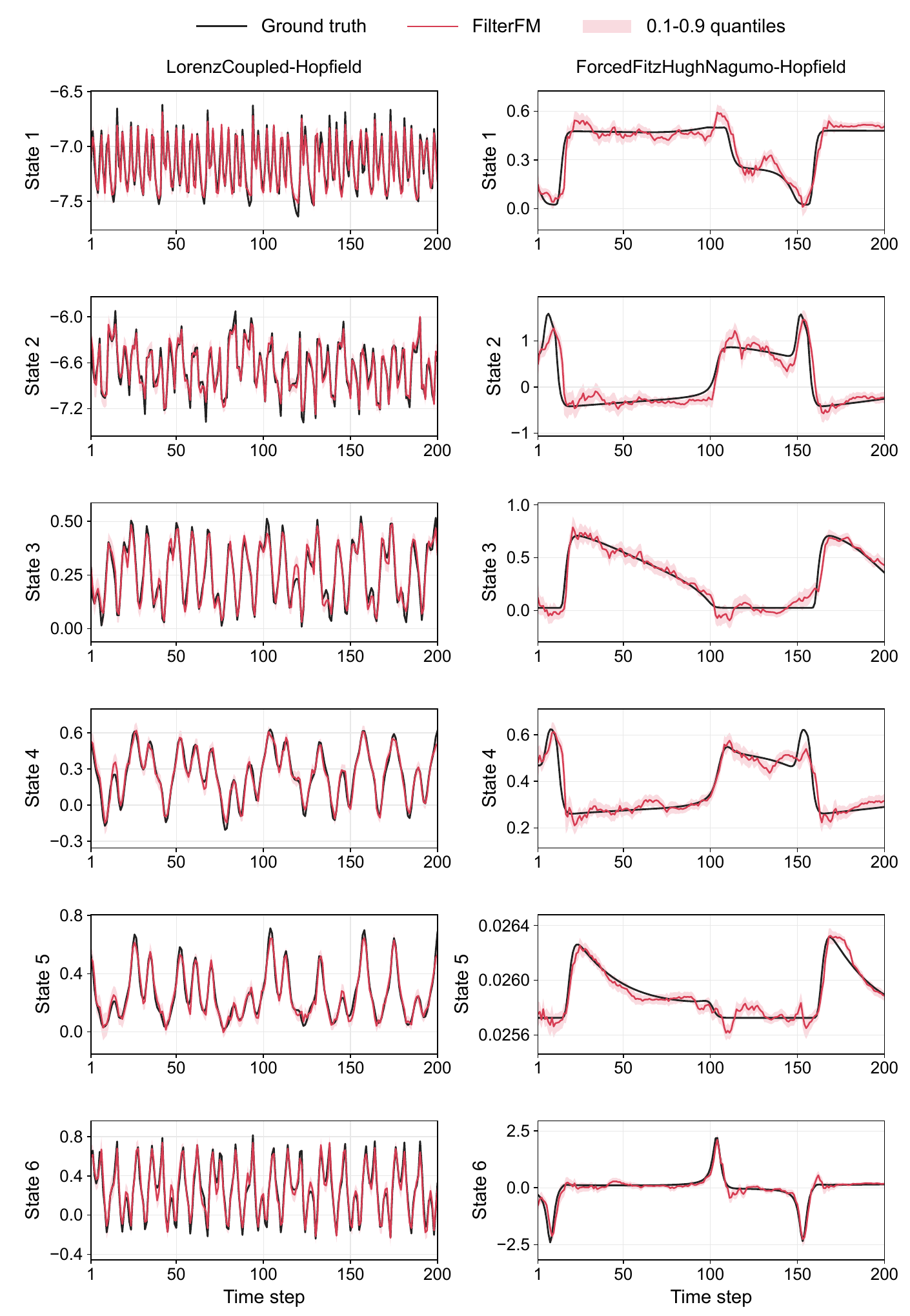}
\caption{Zero-shot state estimation on two additional six-dimensional held-out systems. Each column shows one system and all six state channels, using the same checkpoint and plotting conventions as Figure~\ref{fig:oos_heldout_trajectories_1}.}
\label{fig:oos_heldout_trajectories_16}
\end{figure}

Figures~\ref{fig:oos_heldout_trajectories_1}--\ref{fig:oos_heldout_trajectories_16} show the main model on 32 such systems: eight randomly selected systems at each dimension from three to six, with one random trajectory per system and no selection by prediction error. For each selected trajectory, we compare the estimated and true states in every channel over 200 time steps. Each figure places two systems side by side and arranges their state channels from top to bottom. Black curves show the true states, red curves show the posterior medians, and light-red bands show the predicted $0.1$--$0.9$ quantile intervals. This view exposes differences in estimation quality across channels, including deviations near abrupt changes and turning points.

\clearpage

\clearpage

\clearpage

\clearpage

\clearpage

\clearpage

\clearpage

\clearpage

\clearpage

\clearpage

\clearpage

\clearpage

\clearpage

\clearpage

\clearpage

\clearpage
\section{System-as-Prompt Inputs for LLMFilter}
\label{app:llmfilter_prompts}
\vspace{-0.7em}

The LLMFilter code repository does not include its task-specific System-as-Prompt texts, so we construct a prompt for each of our fourteen benchmark systems following the format described by \citet{LLMFilter}. Each prompt first specifies the filtering task and introduces the underlying dynamical system and observation model, then describes the dataset, including its trajectory length, number of trajectories, and observation-noise variance. A short excerpt of paired observation and state trajectories illustrates the input and target sequences, followed by a final instruction to recover the hidden states from noisy observations. We provide the resulting prompts below.

\tcbinputlisting{systemprompt,title={Tracking},listing file={prompts/system_01.txt}}

\tcbinputlisting{systemprompt,title={Double-Pendulum},listing file={prompts/system_02.txt}}

\tcbinputlisting{systemprompt,title={Selkov},listing file={prompts/system_03.txt}}

\tcbinputlisting{systemprompt,title={Oscillator},listing file={prompts/system_04.txt}}

\tcbinputlisting{systemprompt,title={Hopf},listing file={prompts/system_05.txt}}

\tcbinputlisting{systemprompt,title={Lorenz96},listing file={prompts/system_06.txt}}

\tcbinputlisting{systemprompt,title={Vissio--Lucarini 20 (VL20)},listing file={prompts/system_07.txt}}

\tcbinputlisting{systemprompt,title={LorenzUV40},listing file={prompts/system_08.txt}}

\tcbinputlisting{systemprompt,title={Vanderpol64},listing file={prompts/system_09.txt}}

\tcbinputlisting{systemprompt,title={Rossler96},listing file={prompts/system_10.txt}}

\tcbinputlisting{systemprompt,title={Duffing64},listing file={prompts/system_11.txt}}

\tcbinputlisting{systemprompt,title={Burgers64},listing file={prompts/system_12.txt}}

\tcbinputlisting{systemprompt,title={Fhn64},listing file={prompts/system_13.txt}}

\tcbinputlisting{systemprompt,title={Cubic-Sensoring100},listing file={prompts/system_14.txt}}

\end{document}